%% file: main.tex
\documentclass[10pt,twocolumn,letterpaper]{article}

\usepackage[pagenumbers]{cvpr} % To force page numbers, e.g. for an arXiv version


\definecolor{cvprblue}{rgb}{0.21,0.49,0.74}
\usepackage[pagebackref,breaklinks,colorlinks,allcolors=cvprblue]{hyperref}

\usepackage{graphicx}
\usepackage{amsmath}
\usepackage{amssymb}
\usepackage{multirow}
\usepackage{multicol}
\usepackage{cancel}
\usepackage{booktabs}
\usepackage{comment}
\usepackage{makecell}
\usepackage{booktabs}
\usepackage{wrapfig}
\usepackage{amsmath}
\usepackage{kotex}

\usepackage{float}

\DeclareMathOperator*{\argmin}{arg\,min} % Jan Hlavacek
\def\paperID{} % *** Enter the Paper ID here
\def\confName{CVPR}
\def\confYear{2025}

\newcommand{\Skip}[1]{}

\newcommand{\Loss}[1]
{
	\mathcal{L}
}

\newcommand{\AAEESOP}[1]{AESOP}

\vspace{-10pt}
\title{Auto-Encoded Supervision for Perceptual Image Super-Resolution}
\vspace{-20pt}

\author{MinKyu Lee, Sangeek Hyun, Woojin Jun, Jae-Pil Heo\thanks{Corresponding author.} \\
Sungkyunkwan University\\
{\tt\small \{2minkyulee,  hse1032, junwoojinjin, jaepilheo\}@gmail.com}
}

\begin{document}
\maketitle

\input{article/0_abstract}

\input{article/1_intro}

\input{article/4_relatedworks}
\input{article/2_body}

\input{article/3_experiments}

\input{article/5_conclusion}

\clearpage
% Ack
\paragraph{Acknowledgement}
This work was supported in part by MSIT/IITP (No. 2022-0-00680, 2020-0-01821, RS-2019-II190421, RS-2024-00459618, RS-2024-00360227, RS-2024-00437102, RS-2024-00437633), and MSIT/NRF (No. RS-2024-00357729).

{
    \small
    \bibliographystyle{ieeenat_fullname}
    \bibliography{main}
}

\input{article/X_suppl}

\end{document}

%% file: article/0_abstract.tex
\vspace{-10pt}
\begin{abstract}
This work tackles the \textbf{fidelity} objective in the \textbf{perceptual} super-resolution~(SR) task.
Specifically, we address the shortcomings of pixel-level $\mathcal{L}_\text{p}$ loss ($\mathcal{L}_\text{pix}$) in the GAN-based SR framework.
Since $\mathcal{L}_\text{pix}$ is known to have a trade-off relationship against perceptual quality, prior methods often multiply a small scale factor or utilize low-pass filters.
However, this work shows that these circumventions fail to address the fundamental factor that induces blurring.
Accordingly, we focus on two points: 1) precisely discriminating the subcomponent of $\mathcal{L}_\text{pix}$ that contributes to blurring, and 2) only guiding based on the factor that is free from this trade-off relationship.
We show that they can be achieved in a surprisingly simple manner, with an Auto-Encoder (AE) pretrained with $\mathcal{L}_\text{pix}$. Accordingly, we propose the Auto-Encoded Supervision for Optimal Penalization loss ($\mathcal{L}_\text{AESOP}$), a novel loss function that measures distance in the \textbf{AE space}~\footnote{AE space indicates the space \textit{after} the decoder, not the bottleneck.}, instead of the raw pixel space.
By simply substituting $\mathcal{L}_\text{pix}$ with $\mathcal{L}_\text{AESOP}$, we can provide effective reconstruction guidance without compromising perceptual quality.
Designed for simplicity, our method enables easy integration into existing SR frameworks. Extensive experiments demonstrate the effectiveness of AESOP.
\end{abstract}

%% file: article/1_intro.tex
\vspace{-10pt}
\begin{figure*}[t]
    \centering
    \includegraphics[width=\linewidth]{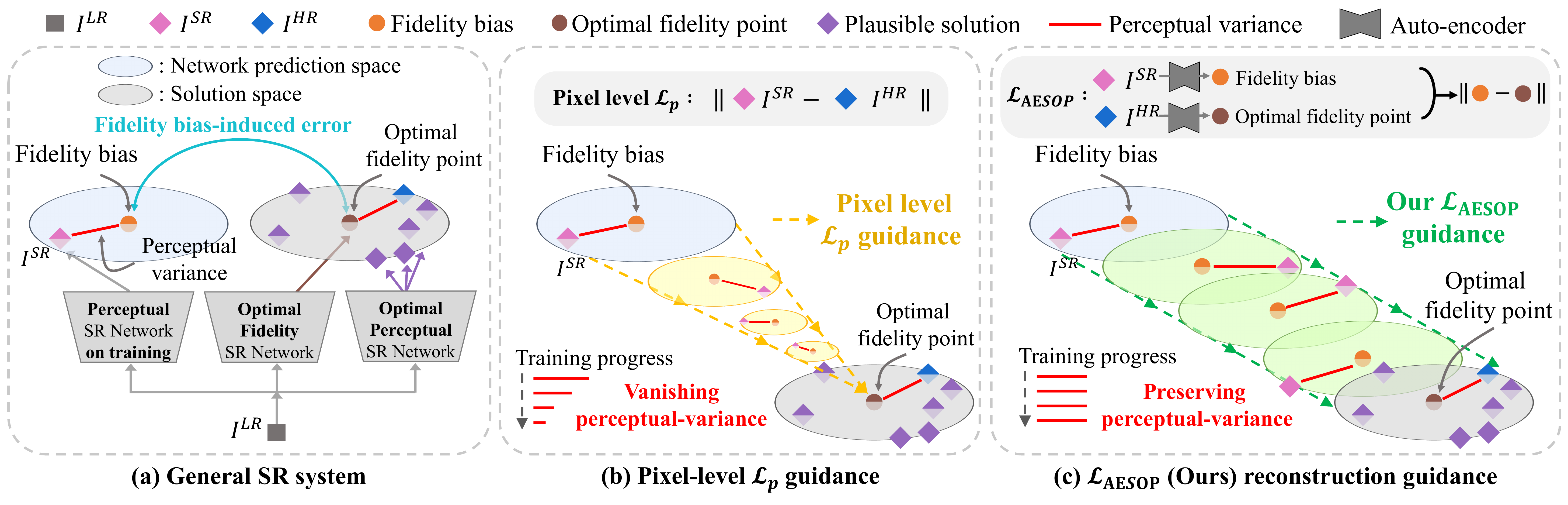}
    \vspace{-25pt}
    \caption{
    Conceptual illustration of the proposed \AAEESOP{} loss and the pixel-level $\mathcal{L}_\text{p}$ reconstruction guidance employed in typical perceptual SR methods.
    \textbf{(a)}
    Fidelity oriented SR network trained with $\mathcal{L}_\text{pix}$ estimates the average over plausible solutions (i.e., the optimal fidelity point).
    Meanwhile, perceptual SR involves a range of multiple solutions, standing around the optimal fidelity point.
    Thus, we identify two fundamental components of a perceptual SR image as 1) the perceptual variance factor~(red line), a factor that possesses randomness and contributes to realistic textures, and 2) the fidelity bias term~(orange dot), the residual blurry component of an SR image, contributing to the overall fidelity, apart from the perceptual variance.  
    \textbf{(b)}
    Typical perceptual SR methods adopt $\mathcal{L}_\text{pix}$ for reconstruction guidance, which pushes the perceptual variance factor to vanish. Thus, when combined with perceptual quality oriented losses that encourage this variance factor, conflict arises, leading to suboptimal performance.
    \textbf{(c)}
    In contrast, $\Loss{}_\text{AESOP}$ only penalizes the fidelity bias-induced error, while preserving these critical perceptual variance factors. This ensures improved fidelity without sacrificing perceptual quality. 
    }
    \label{fig:main_figure}
    \vspace{-10pt}
\end{figure*}

\section{Introduction}
\vspace{-5pt}

% ---- sisr intro
Image Super-Resolution is a fundamental challenge in image processing, where the goal is to reconstruct an unknown high-resolution (HR) image from its low-resolution (LR) counterpart. Recent advances in this field have branched into two distinct mainstreams; fidelity-oriented SR and perceptual quality oriented SR~(perceptual SR).
%
% ---- fidelity
Fidelity-oriented SR methods aim for high fidelity towards the HR image on a pixel-wise basis. These methods
generally adopt the per-pixel reconstruction loss $\Loss{}_\text{pix}$ (i.e., pixel-level $\Loss{}_\text{p}$ loss), thereby regressing the unique point that minimizes the expected error. This unique point of minimum expected error is known as the average point over multiple plausible solutions~\cite{eco}, which we will refer to as the \textit{optimal fidelity point} (brown dot in Fig.\ref{fig:main_figure}.(a)) throughout this work.  
%
% ---- perceptual SR
Another branch of research is the perceptual SR task where the emphasis is on generating visually plausible SR images rather than mere minimum pixel-wise error. Notably, the inherent ill-posedness leads perceptual SR to exhibit a range of variant realistic solutions (multiple purple points in Fig.\ref{fig:main_figure}.(a)), each pivotal to the aforementioned optimal fidelity point.

% ---- PD trade-off and conflict btw L1 and percep

Here, the representative framework in perceptual SR is the SRGAN-based~\cite{SISR6_SRGAN} framework, which utilizes perceptual quality oriented losses~\cite{SISR7_ESRGAN, perceptual_loss_vgg} together with $\Loss{}_\text{pix}$, the \textit{de facto} training scheme.
Yet, since these methods rely on $\Loss{}_\text{pix}$ as their fidelity loss term, they cannot avoid the blurring phenomenon as shown in the perception-distortion (PD) trade-off~\cite{perception_distortion_tradeoff}. 
To address this, they either apply a small coefficient~\cite{SISR6_SRGAN, SISR7_ESRGAN} to $\Loss{}_\text{pix}$ or use low-pass filters~(LPF)~\cite{lowpass1, lowpass2} before calculating $\Loss{}_\text{pix}$.
However, we point out that these circumventions result in suboptimal performance, as they misinterpret the implications of $\Loss{}_\text{pix}$ and fail to distinguish between factors that cause blurring and those that do not.

Accordingly, this work revisits $\Loss{}_\text{pix}$ and aims to correctly analyze the implications of it, by introducing
two key factors of an SR image: 1) the \textit{perceptual variance} factor, and 2) the \textit{fidelity bias} factor. 
Here, the perceptual variance is a necessary variance that captures realistic textures and fine details (red line in Fig.\ref{fig:main_figure}.(a)). Meanwhile, the fidelity bias is the residual component of the SR image, apart from the perceptual variance factor. This can be understood as the blurry average solution without the fine-grained texture that possesses randomness (no variance, Fig.\ref{fig:lossmap}.(d)), or can also be understood as the centroid of the distribution where the SR image was originally expected to be sampled from (orange dot in Fig.\ref{fig:main_figure}.(a)). For an optimal SR image (or the HR), the fidelity bias of itself is the optimal fidelity point.

In terms of these two key components defined above, we will show that $\Loss{}_\text{pix}$ is identical to minimizing both the fidelity bias induced errors and the perceptual variance factor. 
However, while minimizing the fidelity bias induced error is an intended aspect of $\Loss{}_\text{pix}$, vanishing perceptual variance is not suitable in perceptual SR.
%
% =================== Perceptual Variance
Specifically, when the perceptual variance factor is minimized, the prediction space degenerates and the SR image converges to the blurry average image as in Fig.\ref{fig:main_figure}.(b). 
Meanwhile, perceptual quality oriented losses aim to preserve this necessary perceptual variance, which indicates a conflict against $\Loss{}_\text{pix}$. 
Consequently, as long as the SR network receives reconstruction guidance from $\mathcal{L}_\text{pix}$, the perceptual quality oriented losses cannot converge to an optimal point, limiting the visual quality.

Now, we focus on the other counterpart of the SR image, the fidelity bias, and the error induced by it. The fidelity bias induced error indicates the overall degree of misalignment between the prediction space and the solution space.
Accordingly, reducing the fidelity bias induced error is identical to aligning the centroids of the prediction space and the solution space, without altering the range of the prediction space (i.e., preserving perceptual variance), as in Fig.\ref{fig:main_figure}.(c). 
Thus, by reducing the fidelity bias induced error, we can achieve improved fidelity without degrading perceptual variance, which is a highly desired aspect for optimal reconstruction guidance in the perspective of perceptual SR.

Accordingly, this motivates us to design a novel reconstruction loss that can precisely discriminate the fidelity bias and the perceptual variance, and solely penalize based on the fidelity bias term. 
Notably, we will show that this can be surprisingly simplified with a pretrained Auto-Encoder (AE).
We will pretrain an AE with $\Loss{}_\text{pix}$, and take $\Loss{}_\text{p}$ in the AE space instead of raw pixel space.
Since $\Loss{}_\text{pix}$ removes perceptual variances, loss in the AE space will enable us to penalize solely based on fidelity biases. Importantly, we are conversely taking advantage of the vanishing perceptual variance phenomenon of $\mathcal{L}_\text{pix}$, which was observed as a critical limitation. We refer to this as the Auto-Encoded Supervision for Perceptual SR~(\AAEESOP{}). 

In summary, our contributions can be simplified as follows. First, we point out the implications of $\mathcal{L}_\text{pix}$ in the context of the perceptual SR task, which has often been misunderstood. Second, we propose a novel reconstruction loss that only penalizes the fidelity bias factors, thereby preserving visually important perceptual variance factors of SR images. Finally, we provide extensive experiments that validate the effectiveness of \AAEESOP{}, leading to significant improvement in the perceptual SR task. 

\vspace{5pt}
\noindent\textbf{\textit{Disclaimer.}}
Perceptual oriented losses are responsible for handling the preserved perceptual variance term, thereby generating realistic textures and fine-details. We keep improvements on these losses out of the scope of this work.

%% file: article/4_relatedworks.tex
\section{Related work}
\vspace{-5pt}

\noindent\textbf{Fidelity-oriented SR.}
The pioneering works~\cite{SISR1_SRCNN, SISR3_VDSR}, CNN-based~\cite{SISR2_EDSR, SISR4_RCAN, SISR5_SAN, HAN, NLSA, IGNN, oisr} and Swin Transformer~\cite{swin} based methods~\cite{IPT, EDT, SISR_HAT, swinfir, SwinIR} have shown remarkable improvements. The majority of these methods employ the per-pixel loss $\mathcal{L}_\text{pix}$ as their sole objective.
This leads them to estimate the average over multiple solutions~\cite{hyun2020varsr, SRFlow}, resulting in high PSNR scores but is empirically shown to be blurry. 
In our work, the \textit{optimal fidelity point} is the optimal estimation of these fidelity-oriented SR methods.

\noindent\textbf{Perceptual SR.} 
The emphasis here is on visual quality over mere per-pixel error. These methods~\cite{SISR6_SRGAN, USRGAN, SPSR, ranksrgan} commonly adopt the SRGAN~\cite{SISR6_SRGAN}-based framework, integrating $\mathcal{L}_\text{pix}$ with perceptual-quality-oriented losses such as perceptual loss~\cite{perceptual_loss_vgg} and adversarial loss~\cite{gan}. This framework aims to enhance visual quality while retaining a fair amount of fidelity to the HR image. Recently, diffusion-based SR methods~\cite{diff_stablesr, diff_supir, diff_sr3} have shown significant progress. However, due to their limitations in SR tasks without complex degradations and the high computational cost, GAN-based SR remains as one of the main branches of research~\cite{does_diff_beat_gan_sr}. We limit the scope of our work to GAN-based SR methods.

\noindent\textbf{Improvements in the SRGAN-framework.}
Advancements were made on the adversarial loss~\cite{SISR7_ESRGAN, SANGAN}, discriminator architecture~\cite{sed, calgan} and the perceptual loss~\cite{PDL, dsd, srooe}, and also in enhancing GAN stability~\cite{SPSR,details_or_artifact, desra, SANGAN}.
Despite remarkable improvements, all these efforts primarily concentrate on perceptual-oriented loss factors. Meanwhile, the use of $\mathcal{L}_\text{pix}$ for perceptual SR tasks has not been thoroughly investigated. 
To the best of our knowledge, this is the first attempt that successfully tackles the \textit{fidelity}-loss in a \textit{perceptual} SR framework, removing the problematic pixel-level reconstruction loss in the SRGAN-framework.

%% file: article/2_body.tex
\section{Revisiting per-pixel loss in perceptual SR}
\label{Section:RevisitingL1}
\vspace{-5pt}

\textbf{Considering the Oracle case.}
Consider an optimal perceptual SR network that can sample images from the true posterior. By construction, SR images generated from this network are valid solutions but are not necessarily a pixel-wise exact match to the specific HR image instance in our dataset, due to the inherent ill-posed nature of SR. 
Yet, $\mathcal{L}_\text{pix}$ compares two images on a strict pixel-wise basis. This results in penalizing the SR image, despite it being an ideal solution in the context of perceptual SR, by construction.

\noindent\textbf{Revisiting $\mathcal{L}_\text{pix}$ in perceptual SR.}
The phenomenon where even an optimal network gets penalized under $\mathcal{L}_\text{pix}$ results in blurred texture.
This is grounded in the fact that training with $\mathcal{L}_\text{pix}$ is effectively a Maximum Likelihood Estimation, which jointly minimizes both bias errors and also \textit{variance in predictions}. Specifically, prior works \cite{james2003varianceandbias, lee2019harmonizing} have shown that this can be decomposed into jointly minimizing two terms: the systematic-effect~(SE) term and the variance-effect~(VE) term, which are induced by the bias and variance of the prediction, respectively. Formally, given a symmetric loss function $\Loss{}$ and $y \sim p(y|x)$ for an input $x$, the training objective 
$\min_{\hat{y}} \mathbb{E}_{y,\hat{y}}[\Loss{} (y,\hat{y})]$
can be simplified~\footnote{The irreducible variance term of $y$ is omitted here.} as:
\begin{equation}
\begin{aligned}
    \min_{\hat{y}} \{
    \underbrace{
    \mathbb{E}_y[{\Loss{}(y, \mu_{\hat{y}}) - \Loss{}(y, \mu_y)]}
    }_{\text{SE}(y,\hat{y}):~\text{LF + regressable HF}}
    +
    \underbrace{
    \mathbb{E}_{y, \hat{y}}[\Loss{}(y, \hat{y})-\Loss{}(y, \mu_{\hat{y}})]
    }_{\text{VE}(y,\hat{y}):~\text{non-regressable HF}}
    \}
    ,
\end{aligned}
\label{MLE_full}
\end{equation}
with $\hat{y}$ as an estimator of $y$, and $\mu_{y}=\argmin_\mu \mathbb{E}_{y}[\Loss{} (y,\mu)]$ and $\mu_{\hat{y}}=\argmin_\mu \mathbb{E}_{\hat{y}}[\Loss{}(\hat{y},\mu)]$.
For $\Loss{}_\text{2}$, the two terms above are further simplified as follows:
\begin{equation}
    \begin{aligned}
        \text{SE}(y, \hat{y}) 
        &= \mathbb{E}_{y}[(y-\mu_{\hat{y}})^2-(y-\mu_{y})^2]
        = (\mu_{\hat{y}}-\mu_y)^2 
        \\
        \text{VE}(y, \hat{y}) 
        &= \mathbb{E}_{y, \hat{y}}[(y-\hat{y})^2-(y-\mu_{\hat{y}})^2]
        = \mathbb{E}_{\hat{y}}[(\hat{y}- \mu_{\hat{y}})^2]
        .
    \end{aligned}
\label{SE and VE}
\end{equation}

\noindent\textbf{SE and VE in terms of perceptual SR.}
For the perceptual SR task, SE minimization is desired but VE should be sufficiently preserved. We elaborate on the details below.

VE refers to the additional error introduced by generating components with inherent randomness. In perceptual SR, this is a necessary and inevitable error term induced by fine-grained textures, which cannot be learned via regression. Accordingly, we define the VE term as the \textit{perceptual variance} factor. 
Here, minimizing VE can be understood as further reducing the expected pixel-wise error, apart from reducing SE, at the cost of removing visually important fine textures.
In other words, VE is the factor that leads to the PD trade-off due to its randomness, and VE minimization pushes the prediction space to degenerate, only accepting the average solution as in Fig.\ref{fig:main_figure}.(b), which is blurry~\cite{SISR6_SRGAN}.

Focusing now on SE, this is an \textit{unnecessary error term} that can be reduced without inducing PD trade-off since it does not have randomness. Intuitively, it is the overall degree of alignment between the SR and HR image. 
More specifically, this is the distance between the centroids of the two distributions, where the HR and SR images are each expected to be sampled from. 
Since these centroids are the minimum expected error points of each distribution, they resemble the fidelity-oriented SR counterpart of the perceptual SR images~\cite{eco}, as in Fig.\ref{fig:main_figure}.(a).
Accordingly, we define each factor in the SE term ($\mu_{y}, \mu_{\hat{y}}$) as the \textit{fidelity biases}, and SE itself as the fidelity bias induced error.
Note that fidelity biases are not simply low-frequency (LF) components. 
As often reflected in fidelity-oriented SR methods, specific HF components such as simple object boundaries and edges can be learned via pixel-level regression. Fidelity biases include these regressable high-frequency components.

Overall, since VE is the sole factor inducing the PD trade-off, it is straightforward that we can safely reduce SE without harming perception. By minimizing SE for a given VE, we can obtain the maximum fidelity for a given perception level; the ideal PD trade-off~\footnote{Zero SE alone does not indicate optimal \textit{perception}.}. At the same time, a sufficient level of VE should be preserved for visual quality. 
Accordingly, the following sections will elaborate on designing a novel loss that minimizes SE while preserving VE, taking a step toward the optimal perceptual SR network.

\vspace{5pt}
\noindent\textbf{Revisiting prior methods.}
Observations above share some key concepts with the well-known perception-distortion (PD) trade-off \cite{perception_distortion_tradeoff}. However, we highlight aspects of $\mathcal{L}_\text{pix}$ that are often overlooked in most training methods, with the terms defined above. 
Specifically, previous approaches aim to avoid blurring by either 1) introducing LPF before loss calculation \cite{lowpass1, lowpass2} or 2) simply applying a small coefficient for $\mathcal{L}_\text{pix}$ \cite{SISR6_SRGAN, SISR7_ESRGAN, details_or_artifact}. Below, we will show that both are misinterpreting the blurring phenomenon.

First, LPF-based approaches remove more information than required. They assume that the fidelity biases solely consist low-frequency image components. Yet, bias factors ($\mu_y, \mu_{\hat{y}}$) include certain high-frequency components as discussed above. 
Therefore, despite that LPF-based approaches can avoid texture blurring induced by VE, they also fail to guide regressable high-frequency components that are free from texture blurring. By not providing guidance for certain components of SE, they achieve lower fidelity than necessary, failing to reach the optimal PD trade-off. We provide further analysis for this in Sec.\ref{sec:Analysis}.

Meanwhile, applying a small coefficient to $\mathcal{L}_\text{pix}$ misguidedly treats all aspects of $\mathcal{L}_\text{pix}$ as contributing to blurring. This indiscriminative approach unintentionally weakens SE reduction. When combined with the adversarial loss, this also leads to redundant SE, failing to reach the optimal PD trade-off. This is because adversarial loss works in a task-blind, unsupervised manner: it improves realism but does not consider the alignment between the input image and the network output. As this significantly hinders SE  convergence~\cite{details_or_artifact, desra}, strong guidance on the SE factor is required to prevent high SE. However, the small coefficient greatly reduces this guidance, resulting in unnecessary fidelity loss and a suboptimal PD trade-off. Additionally, since $\mathcal{L}_\text{pix}$ fundamentally enforces VE reduction, it can also never achieve optimal perception despite applying a small scale factor. See Appendix.\ref{supp:pdtradeoff} for discussions and graphical illustrations.

\begin{figure}[t]  % wrapfix box size
    \vspace{-10pt}
    \centering
    \includegraphics[trim=0.5cm 0.6cm 0.5cm 0cm, width=1.0\linewidth]{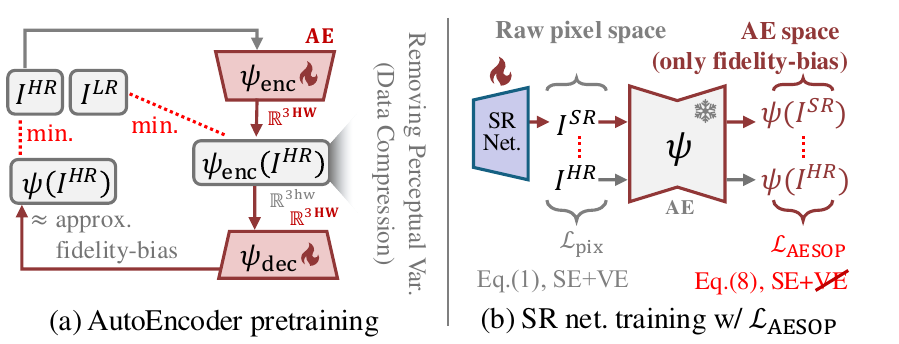}  % actual fig size
    \caption{\textbf{(a)} We pretrain an Auto-Encoder~$\psi_\text{AE}$ that removes perceptual variance factors, thereby establishing a feature space where only the fidelity bias factors reside. \textbf{(b)} The main SR network training step with the proposed $\Loss{}_\text{\AAEESOP{}}$. By applying reconstruction objectives such as the $\mathcal{L}_\text{1}$ loss in the auto-encoded space, we can solely target the fidelity bias induced error without suffering from vanishing perceptual variance (i.e., suffer from blurring). We omit perceptual-quality-oriented losses here. 
    }
    \label{fig:pipeline}
    \vspace{-10pt}
\end{figure}

\section{Method}
\vspace{-5pt}
\noindent\textbf{Motivation.}
An optimal perceptual SR should minimize fidelity bias induced errors while preserving perceptual variance. However, prior methods often fail to achieve this due to their inability to effectively discriminate these factors. This motivates us to design a new method that disentangles them, and solely penalizes based on fidelity biases, bringing us closer to the optimal perceptual SR.

\noindent\textbf{Overview.}
The proposed method can be simplified into two steps. First, we develop an Auto-Encoder~(AE), tailored to create a feature space exclusively for fidelity biases. Second, instead of taking $\Loss{}_\text{p}$ in raw pixel space as typical methods, we calculate $\Loss{}_\text{p}$ in the AE space as in Fig.\ref{fig:pipeline}. Taking $\Loss{}_\text{p}$ in the AE space enables us to provide effective reconstruction guidance with preserved perceptual variance. Note that the term \textit{AE space} indicates the space after the decoder, not the bottleneck. Fundamentally, we are utilizing an AE as a differentiable approximation of an operator ${\psi(\cdot) := \argmin_\mu \mathbb{E}[\Loss{} (\cdot,\mu)]}$ to substitute $\Loss{}_\text{pix}$ with SE.

\vspace{5pt}
\noindent\textbf{Baseline.}
This work aims to make improvements in the GAN-based perceptual SR task. Accordingly, we follow a recent GAN-based SR method LDL~\cite{details_or_artifact}, and set our baseline training objective as below:
\begin{equation}
    \Loss{}_\text{base} = \lambda_1 \Loss{}_\text{pix} + \lambda_2 \Loss{}_\text{percep} + \lambda_3 \Loss{}_\text{adv} + \lambda_4 \Loss{}_\text{artif}
    ,
\end{equation}
where $\Loss{}_\text{pix}$, $\Loss{}_\text{percep}$, $\Loss{}_\text{adv}$, $\Loss{}_\text{artif}$ are the widely used pixel-level $\Loss{}_\text{p}$ loss, perceptual loss~\cite{perceptual_loss_vgg}, the adversarial loss \cite{SISR7_ESRGAN}, and the artifact loss~\cite{details_or_artifact}, respectively, and $\lambda_1$, $\lambda_2$, $\lambda_3$, $\lambda_4$ are coefficients for each loss factors, respectively. We limit the scope of this work to tackling the \textit{fidelity} loss term of \textit{perceptual} SR, thus, we will only modify $\Loss{}_\text{pix}$, while leaving all other loss terms unchanged.

\begin{figure*}[!t]
    \centering
    \vspace{-15pt}
    \includegraphics[width=\linewidth]{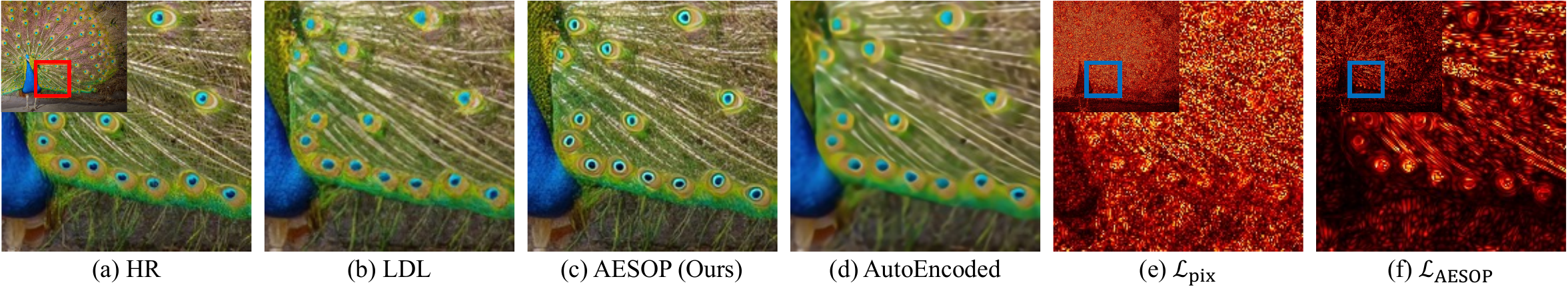}
    \vspace{-25pt}
    \caption{Key components of $\Loss{}_\text{\AAEESOP{}}$ and $\mathcal{L}_\text{pix}$ on SwinIR-backbone. 
    $\mathcal{L}_\text{pix}$ in (e) penalizes perceptually-variance factors, leading to blurry images in (b). In contrast, $\Loss{}_\text{\AAEESOP{}}$ in (f) only penalizes based the fidelity bias (d), which enables us to obtain increased realism as in (c).
    }
    \label{fig:lossmap}
    \vspace{-10pt}
\end{figure*}

\subsection{Auto-Encoder pretraining}
\label{section_ae_pretraining}
\vspace{-5pt}
\textbf{Designing the fidelity bias feature space.}
Our aim is to provide reconstruction supervision focused exclusively on SE. Given that the only components of SE are the fidelity biases ($\mu_y, \mu_{\hat{y}}$), our task simplifies into estimating an operator $\psi$ that estimates the fidelity bias of a given image as ${\psi(\cdot) := \argmin_\mu \mathbb{E}[\Loss{} (\cdot,\mu)]}$. 
To maintain simplicity, we employ a basic Auto-Encoder~(AE) to construct a differentiable approximation of this operator.
This architectural choice is grounded by the nature of SR, where the HR images are conditioned by the LR images and the scale factor. Thus, we model this relationship as $y \sim p(y|x)$ with $x\equiv\phi(y,s)$, where $\phi$ is the $\times s$ downsampling function.
Additionally, the definition of $\psi$ involves minimizing the expected loss over this conditional distribution. Thus, we pretrain the AE with $\Loss{}_{p}$ to learn the forward mapping $x\leftarrow y$, and consecutively, the inverse mapping $y\leftarrow x$. 
Now, the pretrained AE will act as a differentiable approximation of $\psi$, which can decompose the fidelity bias of images and can also be directly plugged into the training framework.

To provide further intuition, we emphasize that the bottleneck of our AE is designed to have the same dimensionality as the LR image. Contrary to most AEs or feature encoders~\cite{perceptual_loss_vgg}, which use a high-dimensional latent space to learn \textit{additional semantics or high-level representations} beyond the raw pixel space, our AE is specifically designed to \textit{remove} particular \textit{low-level} features from the pixel space.
The carefully chosen architecture and pretraining objective form an information bottleneck that effectively compresses out factors that have inherent randomness. Since this is the perceptual variance, we can isolate only the fidelity bias as in Fig.\ref{fig:pipeline}. Thus, $\Loss{}_{p}$ in the AE space resembles loss between fidelity biases, which is fundamentally identical to SE: a term that improves fidelity without inducing blurring. Notably, while vanishing perceptual variance was observed as a critical limitation of $\mathcal{L}_\text{pix}$ in the perceptual SR task, we are conversely taking advantage of it by removing perceptual variance factors in our newly designed fidelity loss term.

\vspace{5pt}
\noindent\textbf{AE pretraining.}
To design a fidelity bias estimator, we pretrain our AE to approximate ${\psi(\cdot) := \argmin_\mu \mathbb{E}[\Loss{} (\cdot,\mu)]}$ for $y \sim p(y|x)$ with $x\equiv\phi(y,s)$. Thus, the AE consecutively estimates the low-resolution counterpart $x$ and then reconstructs $y$.
Accordingly, the pretraining objective is straightforward as follows:
\begin{align}
    \Loss{}^\text{rec}_\text{LR} &= ||\psi_\text{enc}(I^\text{HR}) - I^\text{LR}||_p 
    \label{Eq:LRrecon}
    \\
    \Loss{}^\text{rec}_\text{HR} &=  ||\psi_\text{AE}(I^\text{HR}) - I^\text{HR}||_p
    \label{Eq:HRrecon}
    ,
\end{align}
where $\psi_\text{AE} := \psi_\text{dec} \cdot \psi_\text{enc}$ denotes the AE, $\psi_\text{enc}$, $\psi_\text{dec}$ is the encoder and decoder, and $I^\text{LR}$, $I^\text{HR}$ are LR, HR images. The AE will act as an effective bias estimator, enabling us to design a space where only fidelity biases reside. Note that these losses are only used to pretrain the AE, and will not be used when training the SR network. 

\vspace{5pt}
\noindent\textbf{AE architecture.}
Based on the constructions above, the encoder takes an HR image, and estimates the corresponding LR versions; and the decoder, vice-versa as follows:
\begin{equation}
    \psi_\text{enc} := \mathbb{R}^\text{3HW} \mapsto \mathbb{R}^\text{3hw}
    , \,\,\,\,\,\,\,\,
    \psi_\text{dec} := \mathbb{R}^\text{3hw} \mapsto \mathbb{R}^\text{3HW}
    ,
\end{equation}
where HW and hw each indicate the spatial dimension of the HR and LR images. 
Since the decoding process resembles a fidelity oriented SR task, we employ an off-the-shelf SR architecture RRDBNet \cite{SISR7_ESRGAN}, and initialize the decoder as the pretrained weights for the fidelity-oriented SR task. The encoder is simply a lightweight CNN with downsampling. Refer to the appendix for further details.

\vspace{5pt}
\noindent\textbf{Bottleneck collapse.}
Consider a scenario where the encoder exactly matches the corresponding LR image of the input. If the SR image simply downscales to the original LR image, no loss would backpropagate regardless of the regressable high-frequency component quality of the SR image. Since this can potentially harm the performance, the encoder is jointly optimized with the decoder for Eq.\eqref{Eq:HRrecon}.

\subsection{Auto-Encoded supervision}
\vspace{-5pt}
\noindent\textbf{Defining the \AAEESOP{} loss.}
Since we have obtained a feature space that only retains the fidelity bias factors, we finally define $\Loss{}_\text{\AAEESOP{}}$ as $\mathcal{L}_\text{p}$ with auto-encoded versions of HR and SR images.   Contrary to $\mathcal{L}_\text{pix}$ which minimizes both SE and VE, the proposed $\mathcal{L}_\text{AESOP}$ only minimizes SE is as follows:
\begin{align}
    \Loss{}_\text{pix} &= || I^\text{HR} - I^\text{SR} ||_p 
    ~~~~~~~~~~~~~~~~~~~~~~(= \text{SE} + \text{VE})
    ,
    \label{eq:lpix}
    \\
    \Loss{}_\text{\AAEESOP{}} &= || \psi_\text{AE}(I^\text{HR}) - \psi_\text{AE}(I^\text{SR}) ||_p~~~(\approx \text{SE} + \cancel{\text{VE}})
    ,
    \label{eq:AESOP}
\end{align}
where $I^\text{HR}$, $I^\text{SR}$ represent HR, SR images, respectively. 
Considering the AE pretraining, $\psi_\text{AE}$ is a differential approximation of a fidelity bias estimator. Thus, $\Loss{}_\text{\AAEESOP{}}$ is now fundamentally identical to only penalizing the SE factor of Eq.\eqref{SE and VE} or Eq.\eqref{eq:lpix}.
Since these features are decoupled from the perceptual variance factors by construction, $\Loss{}_\text{\AAEESOP{}}$ leads to increased fidelity without forcing visually important textures to vanish.
Also, note that \textit{auto-encoded} indicates the space after the decoder, not the bottleneck.

\vspace{5pt}
\noindent\textbf{Final objective function.}
Since we focus on improving the \textit{fidelity} loss term of the framework, we substitute $\Loss{}_\text{pix}$ with $\Loss{}_\text{\AAEESOP{}}$, leading to the overall objective function as follows:
\begin{equation}
\label{Eq:final}
    \Loss{}_\text{total} = \lambda_\text{\AAEESOP{}} \Loss{}_\text{\AAEESOP{}} + \lambda_2 \Loss{}_\text{percep} + \lambda_3 \Loss{}_\text{artif} + \lambda_4 \Loss{}_\text{adv}
    ,
\end{equation}
where $\lambda_\text{\AAEESOP{}}$, $\lambda_2$, $\lambda_3$, $\lambda_4$ are coefficients for each loss factors. The overall pipeline of our training strategy is visualized in Fig.\ref{fig:pipeline}.
Based on the constructions above, the proposed $\Loss{}_\text{\AAEESOP{}}$ provides reconstruction guidance without facing conflicts with the perceptual-quality-oriented losses. This indicates that both the $\Loss{}_\text{\AAEESOP{}}$ and perceptual-quality-oriented losses can converge to an optimal point, leading to increased performance in terms of perception-distortion trade-off \cite{perception_distortion_tradeoff}.
Accordingly, while typical methods multiply a very small coefficient to the reconstruction loss (generally chosen as 0.01 \cite{SISR7_ESRGAN, details_or_artifact}) to prevent blurring effects, we let $\lambda_\text{\AAEESOP{}}$ = 1. This way, we can provide significantly stronger reconstruction guidance without suffering from unintended blurring, leading to both lower levels of artifacts~\cite{details_or_artifact} and enhanced realism. For the other coefficients, we follow our baseline~\cite{details_or_artifact} settings and choose $\lambda_2$ = 1, $\lambda_3$ = 1, $\lambda_4$ = 0.005.

\vspace{5pt}
\noindent\textbf{AE collapse.}
Eq.\eqref{eq:\AAEESOP{}} leads to a trivial solution when the AE outputs the same value regardless of the input. To prevent this, we keep the AE frozen when training the SR network.

%% file: article/3_experiments.tex
\input{table/main_table}

%%%%%%%%%%%%%%%%%%%%%%%%%%%%%%%%%%%%%%%%%%%%%%%%%%%%%% Benchmark
\section{Experiments}
\vspace{-5pt}

\subsection{Benchmark evaluation}
\vspace{-5pt}
\textbf{Experimental setup.}
We employ benchmark datasets including Set14~\cite{set14}, Manga109~\cite{manga109}, General100~\cite{general100}, Urban100~\cite{urban100}, DIV2K~\cite{div2k}, BSD100~\cite{bsd100}, LSDIR~\cite{lsdir}. For both the AE pretraining and the SR network training, we use DF2K, a combination of DIV2K~\cite{div2k} and Flickr2K~\cite{flickr2k}. We report PSNR and SSIM~\cite{ssim} scores for distortion metrics and LPIPS~\cite{lpips}, DISTS~\cite{dists} for perceptual quality metrics. 
We use both RRDB-based models: ESRGAN \cite{SISR7_ESRGAN}, SPSR \cite{SPSR}, LDL \cite{details_or_artifact}, CALGAN\cite{calgan}; and SwinIR-based versions of each, if available.
AESOP is used interchangeably to indicate either $\Loss{}_\text{AESOP}$ or the SR networks trained with Eq.\eqref{Eq:final}. Refer to the Appendix for details on experimental settings.

\begin{figure}  % wrapfix box size
    \begin{center}
    \includegraphics[trim=0mm 0mm 8mm 12mm, clip, width=1.0\columnwidth]
    {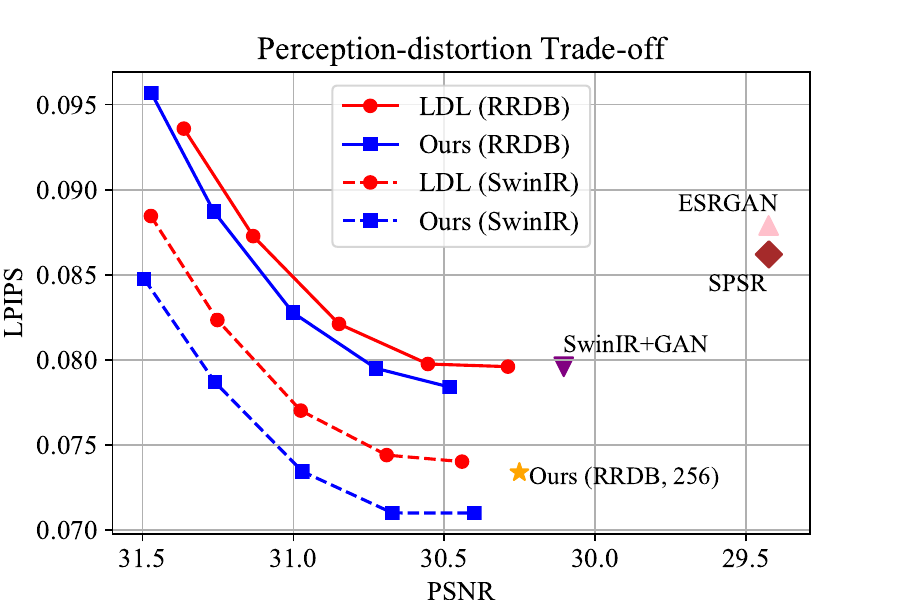} % actual fig size
    \end{center}
    \vspace{-15pt}
    \caption{
    The PD trade-off curve. The backbone and training patch size are indicated; if not specified, the default patch size is 128.
    }
    \label{fig:tradeoff}
    \vspace{-15pt}
\end{figure}

% Quality
\begin{figure*}[t]
    \begin{center}
    \includegraphics[width=1.0\linewidth]{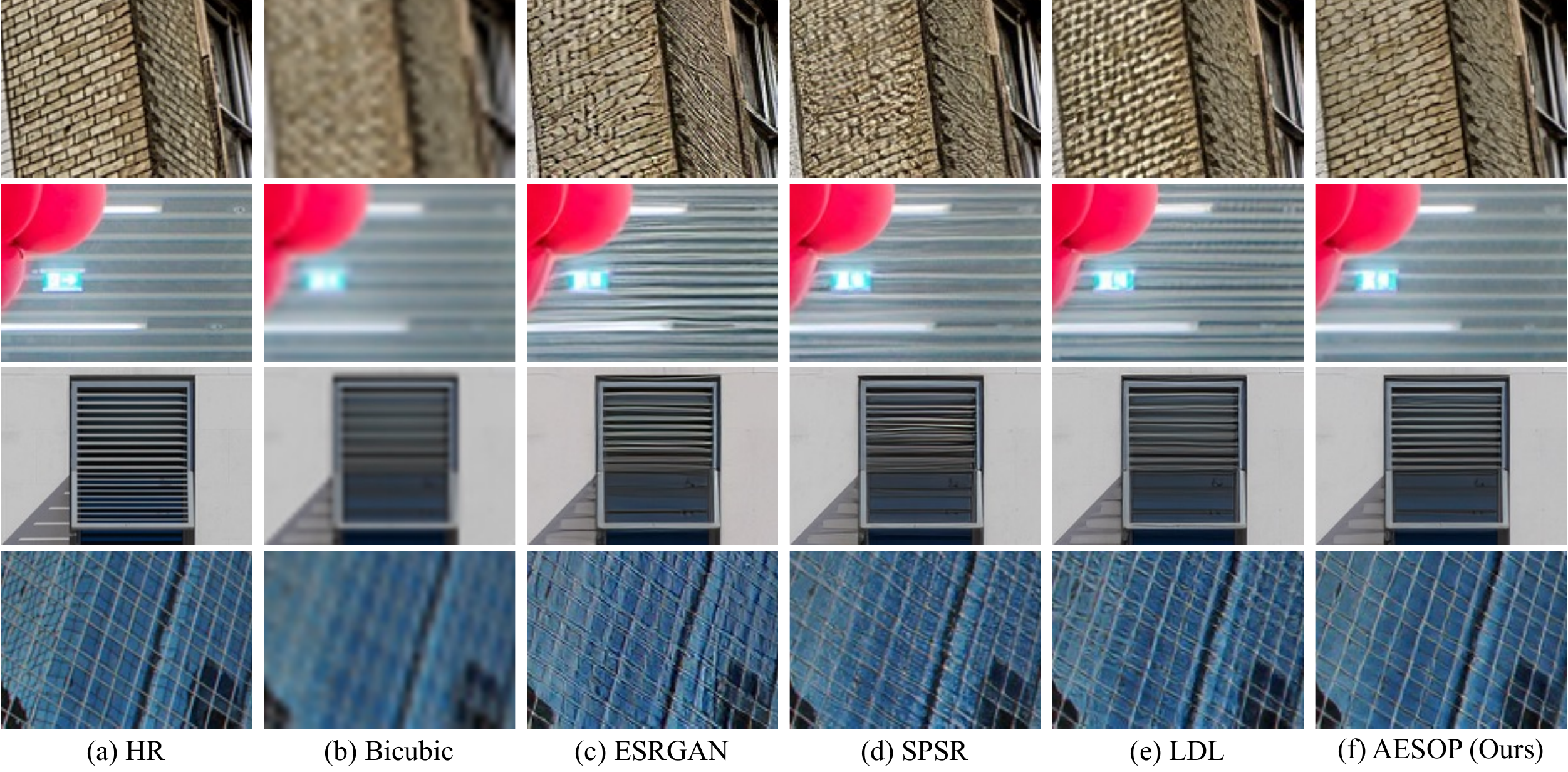}
    \end{center}
    \vspace{-20pt}
    \caption{
        Visual comparison of \AAEESOP{} with baseline methods for the $\times$4 SR task on the RRDB backbone. Our method produces images with fewer visual artifacts. See the Appendix for more visual examples of AESOP's improvement in realism and fine details.
    }
    \label{fig:visual_comparsion_synthetic}
    \vspace{-15pt}
\end{figure*}

\vspace{5pt}
\noindent\textbf{Quantitative comparison.}
In Tab.\ref{tab:maintable}, we perform a quantitative evaluation against baseline GAN-based perceptual SR methods.
%
% About RRDB
Against all RRDB \cite{SISR7_ESRGAN} based baseline methods, \AAEESOP{} significantly improves both distortion metrics and perceptual scores. We find the key factor of this improvement as the carefully designed feature space and pretrained AE. It provides an increased level of reconstruction guidance, specifically to fidelity bias factor, while providing additional freedom to perceptual-quality-oriented losses. 
%
%
%
% About SwinIR v2
For SwinIR~\cite{SwinIR} backbone methods, we report both the final results at 300K iterations and intermediate training results at 200K iterations (indicated as AESOP$^*$). At 200K, it can be seen that AESOP leads to improvements in both fidelity scores and perceptual metrics, similar to the RRDB-backbone. When we fully train our model up to 300K, we observe further enhancements in perception scores. Since improved perception leads to lower fidelity due to the PD trade-off, we provide the PD trade-off curves in Fig.\ref{fig:tradeoff}, with PSNR and LPIPS scores on the General100 dataset. AESOP leads to improved trade-off relationships on both RRDB and SwinIR backbones. 
Additionally, we provide RRDB-backbone results using a training HR patch size of 256 (denoted as \AAEESOP{}$^\dagger$) to align with the training settings of SwinIR.
\AAEESOP{} also improves realism for real-world SR tasks as in Tab.\ref{tab:main_tiny_realworldsr_nriqa}. Refer to Appendix.\ref{supp:realworldsr} for more results.

\vspace{5pt}
\noindent\textbf{Qualitative comparison.}
In Fig.\ref{fig:visual_comparsion_synthetic}, baseline methods often suffer from unpleasant GAN artifacts~\cite{details_or_artifact} while \AAEESOP{} presents a significantly lower level of these artifacts.
This is due to the strong reconstruction guidance $\Loss{}_\text{AESOP}$ provides, since it does not require a small scaling factor.
Additionally, refer to the Appendix for visual examples of AESOP improving realism on fine details and complex textures.

\vspace{5pt}
\noindent\textbf{Ablation studies.}
% \vspace{-5pt}
To verify the effects of each component, we perform ablation studies on \AAEESOP{}~(128). In Tab.\ref{tab:ablation_table}, we report DISTS and PSNR scores for each setting on DIV2K validation set. The first line indicates the baseline setting that matches LDL, where none of our proposed components are applied. 
%
% a 세팅 
Tab.\ref{tab:ablation_table}.(a) indicates employing a decoder, but using a simple bicubic downsampling operation instead of the encoder. As reported, the perceptual quality and fidelity are improved even when solely utilizing the decoder due to its ability in offering stronger reconstruction loss without conflict with perceptual objectives. However, the usage of bicubic downsampling with a decoder corresponds to the bottleneck collapse, where no loss backpropagates if the SR image downsamples to the LR image, thus, leading to slightly lower performance against our full method.
In Tab.\ref{tab:ablation_table}.(b), we further introduce a learnable encoder together with the decoder, but without Eq.\eqref{Eq:LRrecon} (i.e., without estimating the LR image). 
Given that both fidelity bias factors and perceptual variance factors are determined by the LR counterpart, the model cannot properly estimate the fidelity bias and the perceptual variance.
Consequently, the perceptual variance factor can be penalized, which is not intended, leading to the lowest perceptual score.
Our full model in Tab.\ref{tab:ablation_table}.(c) further employs Eq.\eqref{Eq:LRrecon}. Thus, it better models the LR than Tab.\ref{tab:ablation_table}.(b), while also avoiding the bottleneck collapse which Tab.\ref{tab:ablation_table}.(a) has suffered, leading to the best scores in terms of perceptual quality.

\input{table/realworld_nriqa_tiny}

\begin{table}[t]
    \centering
    \footnotesize
    \setlength{\tabcolsep}{3pt}
    \renewcommand{\arraystretch}{0.8} % Adjust row spacing
    \begin{tabular}{l|ccc|cc}
    \toprule
          & Decoder & Encoder & $\Loss{}^\text{rec}_\text{LR}$ & DISTS~$\downarrow$ & PSNR~$\uparrow$ \\
    \midrule
        Baseline (LDL) & & & & 0.0526 & 28.819 \\
    \midrule
        Config-(a) & \checkmark    &               &               & 0.0521 & 29.060 \\
        Config-(b) & \checkmark    & \checkmark    &               & 0.0526 & \textbf{29.150} \\
        Config-(c) (Ours) & \checkmark    & \checkmark    & \checkmark    & \textbf{0.0518} & 29.079 \\
    \hline
    \end{tabular}
    \vspace{-5pt}
    \caption{Ablation study on each component of AESOP~(128).}
    \vspace{-10pt}
    \label{tab:ablation_table}
\end{table}

\subsection{Analysis}
\label{sec:Analysis}
\begin{figure}[t]
    \begin{center}
    \includegraphics[width=1\linewidth]{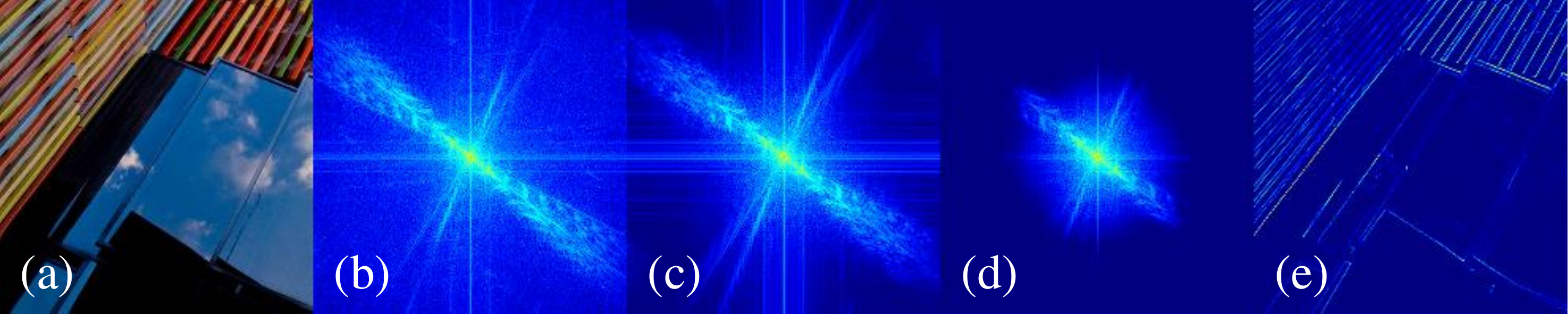}
    \end{center}
    \vspace{-15pt}
    \caption{
        Visual comparison between AE and LPF. \textbf{(a)} Original image. \textbf{(b)} Original image in spectral domain. \textbf{(c)} Forwarding through AE. \textbf{(d)} Applying LPF. \textbf{(e)} Absolute difference between (c) and (d). (Electronic viewer highly recommended.)
    }
    \vspace{0pt}
    \label{fig:visual_comparsion_of_AE_LP}
\end{figure}

\vspace{-5pt}
\noindent\textbf{Spectral analysis.}
Several prior works \cite{lowpass1, lowpass2} employ low-pass filtering (LPF) to avoid the blurring effect of $\Loss{}_\text{pix}$.
To identify the difference between our AE and LPF, we analyze the spectral magnitudes of Fig.\ref{fig:visual_comparsion_of_AE_LP}.(a) after applying AE and LPF, in Fig.\ref{fig:visual_comparsion_of_AE_LP}.(c)-(d), respectively.
As can be seen, LPF blindly removes all HF components, contrary to AE, where certain patterns in the HF regions are preserved. As discussed in Sec.~\ref{Section:RevisitingL1}, the remaining HF components in Fig.\ref{fig:visual_comparsion_of_AE_LP}.(c) are factors that can be learned by pixel-level regression, and the removed HF components are the non-regressable factors (VE) that lead to blurring when minimized. 

Here, we focus on the \textit{remaining} HF components and visualize it in Fig.\ref{fig:visual_comparsion_of_AE_LP}.(e), by taking the absolute difference between Fig.\ref{fig:visual_comparsion_of_AE_LP}.(c)-(d). Simple object edges are highlighted, which are regions where even fidelity-oriented SR networks (trained with pixel-level regression loss, $\Loss{}_\text{pix}$), can also sharply reconstruct. This indicates that specific HF components \textit{can} be regressed, and importantly, these HF components \textit{cannot} be disentangled from other HF components by band-pass filters (e.g., LPF) or frequency selection~\cite{mimounet, sfnet}; since they are intertwined within the same frequency band. 

Meanwhile, LPF is expected to provide limited supervision in these regressable HF components, leading to degraded performance. To validate this statement, we compare AESOP against LPF in Tab.\ref{tab:ablation2_table}.(b). We apply LPF on SR and HR images before calculating $\Loss{}_\text{pix}$ on top of our baseline method LDL. A notable degradation in performance can be observed, highlighting the superiority of our AE-based method in maintaining high-quality reconstruction guidance over conventional frequency-based methods.

Overall, we conclude that our AE can successfully disentangle fidelity biases and perceptual variances, by capturing specific HF components that 
1) cannot be obtained by simple frequency selection,
2) but can be learned via regression, and 
3) significantly contribute to improved fidelity.

\vspace{5pt}
\noindent\textbf{Loss map comparison.}
Fig.\ref{fig:lossmap} visualizes key components of $\Loss{}_\text{\AAEESOP{}}$ and $\mathcal{L}_\text{pix}$. In Fig.\ref{fig:lossmap}$.(e), \mathcal{L}_\text{pix}$ cannot distinguish perceptual variance factors and fidelity bias factors. Thus, visually important fine-textures are penalized, leading to a blurry result as in Fig.\ref{fig:lossmap}.(b). Meanwhile, $\Loss{}_\text{\AAEESOP{}}$ in Fig.\ref{fig:lossmap}.(f) successfully extracts and penalizes only the fidelity bias factor Fig.\ref{fig:lossmap}.(d), leading to increased realism as in Fig.\ref{fig:lossmap}.(c).

\vspace{5pt}
\noindent\textbf{Fidelity bias estimation.}
Since $\Loss{}_\text{AESOP}$ does not lead to blurring, we do not multiply a small scaling factor. Accordingly, we can provide significantly stronger guidance on fidelity biases. 
Here, we measure how well each network estimates the fidelity biases, apart from PSNR scores which are influenced by perceptual variances.
To do this, we introduce \textit{AE-PSNR}, which measures the PSNR between auto-encoded SR and HR images. This score reflects how well an image captures the fidelity bias of the reference image.
However, since AESOP is trained using the AE, there may be unintended biases introduced by the AE itself. Thus, we additionally report LR-PSNR as an unbiased metric independent of the AE, which measures the PSNR between downscaled SR and the original LR images.
This captures how well the fidelity biases align, but without being influenced by the AE. However, note that this measure only reflects the 
LF feature, a subcomponent of the fidelity bias.
In Tab.\ref{tab:lrpsnr_aepsnr_table_swinir}, AESOP shows improvements on both AE-PSNR and LR-PSNR scores, demonstrating the superiority of $\Loss{}_\text{AESOP}$ against scaled $\Loss{}_\text{pix}$ in effectively reducing the SE term. See the Appendix for scores on RRDB-backbone methods.

\vspace{5pt}
\noindent\textbf{Architectural choice of AE.}
Performance of \AAEESOP{} with SRResNet-based \cite{SISR6_SRGAN} AE is reported in Tab.\ref{tab:ablation2_table}.(a). We observe a slight performance drop against our RRDB-based AE, but it is still superior against LDL. 
While this indicates that AESOP relies on a well-performing AE, this reliance does not pose a practical issue.
In our pretraining framework, we initialize the decoder of AE as the fidelity-oriented SR network, which is expected to be already in place under the SRGAN-based training framework.

\input{table/lraepsnr_swin_table}

\begin{table}[]
    \centering
    \footnotesize
    \renewcommand{\arraystretch}{0.7} % Adjust row spacing
    \begin{tabular}{c|c|c|c|c}
            \midrule
            \multicolumn{1}{c}{} & \multicolumn{1}{|c}{Manga109} & \multicolumn{1}{|c}{General100} & \multicolumn{1}{|c}{Urban100} & \multicolumn{1}{|c}{DIV2K100} \\ 
            \hline\midrule
            $\text{Ours}$ & \textbf{29.97}/\textbf{.0525} & \textbf{30.48}/\textbf{.0784} & \textbf{25.63}/.1064 & \textbf{29.08}/\textbf{.0977}  \\
            \midrule
            $\text{(a)}$  & 29.78/.0534 & 30.35/.0789 & 25.55/\textbf{.1054} & 28.97/.0982 \\ % SRResNet
            $\text{LDL}$  & 29.62/.0544 & 30.29/.0796 & 25.46/.1084 & 28.82/.0999  \\
            $\text{(b)}$  & 29.55/.0545 & 30.20/.0801 & 25.39/.1090 & 28.75/.1005  \\ % LP
            \hline
    \end{tabular}
    \vspace{-5pt}
    \caption{ \small PSNR/LPIPS scores. \textbf{(a)} \AAEESOP{} with SRResNet-based AE~($\psi$). \textbf{(b)} Applying LPF before calculating $\Loss{}_\text{pix}$ with LDL~\cite{details_or_artifact}.}
    \vspace{-5pt}
    \label{tab:ablation2_table}
\end{table}

%% file: table/main_table.tex
\begin{table*}[!t]
\vspace{-5pt}
\centering
\footnotesize
\renewcommand{\arraystretch}{0.7} % Adjust row spacing
\resizebox{\textwidth}{!}{%
\begin{tabular}{@{}cl|cccc|c|ccc|c@{}}
\midrule
 & Backbone & \multicolumn{5}{c|}{RRDB} & \multicolumn{4}{c}{SwinIR} \\ 
\hline\hline
\multicolumn{1}{c}{Metrics} &
  \multicolumn{1}{l|}{Benchmark} &
  ESRGAN &
  SPSR &
  LDL &
  \multicolumn{1}{c|}{AESOP} &
  \multicolumn{1}{c|}{AESOP$^\dagger$} &
  \multicolumn{1}{c}{+GAN} &
  \multicolumn{1}{c}{LDL} &
  \multicolumn{1}{c|}{AESOP*} &
  \multicolumn{1}{c}{AESOP} \\ 
\hline
\multicolumn{2}{l|}{\makecell{Recon. Objective}} & $\Loss{}_\text{pix}$ & $\Loss{}_\text{pix}$ & $\Loss{}_\text{pix}$ & \multicolumn{1}{c|}{\textbf{$\Loss{}_\text{\AAEESOP{}}$}} & \multicolumn{1}{c|}{\textbf{$\Loss{}_\text{\AAEESOP{}}$}} & $\Loss{}_\text{pix}$ & $\Loss{}_\text{pix}$ & \multicolumn{1}{c|}{\textbf{$\Loss{}_\text{\AAEESOP{}}$}} & \multicolumn{1}{c}{\textbf{$\Loss{}_\text{\AAEESOP{}}$}}\\
\hline
\multicolumn{2}{l|}{\makecell{Patch Size (Training)}}         & 128          & 128          & 128                    & 128                          & 256                          & 256             & 256                    & 256                      & 256                  \\  \hline\midrule 
\multirow{7}{*}{LPIPS~$\downarrow$} 
                       & Set14                                & 0.1241       & 0.1207       & 0.1132                 & \textbf{0.1067}              & 0.1053                       & 0.1160          & 0.1091                 & \textbf{0.1023}          & 0.1027               \\
                       & Manga109                             & 0.0649       & 0.0672       & 0.0544                 & \textbf{0.0525}              & 0.0494                       & 0.0542          & 0.0469                 & \textbf{0.0440}          & 0.0461               \\
                       & General100                           & 0.0879       & 0.0862       & 0.0796                 & \textbf{0.0784}              & 0.0734                       & 0.0796          & 0.0740                 & \textbf{0.0717}          & 0.0710               \\
                       & Urban100                             & 0.1229       & 0.1184       & 0.1084                 & \textbf{0.1064}              & 0.1033                       & 0.1077          & 0.1021                 & \textbf{0.0961}          & 0.0945               \\
                       & DIV2K-val                            & 0.1154       & 0.1099       & 0.0999                 & \textbf{0.0977}              & 0.0936                       & 0.1038          & 0.0944                 & \textbf{0.0909}          & 0.0893               \\
                       & BSD100                               & 0.1616       & 0.1609       & 0.1535                 & \textbf{0.1515}              & 0.1443                       & -               & 0.1572                 & \textbf{0.1441}          & 0.1385               \\
                       & LSDIR                                & 0.1378       & 0.1312       & 0.1180                 & \textbf{0.1152}              & 0.1123                       & -               & 0.1132                 & \textbf{0.1094}          & 0.1071               \\  \midrule
\multirow{7}{*}{DISTS~$\downarrow$} 
                       & Set14                                & 0.0951       & 0.0920       & 0.0866                 & \textbf{0.0852}              & 0.0825                       & 0.0930          & 0.0869                 & \textbf{0.0809}          & 0.0819               \\  
                       & Manga109                             & 0.0471       & 0.0463       & \textbf{0.0355}        & 0.0360                       & 0.0356                       & 0.0365          & \textbf{0.0315}        & 0.0327                   & 0.0328               \\
                       & General100                           & 0.0874       & 0.0884       & 0.0801                 & \textbf{0.0798}              & 0.0773                       & 0.0835          & 0.0794                 & \textbf{0.0768}          & 0.0762               \\
                       & Urban100                             & 0.0880       & 0.0849       & \textbf{0.0793}        & \textbf{0.0793}              & 0.0768                       & 0.0835          & 0.0800                 & \textbf{0.0751}          & 0.0742               \\
                       & DIV2K-val                            & 0.0593       & 0.0546       & 0.0526                 & \textbf{0.0518}              & 0.0484                       & 0.0531          & 0.0507                 & \textbf{0.0469}          & 0.0459               \\
                       & BSD100                               & 0.1165       & 0.1176       & 0.1163                 & \textbf{0.1117}              & 0.1089                       & -               & 0.1185                 & \textbf{0.1078}          & 0.1072               \\
                       & LSDIR                                & 0.0764       & 0.0699       & 0.0650                 & \textbf{0.0641}              & 0.0612                       & -               & 0.0650                 & \textbf{0.0601}          & 0.0591               \\  \midrule
\multirow{7}{*}{PSNR~$\uparrow$}  
                       & Set14                                & 26.594       & 26.860       & 27.228                 & \textbf{27.361}              & 27.246                       & 27.282          & 27.526                 & \textbf{27.822}          & 27.421               \\  
                       & Manga109                             & 28.413       & 28.561       & 29.620                 & \textbf{29.973}              & 29.747                       & 29.345          & 30.143                 & \textbf{30.453}          & 30.061               \\
                       & General100                           & 29.425       & 29.424       & 30.289                 & \textbf{30.482}              & 30.251                       & 30.104          & 30.441                 & \textbf{30.752}          & 30.401               \\
                       & Urban100                             & 24.365       & 24.804       & 25.459                 & \textbf{25.630}              & 25.541                       & 25.736          & 26.231                 & \textbf{26.398}          & 26.148               \\
                       & DIV2K-val                            & 28.175       & 28.182       & 28.819                 & \textbf{29.079}              & 28.910                       & 28.784          & 29.117                 & \textbf{29.543}          & 29.137               \\
                       & BSD100                               & 25.313       & 25.501       & 25.954                 & \textbf{26.080}              & 25.904                       & -               & 26.216                 & \textbf{26.405}          & 25.930               \\
                       & LSDIR                                & 23.882       & 24.232       & 24.663                 & \textbf{24.933}              & 24.845                       & -               & 25.129                 & \textbf{25.419}          & 25.038               \\  \midrule
\multirow{7}{*}{SSIM~$\uparrow$}  
                       & Set14                                & 0.7144       & 0.7254       & 0.7358                 & \textbf{0.7402}              & 0.7371                       & 0.7407          & 0.7478                 & \textbf{0.7578}          & 0.7438               \\  
                       & Manga109                             & 0.8595       & 0.8590       & 0.8734                 & \textbf{0.8827}              & 0.8802                       & 0.8796          & 0.8880                 & \textbf{0.8949}          & 0.8880               \\
                       & General100                           & 0.8095       & 0.8091       & 0.8280                 & \textbf{0.8335}              & 0.8269                       & 0.8305          & 0.8347                 & \textbf{0.8415}          & 0.8328               \\
                       & Urban100                             & 0.7341       & 0.7474       & 0.7661                 & \textbf{0.7724}              & 0.7697                       & 0.7786          & 0.7918                 & \textbf{0.7947}          & 0.7884               \\
                       & DIV2K-val                            & 0.7759       & 0.7720       & 0.7897                 & \textbf{0.7978}              & 0.7951                       & 0.7911          & 0.8011                 & \textbf{0.8121}          & 0.8023               \\
                       & BSD100                               & 0.6527       & 0.6596       & 0.6813                 & \textbf{0.6841}              & 0.6783                       & -               & 0.6923                 & \textbf{0.6982}          & 0.6813               \\
                       & LSDIR                                & 0.6866       & 0.6966       & 0.7117                 & \textbf{0.7220}              & 0.7202                       & -               & 0.7316                 & \textbf{0.7397}          & 0.7289               \\
                       \midrule\hline
\end{tabular}%
}
\vspace{-10pt}
\caption{
Quantitative comparison between \AAEESOP{}~(Ours) and baseline methods. The best results of each group are highlighted in \textbf{bold}. \AAEESOP{}$^*$ indicates only training 200K iterations, \AAEESOP{}$^\dagger$ indicates training with a larger patch.
}
\vspace{-10pt}
\label{tab:maintable}
\end{table*}

%% file: table/realworld_nriqa_tiny.tex
\begin{table}[t]
\centering
\setlength{\tabcolsep}{3pt}
\footnotesize
\renewcommand{\arraystretch}{0.8} % Adjust row spacing
{%
\begin{tabular}{l|l|c|cc}
\hline
\multicolumn{1}{c|}{Dataset} & \multicolumn{1}{c|}{Method} & \multicolumn{1}{c|}{Recon. Obj.} & \multicolumn{1}{c}{NIQE$\downarrow$} & \multicolumn{1}{c}{MANIQA$\uparrow$} \\ \hline\midrule
\multirow{2}{*}{\centering RealSRv3 \cite{RealSR_dataset1_RealSRv3}} 
                          & Real-ESRGAN & $\Loss{}_\text{pix}$                  & 4.6790 &	0.3662  \\
                          & AESOP~(Ours)  & $\Loss{}_\text{AESOP}$ & \textbf{4.2337} &	\textbf{0.4136}  \\ \midrule
\multirow{2}{*}{\centering DRealSR \cite{RealSR_datasets3_DRealSR}}  
                          & Real-ESRGAN   & $\Loss{}_\text{pix}$                & 4.7152 &	0.3404  \\
                          & AESOP~(Ours)  & $\Loss{}_\text{AESOP}$                                & \textbf{4.1922} &	\textbf{0.3917}  \\ \midrule
                          
\end{tabular}%
}
\vspace{-10pt}
\caption{Quantitative results of AESOP in real-world settings. Refer to the Appendix for further results, including visual examples.
}
\label{tab:main_tiny_realworldsr_nriqa}
\end{table}

%% file: table/lraepsnr_swin_table.tex
\begin{table}[t]
\vspace{-5pt}
\centering
\setlength{\tabcolsep}{2pt}
\footnotesize
\renewcommand{\arraystretch}{0.9} % Adjust row spacing
\begin{tabular}{@{}c|c|ccccccc@{}}
\midrule
\multirow{2}{*}{\rotatebox{90}{}} &  Method & Set14 & Mg109 & Gen100 & Urb100 & DIV2K & B100 & LSDIR \\
\hline\midrule
\multirow{2}{*}{\rotatebox{90}{AE-}} 
& LDL~\cite{details_or_artifact}    & 31.525	& 33.215 &	33.994 &	29.374 &	32.855 &	29.792 &	29.071 \\
& Ours & \textbf{32.111} & \textbf{33.635} & \textbf{34.535} & \textbf{29.666} & \textbf{33.490} & \textbf{30.366} & \textbf{29.552} \\
\midrule
\multirow{2}{*}{\rotatebox{90}{LR-}} 
& LDL~\cite{details_or_artifact}    & 46.899 &	49.135 &	48.663 &	47.404 &	48.084 &	45.494 &	45.731 \\
& Ours & \textbf{48.245} & \textbf{50.042} & \textbf{49.733} & \textbf{48.564} & \textbf{49.856} & \textbf{47.578} & \textbf{47.476} \\
\hline
\end{tabular}
\vspace{-5pt}
\caption{
AE-PSNR and LR-PSNR scores with SwinIR-backbone.
}
\label{tab:lrpsnr_aepsnr_table_swinir}
\end{table}

%% file: article/5_conclusion.tex
\section{Conclusion}
\vspace{-5pt}
This work analyzes limitations of $\mathcal{L}_\text{pix}$ (i.e., the conventional pixel-level $\mathcal{L}_\text{p}$) in the context of perceptual SR.
Further, we highlight the shortcomings of prior circumvention to avoid blurring, in terms of fidelity biases and perceptual variance factors.
We tackle this issue by introducing $\Loss{}_\text{\AAEESOP{}}$, a novel reconstruction loss that separates fidelity bias factors from perceptual variance factors using an AE, pretrained for a reconstruction task. This allows us to focus on enhancing fidelity while preserving the visual quality of SR images.
Experimental results validate that the proposed method leads to significant improvement in the perceptual SR task.

%% file: article/X_suppl.tex
\clearpage
\appendix
\setcounter{page}{1}

% \maketitlesupplementary
\twocolumn[{
\begin{center}
    \Large\textbf{Auto-Encoded Supervision for Perceptual Image Super-Resolution}\\
    \vspace{0.5em}Supplementary Material \\
    \vspace{1.0em}
    \vspace{10pt}
    \captionsetup{type=figure}
    \includegraphics[width=0.9\textwidth]{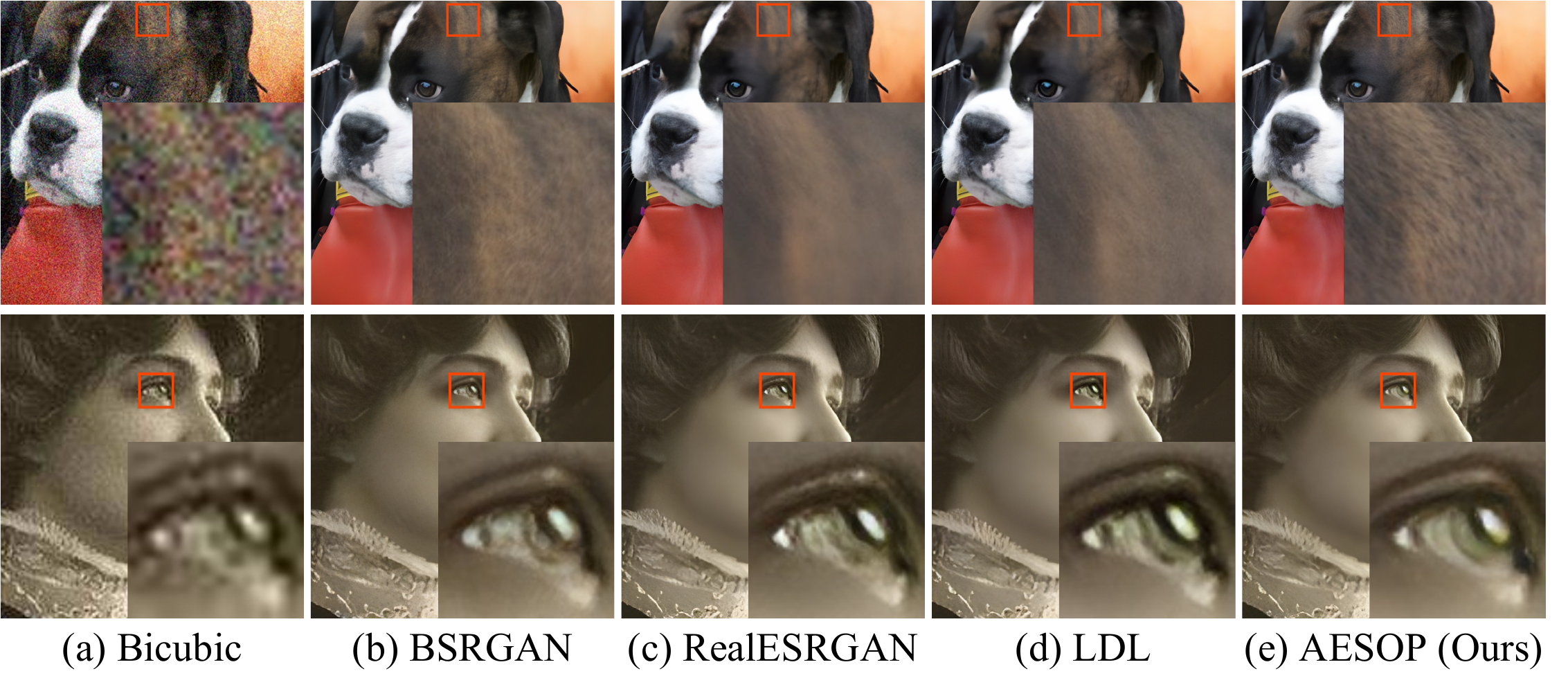}
    \vspace{-0.2cm}
    \captionof{figure}{
    Visual comparison between \AAEESOP{}~(ours) against baseline methods for the real-world $\times$4 SR task. \AAEESOP{} leads to improved realism (top) with a lower level of visual artifacts (bottom). \textbf{Zoom in for best view.}
    }
    \label{fig:visual_comparsion_realworld}
    \vspace{0.2cm}
    \end{center}
}]

\input{supp/realworld_nriqa}

% \vspace{20pt}
% \begin{figure}[h]  % [!t] forces the figure to appear at the top
%     \centering
%     \includegraphics[width=0.9\textwidth]{figures/realworld_comp.pdf}
%     \vspace{-10pt}
%     \caption{
%         Visual comparison between \AAEESOP{}~(ours) against baseline methods for the real-world $\times$4 SR task. \AAEESOP{} leads to improved realism (top) with a lower level of visual artifacts (bottom). \textbf{Zoom in for best view.}
%     }
%     \label{fig:visual_comparsion_realworld}
% \end{figure}

%%%%%%%%%%%%%%%%%%%%%%%%%%%%%%%%%%
%%%%%%%%%%%%%%%%%%%%%%%%%%%%%%%%%% Misc
%%%%%%%%%%%%%%%%%%%%%%%%%%%%%%%%%% 

\section{Implementation and experimental details}
\vspace{-5pt}

\paragraph{Network architecture and weight initialization.}
Following previous works, we initialize our SR networks with the official weight of the fidelity-oriented model of either ESRGAN~\cite{SISR7_ESRGAN} or SwinIR~\cite{SwinIR}. Similarly, the decoder of the AE follows the architecture of RRDB and is initialized with the fidelity-oriented weights. The overall architecture of the encoder is implemented in a straightforward manner. We simply design it as a series of two convolutional layers (fromRGB layer), followed by a pixel-unshuffle operation and two  RRDB blocks \cite{SISR7_ESRGAN}, concluding with additional two convolutional layers (toRGB layer).
The RRDB block is identical to that of the SR networks.
The pixel-unshuffle acts as a $\times$s downscaling operation, effectively reducing the image dimension to match that of the LR image.
Since the channel size is increased due to the pixel-shuffle operation, the second layer of the fromRGB layer reduces the channel size $\times \text{s}^2$ smaller than that of the RRDB block.
The kernel size is 3$\times$3 for all convolutional layers.

\vspace{5pt}
\noindent\textbf{Training and evaluation details.}
The optimizer is chosen as the Adam~\cite{adam} optimizer with a learning rate of 0.0001, for both the Auto-Encoder and the SR network. Following conventions, we choose p = 1 for $\Loss{}_\text{p}$ and the coefficient of loss factors are $\lambda_\text{\AAEESOP{}}$ = 1, $\lambda_2$ = 1, $\lambda_3$ = 1, $\lambda_4$ = 0.005. The Auto-Encoder is pretrained up to 100K iterations, and the SR networks are trained up to 300K iterations. Unless specified, the HR training patch size is 128. PSNR and SSIM scores are evaluated on the Y channel (luminance channel) in the YCbCr space and pixels up to the scale factors in the border were ignored. We use the default \texttt{alex} option for LPIPS~\cite{lpips}. Training and evaluation are performed on top of BasicSR~\cite{basicsr}. Networks are trained and evaluated with either 4 NVIDIA A6000s or 4 NVIDIA RTX 3090s.

\section{Evaluation on real-world SR datasets}
\label{supp:realworldsr}
\vspace{-5pt}

% %%%%%%%%%%%%%%%%%%%%%%%%%%%%%%%%%%
% %%%%%%%%%%%%%%%%%%%%%%%%%%%%%%%%%% Real world sr
% %%%%%%%%%%%%%%%%%%%%%%%%%%%%%%%%%% 

\noindent\textbf{AESOP on real-world SR.}
In the real-world SR task, the overall task becomes more complex and the range of plausible solutions is larger than that of the conventional bicubic SR task. Accordingly, the conflict between $\mathcal{L}_\text{pix}$ and perceptual quality-oriented objectives gets severe, and the blurring tendency of conventional $\mathcal{L}_\text{pix}$ loss may become more significant. We further validate the effectiveness of the proposed method in the real-world $\times$4 SR task. For comparison, we use representative baseline real-world SR methods utilizing $\mathcal{L}_\text{pix}$, including RealESRGAN \cite{RealESRGAN}, BSRGAN \cite{BSRGAN}, and LDL \cite{details_or_artifact}.

\vspace{5pt}
\noindent\textbf{Qualitative results.}
In Fig.\ref{fig:visual_comparsion_realworld}, we provide a visual comparison of \AAEESOP{} against baseline methods for the $\times$4 real-world SISR task on RealSRSet \cite{BSRGAN}. 
We only replace the $\mathcal{L}_\text{pix}$ term of \cite{details_or_artifact} while keeping all other training settings identical. Since we do not have ground-truth HR images, we only provide bicubic upsampled images and SR results from each method.
Due to the inherent high complexity of the real-world task, baseline networks fail in generating fine-grained textures (first row of Fig.\ref{fig:visual_comparsion_realworld}) and generate visually unpleasing artifacts (second row of Fig.\ref{fig:visual_comparsion_realworld}). In contrast, \AAEESOP{} successfully recovers fine textures with fewer artifacts.

\vspace{5pt}
\noindent\textbf{Quantitative results.}
We report quantitative results on RealSRv3~\cite{RealSR_dataset1_RealSRv3} and DRealSR~\cite{RealSR_datasets3_DRealSR}. To assess perceptual quality, we utilize NIQE~\cite{niqe}, MANIQA~\cite{yang2022maniqa}, MUSIQ~\cite{ke2021musiq}, and CLIP-IQA~\cite{wang2023exploring_clipiqa} scores. Due to memory constraints, images were divided into four quadrants when evaluating the CLIP-IQA scores for the DRealSR dataset. \AAEESOP{} demonstrates superior performance against baselines in all evaluation metrics, which verifies the effectiveness of our method for practical applications.

% %%%%%%%%%%%%%%%%%%%%%%%%%%%%%%%%%%
% %%%%%%%%%%%%%%%%%%%%%%%%%%%%%%%%%% More results on bicubic SR
% %%%%%%%%%%%%%%%%%%%%%%%%%%%%%%%%%% 

% \clearpage

\input{supp/fid_table}

\section{Additional results for the Bicubic SR task}
\vspace{-5pt}

\vspace{5pt}
\paragraph{FID scores.}
In Tab.\ref{tab:fid_table}, we report  Frechet Inception Distance~(FID)~\cite{fid} scores to further evaluate the proposed AESOP against baseline methods for the bicubic $\times$4 SR task. FID, widely used for generative tasks \cite{karras2019style}, has recently been adopted for super-resolution tasks~\cite{details_or_artifact, calgan}. However, its standard approach of resizing images to 299$\times$299 may not be suitable to assess SR methods.
Resizing can alter important details that SR aims to improve, directly conflicting with the objectives of SR focusing on enhancing image quality at higher resolutions.

\vspace{5pt}
\noindent\textbf{Patch FID scores.}
Accordingly, we additionally report the patch-FID~(pFID)~\cite{chai2022any_pfid} scores, which does not require image resizing. For patch-FID evaluation, 299$\times$299 non-overlapping patches are extracted from the images. If an image is smaller than 299 pixels in any dimension, we use zero-padding to meet the required size.

\vspace{5pt}
\noindent\textbf{Fidelity bias estimation.}
As discussed in the main article, we do not multiply a small scaling factor to $\Loss{}_\text{AESOP}$ which leads to significantly stronger guidance on fidelity biases (Fig.\ref{fig:supp_lossscalegraph}). 
Accordingly, we have measured how well AESOP and the baseline methods estimate the fidelity biases by reporting \text{AE-PSNR} which captures the distance between the fidelity bias of the SR image and the fidelity bias of the HR image. Additionally, we have shown \text{LR-PSNR} scores to provide a metric that is not biased by the Auto-Encoder.
In Tab.\ref{tab:lrpsnr_aepsnr_table_rrdb}, we additionally provide \text{AE-PSNR} and \text{LR-PSNR} scores on top of the RRDB~\cite{SISR7_ESRGAN} backbone. Similar to results in Tab.\ref{tab:lrpsnr_aepsnr_table_swinir}, AESOP shows improvements in both AE-PSNR and LR-PSNR scores, highlighting the superiority of AESOP in effectively reducing the SE term.

\vspace{5pt}
\noindent\textbf{AESOP on recent backbone network architecture.}
We report additional quantitative results on the benchmark datasets in Tab.\ref{tab:supp_maintable_supp}.
First, we employ DRCT~\cite{drct}, a recent state-of-the-art Swin Transformer-based method that leverages dense residual connections within a fidelity-oriented SR framework. We implement LDL on top of DRCT and compare it to our proposed AESOP. AESOP consistently outperforms the baseline in terms of both fidelity and perceptual quality, demonstrating its effectiveness even with advanced network architectures. Notably, the performance improvement is more significant compared to the RRDB backbone, suggesting that AESOP may yield even greater benefits with larger-capacity network architectures.

\vspace{5pt}
\noindent\textbf{Regarding recent perceptual-oriented losses.}
We report quantitative results of another recent state-of-the-art method, CALGAN~\cite{calgan}. This work is a different branches of research in the field of perceptual SR, focusing on improvements in perceptual quality-oriented losses. Interestingly, AESOP outperforms CALGAN in most cases, even without the Mixture of Experts~(MoE)–based discriminator proposed in CALGAN~\cite{calgan}. This signifies the effectiveness of AESOP.
However, note that improvements in network architectures and perceptual-oriented losses are beyond the scope of this work. The focus of this study is on the fidelity loss term $\Loss{}_\text{pix}$ within the perceptual SR framework.
We leave the integration of $\Loss{}_\text{AESOP}$ (fundamentally a \textit{fidelity} loss), with the enhanced perceptual-oriented losses of CALGAN to future work due to limited computational budget.

\input{supp/lraepsnr_rrdb_table}

\input{supp/supp_maintable_supp}

% %%%%%%%%%%%%%%%%%%%%%%%%%%%%%%%%%%
% %%%%%%%%%%%%%%%%%%%%%%%%%%%%%%%%%% More visualizaion
% %%%%%%%%%%%%%%%%%%%%%%%%%%%%%%%%%% 

% %%%%%%%%%%%%%%%%%%%%%%%%%%%%%%%%%%
% %%%%%%%%%%%%%%%%%%%%%%%%%%%%%%%%%% Discussion
% %%%%%%%%%%%%%%%%%%%%%%%%%%%%%%%%%% 

\vspace{15pt}
\section{Further discussion on AESOP}
\label{supp:discussion}
\vspace{-5pt}

\begin{figure*}
    \centering
    \includegraphics[width=\linewidth]{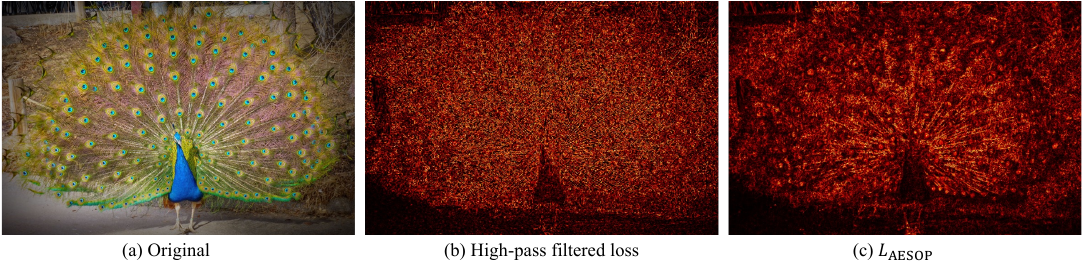}
    \vspace{-20pt}
    \caption{Loss map comparison between high-pass filtered (HPF) loss and $\Loss{}_\text{AESOP}$. Refer to Appendix.\ref{supp:discussion} for further discussion.}
    \label{fig:supp_HPF}
\end{figure*}

\noindent\textbf{Regarding the loss maps and spectral magnitudes.}
Here we provide further discussions regarding the loss maps and the spectral analysis in the main article. In Sec.\ref{sec:Analysis}, we have discussed the differences between AESOP and \textit{low}-pass filtering-based methods.
However, the loss maps reveal object edges, which are the regressable high-frequency components, aligning to \textit{high}-pass filters. Accordingly, we provide further discussion and compare AESOP against high-pass filter based losses or similarly against edge filters from two perspectives: 1) regions with low loss values under $\Loss{}_\text{AESOP}$ and 2) regions with high loss values under HPF losses. (Fig.\ref{fig:supp_HPF})

First, we emphasize that regions with low loss values under $\Loss{}_\text{AESOP}$ do not imply that $\Loss{}_\text{AESOP}$ neglects these areas. Instead, they simply indicate that the network has accurately estimated the fidelity bias in those regions. This is clearly different from frequency filters, which entirely ignore these regions. For instance, consider a scenario where the SR network produces low-frequency artifacts due to adversarial training instability. In such cases, $\Loss{}_\text{AESOP}$ effectively guides the network toward proper estimation, whereas HPF loss ignores these artifacts, resulting in suboptimal performance. This also suggests that the components that require reconstruction guidance and those that do not require reconstruction guidance are inherently intertwined within each pixel. Thus, they cannot be disentangled merely by selecting which pixels to penalize.

Meanwhile, for regions that receive high loss activations under high-pass filtering (HPF) loss, these typically correspond to areas with fine textures. This is exactly the problematic issue raised in $\Loss{}_\text{pix}$, where such activations contribute to blurring. Consequently, this represents an undesirable aspect of HPF-based methods.

\begin{figure}[!t]
    \centering
    \includegraphics[width=1.0\columnwidth]{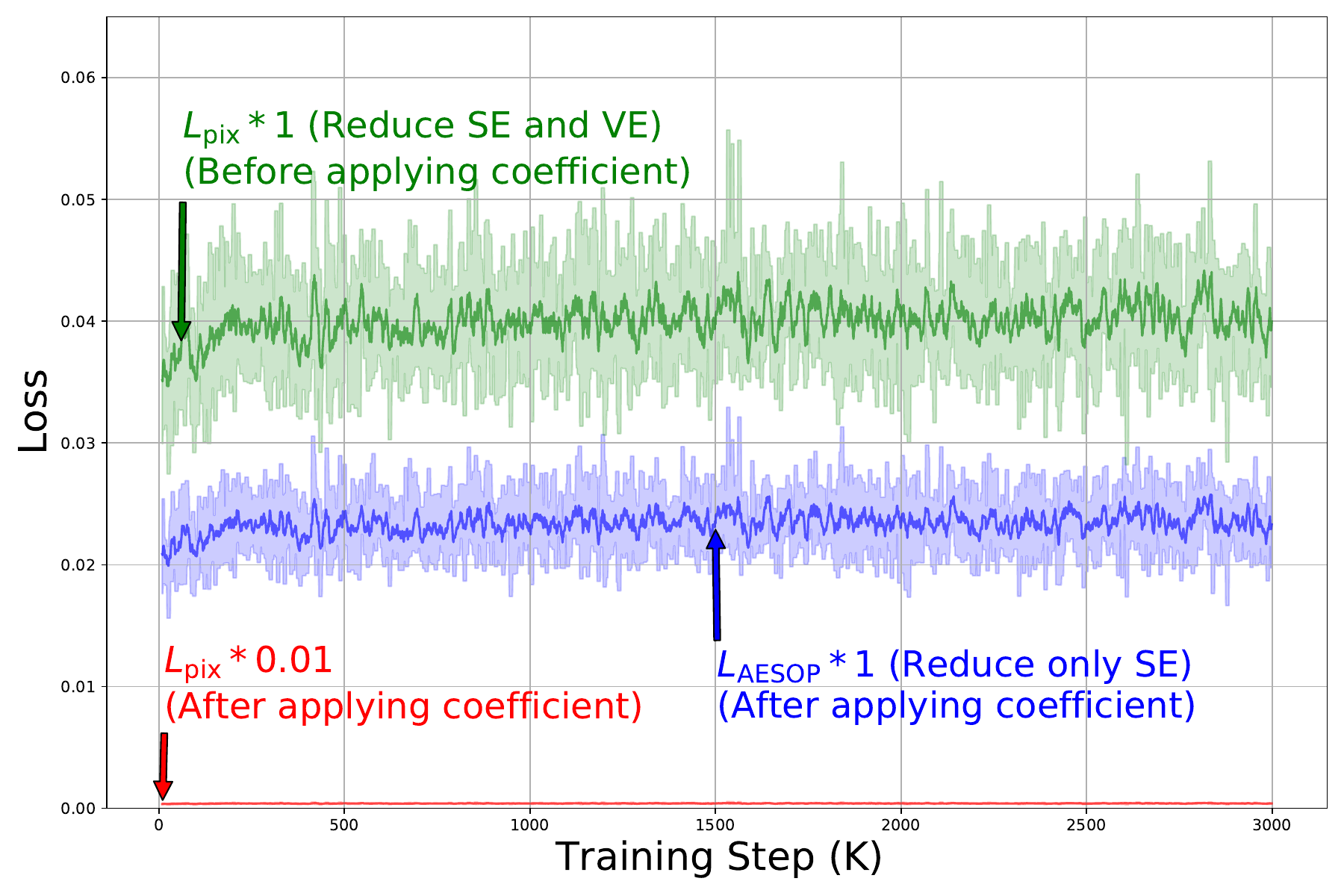}
    \vspace{-10pt}
    \caption{The loss trajectory of $\Loss{}_\text{pix}$ before applying the coefficient (green) is visualized by scaling the original loss (red). The loss trajectory of $\Loss{}_\text{AESOP}$ (blue) is visualized as-is, since we do not scale it. Refer to Appendix.\ref{supp:discussion} for further discussion.
    }
    \label{fig:supp_lossscalegraph}
\end{figure}

\begin{figure}[t]
    \centering
    \includegraphics[width=0.98\columnwidth]{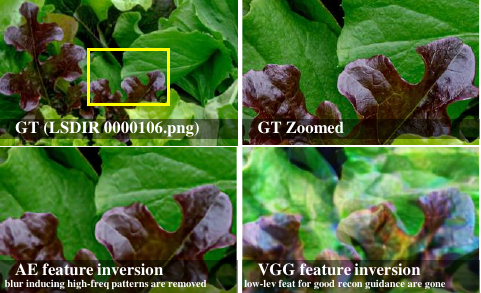}
    % \vspace{-10pt}
    \caption{Comparison between feature inversion results obtained from deep features extracted by VGG and our proposed AE. Deep features of VGG lose important low-level features crucial for a reconstruction loss. Meanwhile, the carefully chosen network architecture and the pretraining objective of our AE enable precise control over which information to remove (blur-inducing high-frequency patterns) and preserve (structural edges). 
    }
    \label{fig:feature_inversion}
\end{figure}

\vspace{5pt}
\noindent\textbf{Intuitions on $\Loss{}_\text{AESOP}$ based on loss scales.}
In Fig.\ref{fig:supp_lossscalegraph}, we compare the loss scales of $\Loss{}_\text{pix}$ and $\Loss{}_\text{AESOP}$, both before and after applying their loss coefficients. Before the loss coefficients are applied, $\Loss{}_\text{pix}$ (green) exhibits greater loss values than $\Loss{}_\text{AESOP}$ (blue). This observation aligns with our theoretical analysis and construction of the Auto-Encoder, where $\Loss{}_\text{AESOP}$ only penalizes a subcomponent of $\Loss{}_\text{pix}$. Specifically, while $\Loss{}_\text{pix}$ minimizes both perceptual variance (VE) and fidelity bias induced error (SE), our carefully designed $\Loss{}_\text{AESOP}$ only targets the SE term, leading to lower loss values. Consequently, the gap between the green loss trajectory and the blue one quantifies the VE loss component embedded within $\Loss{}_\text{pix}$.
After the loss coefficients are applied to each reconstruction loss, $\Loss{}_\text{AESOP}$~(blue) provides an order of magnitude stronger reconstruction guidance compared to scaled $\Loss{}_\text{pix}$~(red). Regardless of this strengthened fidelity guidance, SR networks trained with $\Loss{}_\text{AESOP}$ do not have to suffer from blurring and can achieve improved perceptual quality over $\Loss{}_\text{pix}$.

\begin{figure*}[!t]
    \centering
    \includegraphics[width=\linewidth]{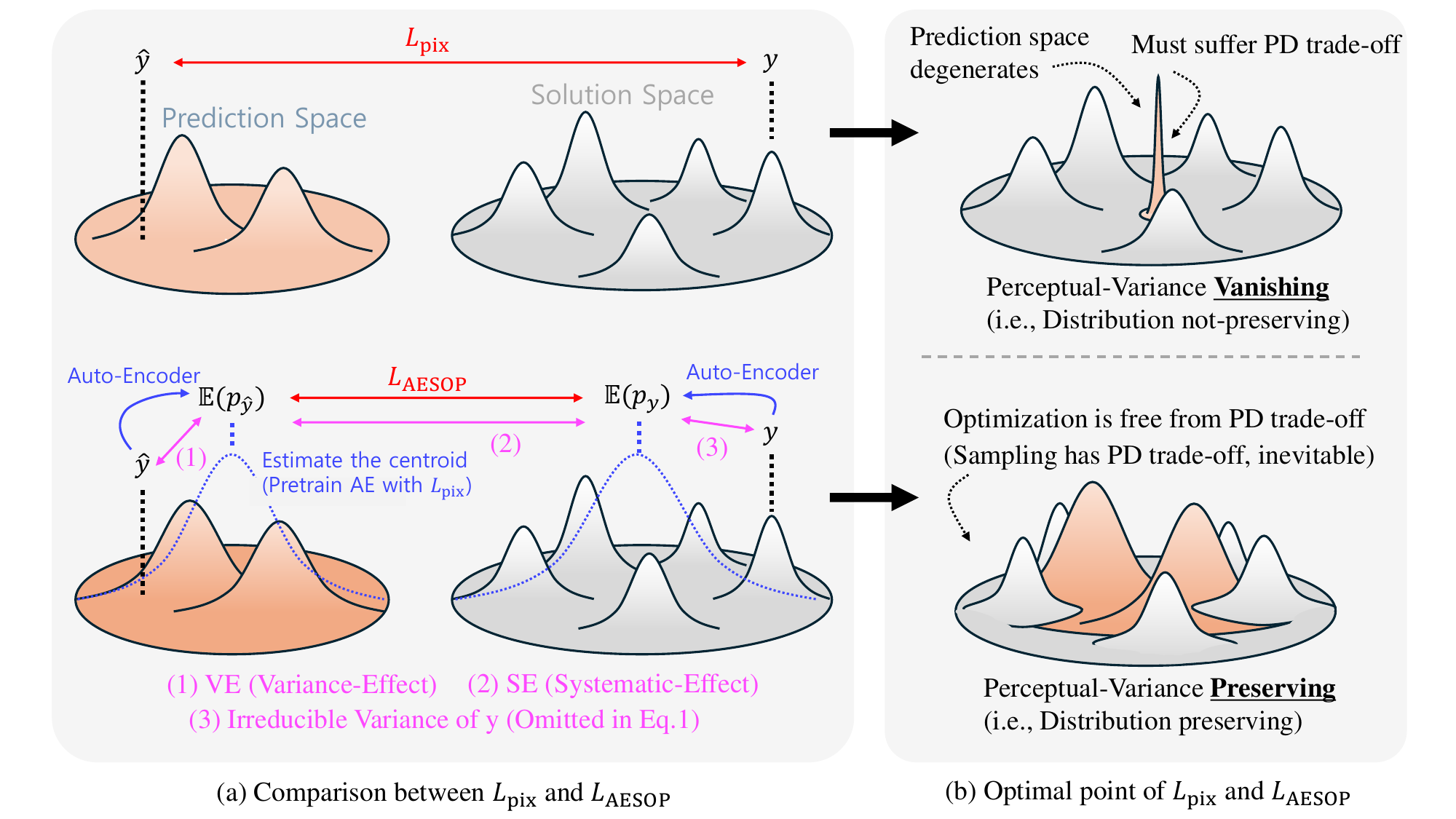}
    \vspace{-10pt}
    \caption{Graphical illustration of the optimization procedure and the optimal point for $\Loss{}_\text{pix}$ and $\Loss{}_\text{AESOP}$.}
    \label{fig:supp_aesopintuition}
\end{figure*}

\vspace{5pt}
\noindent\textbf{Intuitions on $\Loss{}_\text{pix}$ and $\Loss{}_\text{AESOP}$.}
Apart from Fig.\ref{fig:main_figure}, we show additional graphical illustration in Fig.\ref{fig:supp_aesopintuition} to provide further intuitions on the overall optimization procedure and the optimal point of each $\Loss{}_\text{pix}$ and $\Loss{}_\text{AESOP}$. As can be seen, $\Loss{}_\text{AESOP}$ consecutively estimates the centroid (fidelity bias) of the prediction and solution space, and minimizes the distance between them (i.e., minimizes the SE factor). Accordingly, $\Loss{}_\text{AESOP}$ reaches the optimal point when the two distributions are aligned. However, $\Loss{}_\text{pix}$ converges to the minimum expected error point, which is the blurry average solution. Thus, the prediction space degenerates.

\vspace{5pt}
\noindent\textbf{Comparison between $\Loss{}_\text{percep}$ and $\Loss{}_\text{AESOP}$.}
The proposed loss $\mathcal{L}_\text{AESOP}$ and the perceptual loss $\mathcal{L}_\text{percep}$ share the characteristic of utilizing a pretrained neural network for guidance. However, they differ fundamentally in their objectives and mechanisms. Below, we clarify these differences in two different aspects.

First, the primary objectives of these two losses differ significantly.$\mathcal{L}_\text{percep}$ belongs to the category of perceptual-oriented losses. Its main purpose is to explicitly improve perceptual quality by providing \textit{high-level} semantics and textural guidance. Accordingly, $\mathcal{L}_\text{percep}$ measures the discrepancy between the SR and HR images within a high-dimensional feature space derived from a pretrained feature extractor (such as VGG~\cite{vggnet}), where the high-dimensional space captures additional semantic and textural details beyond those available in the raw pixel domain.
In contrast, $\mathcal{L}_\text{AESOP}$ is fundamentally a reconstruction (fidelity) loss that provides guidance based on low-level features, similar to the conventional $\Loss{}_\text{pix}$, but specifically tailored for perceptual SR tasks so that it does not show conflicts with perceptual-oriented losses. $\mathcal{L}_\text{AESOP}$ employs an Auto-Encoder (AE) architecture with a low-dimensional bottleneck, pretrained for low-level reconstruction. Due to its design and pretraining objective, the AE inherently compresses the input and selectively discards certain information, while preserving important low-level features. Consequently, the Auto-Encoded output contains \textit{less} information compared to the original image, as opposed to the enriched, high-dimensional features used in $\mathcal{L}_\text{percep}$.

Second, the underlying mechanism and the information each feature encoder embeds are different.
In order to utilize a feature encoder as a loss function in low-level vision tasks, precise control over which information to remove and preserve is important. Considering that a reconstruction loss in perceptual SR task should (1) provide sufficient reconstruction guidance while (2) avoid blurring; feature encoders should be able to preserve important low-level features while removing blur-inducing factors.
However, feature encoders pretrained on image classification tasks (such as VGG) naturally discard many low-level features not relevant to classification, resulting in uncontrollable loss of critical reconstruction information. In contrast, the carefully designed AE preserves essential low-level features, particularly structural edges, while the blur inducing perceptual variance factors are removed.

We empirically verify these properties through feature inversion results shown in Fig.\ref{fig:feature_inversion}. Clearly, deep features extracted from VGG omit critical low-level reconstruction details. On the other hand, our AE-derived deep features successfully retain sharp edges and structural alignment while the blur inducing high-frequency textural information is removed as intended. 
Overall, this verifies that $\mathcal{L}_\text{percep}$ cannot act as a standalone reconstruction loss in low-level vision tasks, while $\mathcal{L}_\text{AESOP}$ can, and is even shown to outperform conventional $\mathcal{L}_\text{pix}$ through extensive experiments.

\paragraph{Disclaimer.} We clarify that the improvement in perceptual scores by raising $\Loss{}_\text{AESOP}$ is since it does not hinder the perceptual-oriented guidance provided by perceptual-oriented losses under the SRGAN-framework. $\Loss{}_\text{AESOP}$ itself will not guide towards realism. We keep improvements in perceptual-oriented losses out of the scope of this work.

\section{Further intuition regarding the PD trade-off}
\label{supp:pdtradeoff}
\vspace{-5pt}

\begin{figure*}
    \centering
    \includegraphics[trim=40mm 15mm 10mm 15mm, clip, width=0.95\linewidth]{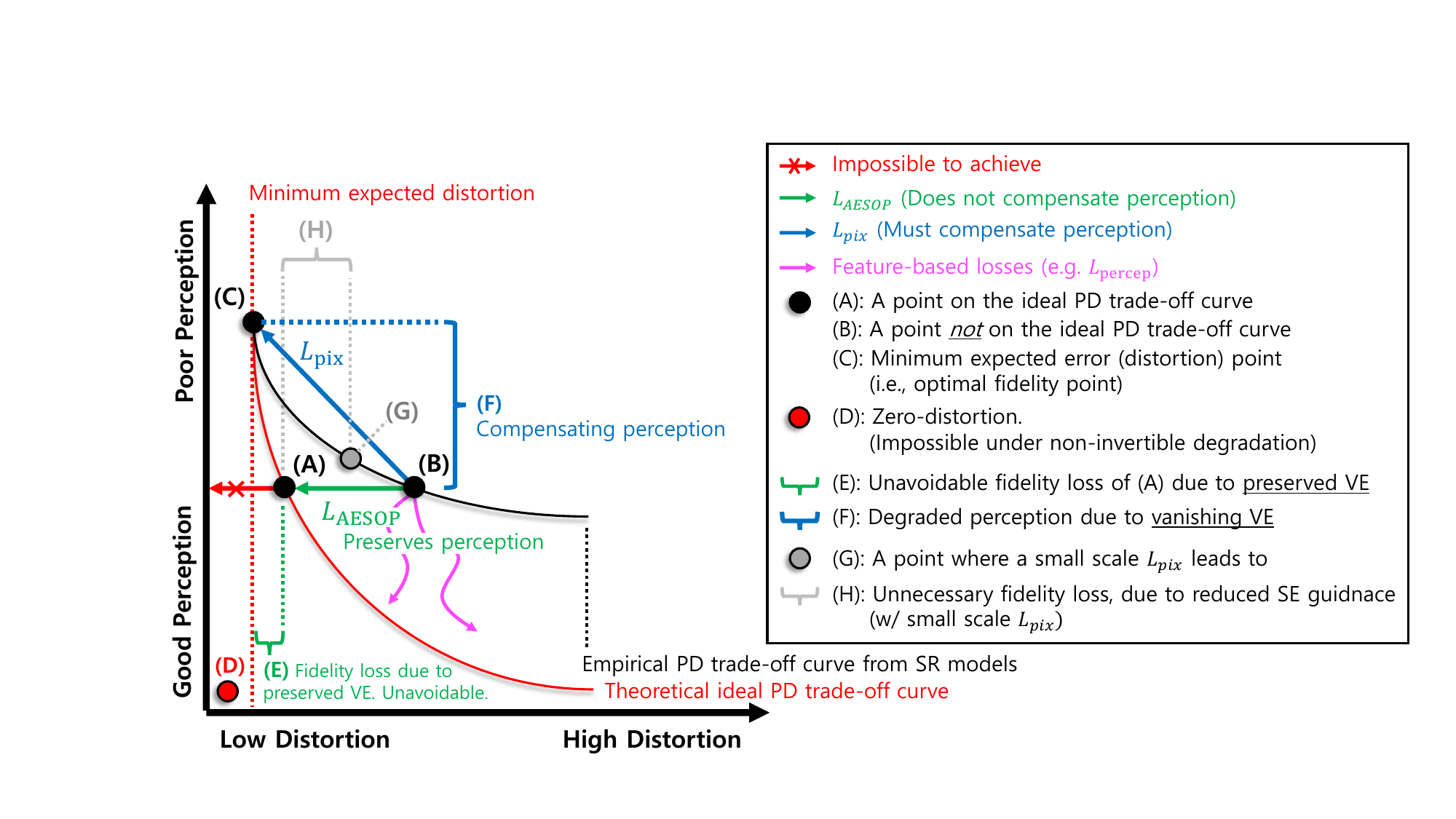}
    \caption{Graphical illustration of $\Loss{}_\text{pix}$ and $\Loss{}_\text{AESOP}$ in terms of the perception-distortion trade-off.}
    \vspace{-10pt}
    \label{fig:supp_aesop_pdcurvelecture}
\end{figure*}

\noindent\textbf{Comparison between $\Loss{}_\text{pix}$ and $\Loss{}_\text{AESOP}$.}
Fig.\ref{fig:supp_aesop_pdcurvelecture} represents the guidance $\Loss{}_\text{pix}$ and $\Loss{}_\text{AESOP}$ provides in terms the perception-distortion (PD) trade-off. We start our discussion with point (B), which represents an image that is not optimal in both fidelity and perception. Given this image, $\Loss{}_\text{pix}$ with a large coefficient guides the image towards point (C). This is the blurry image with the lowest expected distortion, or simply the fidelity bias of the image. Meanwhile, with a smaller coefficient, it achieves improved perception as point (G). However, it leads to unnecessary fidelity loss (H) since SE reduction is significantly weakened while the adversarial loss continuously hinders SE convergence. Meanwhile, $\Loss{}_\text{AESOP}$ removes the VE minimization term of $\Loss{}_\text{pix}$. Thus, it improves fidelity without suffering from blurring, thereby guides point (B) towards point (A). However, we clarify that $\Loss{}_\text{AESOP}$ cannot further improve the fidelity beyond the ideal PD trade-off curve. This is impossible as (E), under non-invertible degradation~\cite{perception_distortion_tradeoff} including image super-resolution. This statement even holds for the case with an optimal perceptual SR network that can sample images from the true posterior. Note that $\Loss{}_\text{AESOP}$ reaches zero for point (A).

\vspace{5pt}
\noindent\textbf{Is $\Loss{}_\text{AESOP}$ a distortion measure?}
Blau et. al.~\cite{perception_distortion_tradeoff} have shown that we must compensate perception when aiming to reduce \textit{any} distortion measure; the perception-distortion trade-off. 
This might seem contradictory with $\Loss{}_\text{AESOP}$ at first glance, since $\Loss{}_\text{AESOP}$ is designed to improve fidelity without degrading perception. However, fortunately, $\Loss{}_\text{AESOP}$ does not fall within the definition of distortion metric defined in Blau et. al.~\cite{perception_distortion_tradeoff}. A distortion measure $\Delta$ that induces PD trade-off requires: $\Delta(y_1, y_2)>0$ for $y_1 \neq y_2$ by definition.
However, for AESOP, it is straightforward (and also intended) that multiple different images can share an identical fidelity bias. Formally, there exists $y_1, y_2$ s.t. $\Loss{}_\text{AESOP}(y_1, y_2)=0$ and $y_1 \neq y_2$. As this does not satisfy the constraints of a distortion measure, $\Loss{}_\text{AESOP}$ is not guaranteed to raise PD trade-off. However, we clarify that this does not imply that SR networks trained with $\Loss{}_\text{AESOP}$ can generate images that are free from the PD trade-off. This is impossible even with an oracle perceptual SR network, as discussed in prior sections.

\section{Additional visualization}
\label{supp:additional_loss_and_spectral_vis}

\begin{figure*}
    \centering
    \includegraphics[width=\linewidth]{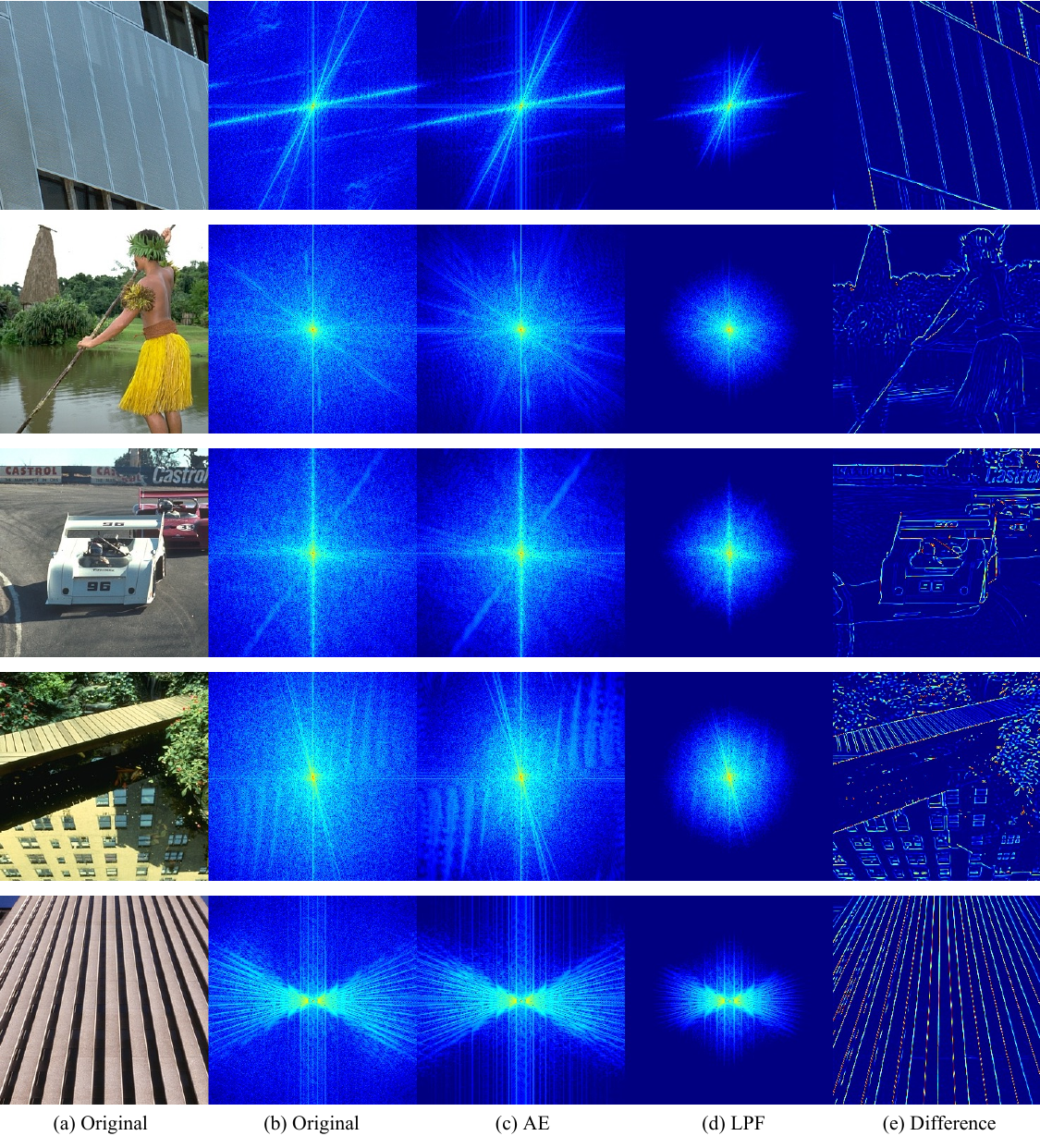}
    \caption{Visual comparison between Auto-Encoding and low-pass filtering. \textbf{(a)} Original image. \textbf{(b)} Original image in spectral domain. \textbf{(c)} Auto-Encoded image. \textbf{(d)} Low-pass filtered image. \textbf{(e)} Absolute difference between the Auto-Encoded image and the low-pass filtered image. \textbf{Electronic viewer recommended}.}
    \label{fig:supp_additional_spectral_1}
\end{figure*}

\noindent\textbf{Spectral magnitudes.}
Fig.\ref{fig:supp_additional_spectral_1} provides visual examples of the spectral magnitudes, aligning with Fig.\ref{fig:visual_comparsion_of_AE_LP}.
The spectral magnitudes reflect the effectiveness of the pretrained Auto-Encoder in discriminating non-regressable factors that lead to blurring and the regressable high-frequency components that enhance fidelity without causing blurring. Meanwhile, low-pass filters fail to achieve this since the regressable and non-regressable components cannot be disentangled using simple frequency filters. They are intertwined within the same frequency band.

\begin{figure*}
    \centering
    \includegraphics[width=\linewidth]{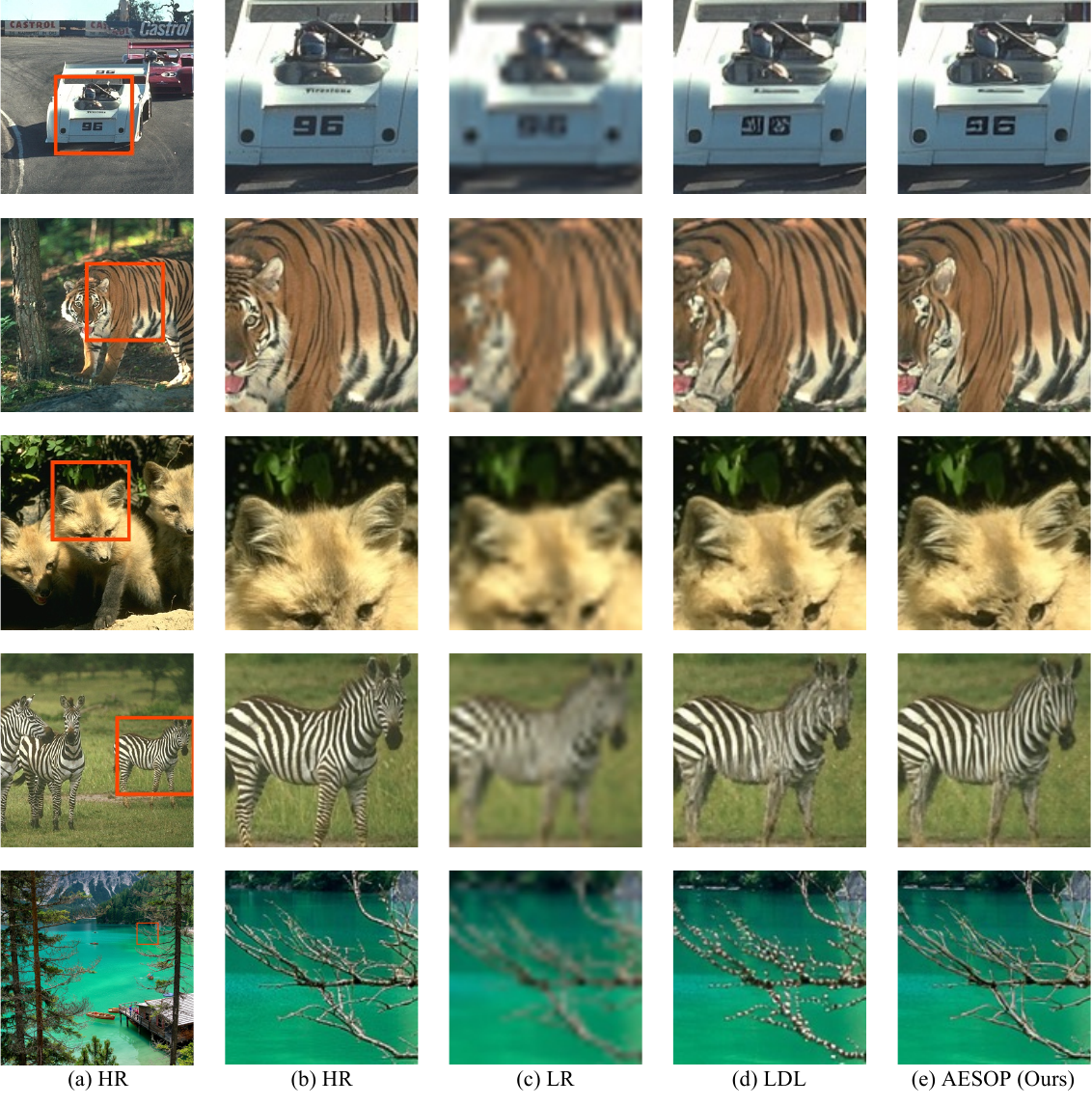}
    \caption{Visualization of AESOP (ours) and the baseline method for the bicubic ×4 SR task with SwinIR backbone. AESOP can generate fine-grained textures with a lower level of visual artifacts. \textbf{Zoom in for best view}.}
    \label{fig:supp_swinir1}
\end{figure*}

\begin{figure*}
    \centering
    \includegraphics[width=\linewidth]{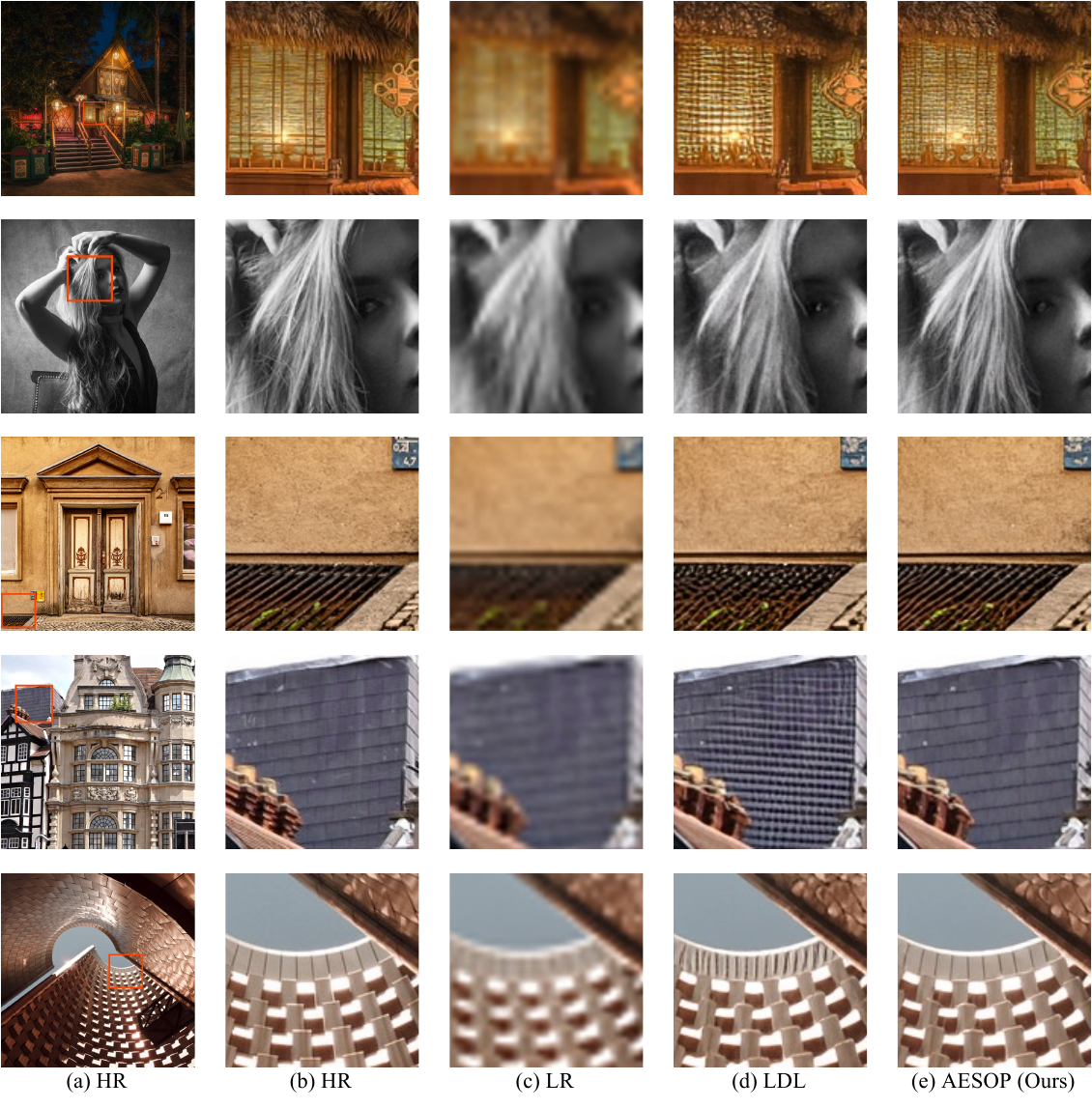}
    \caption{Visualization of AESOP (ours) and the baseline method for the bicubic ×4 SR task with SwinIR backbone. AESOP can generate fine-grained textures with a lower level of visual artifacts. \textbf{Zoom in for best view}.}
    \label{fig:supp_swinir2}
\end{figure*}

\begin{figure*}
    \centering
    \includegraphics[width=\linewidth]{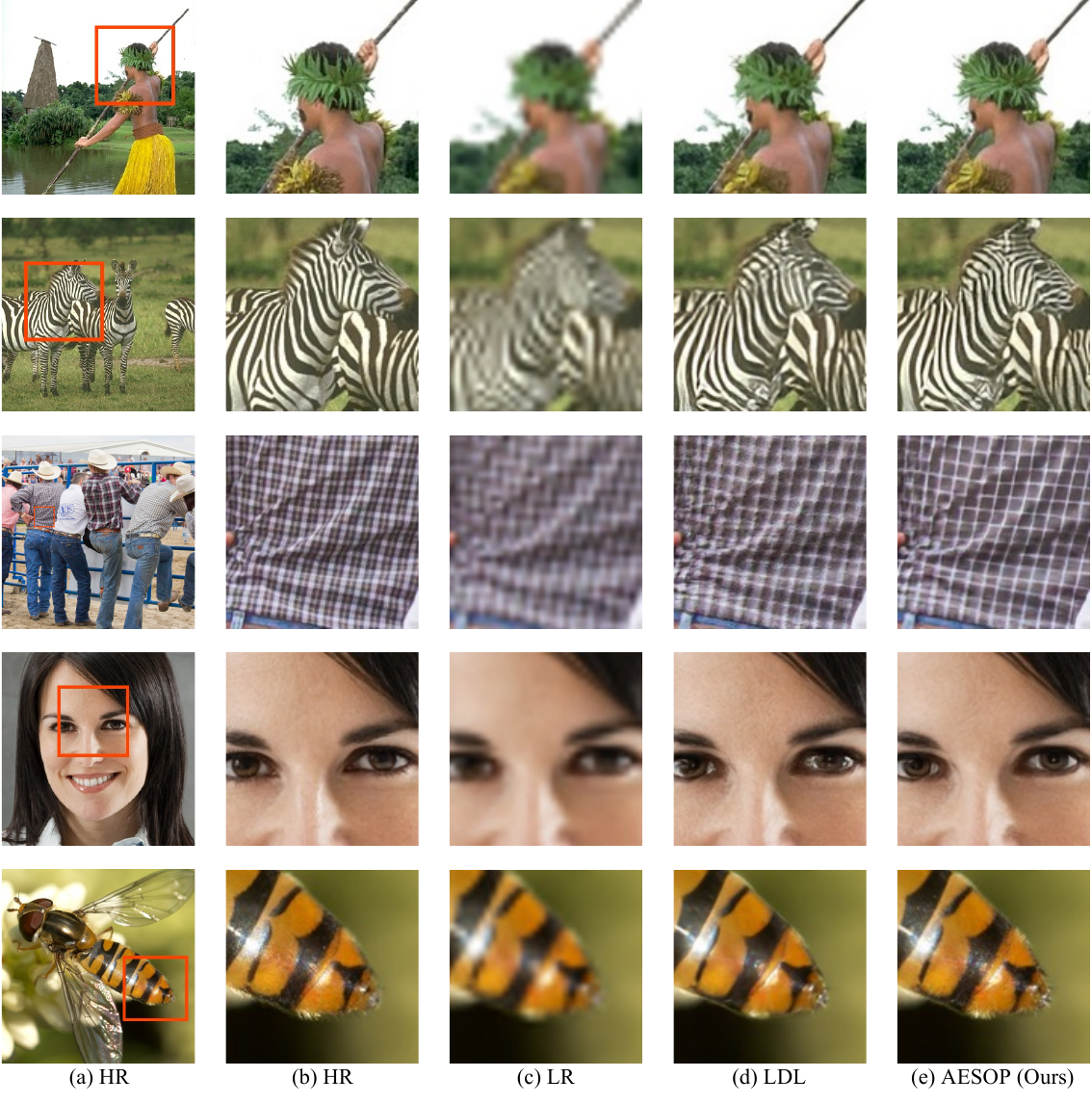}
    \caption{Visualization of AESOP (ours) and the baseline method for the bicubic ×4 SR task with RRDB backbone. AESOP can generate fine-grained textures with a lower level of visual artifacts. \textbf{Zoom in for best view}.}
    \label{fig:supp_rrdb1}
\end{figure*}

\begin{figure*}
    \centering
    \includegraphics[width=\linewidth]{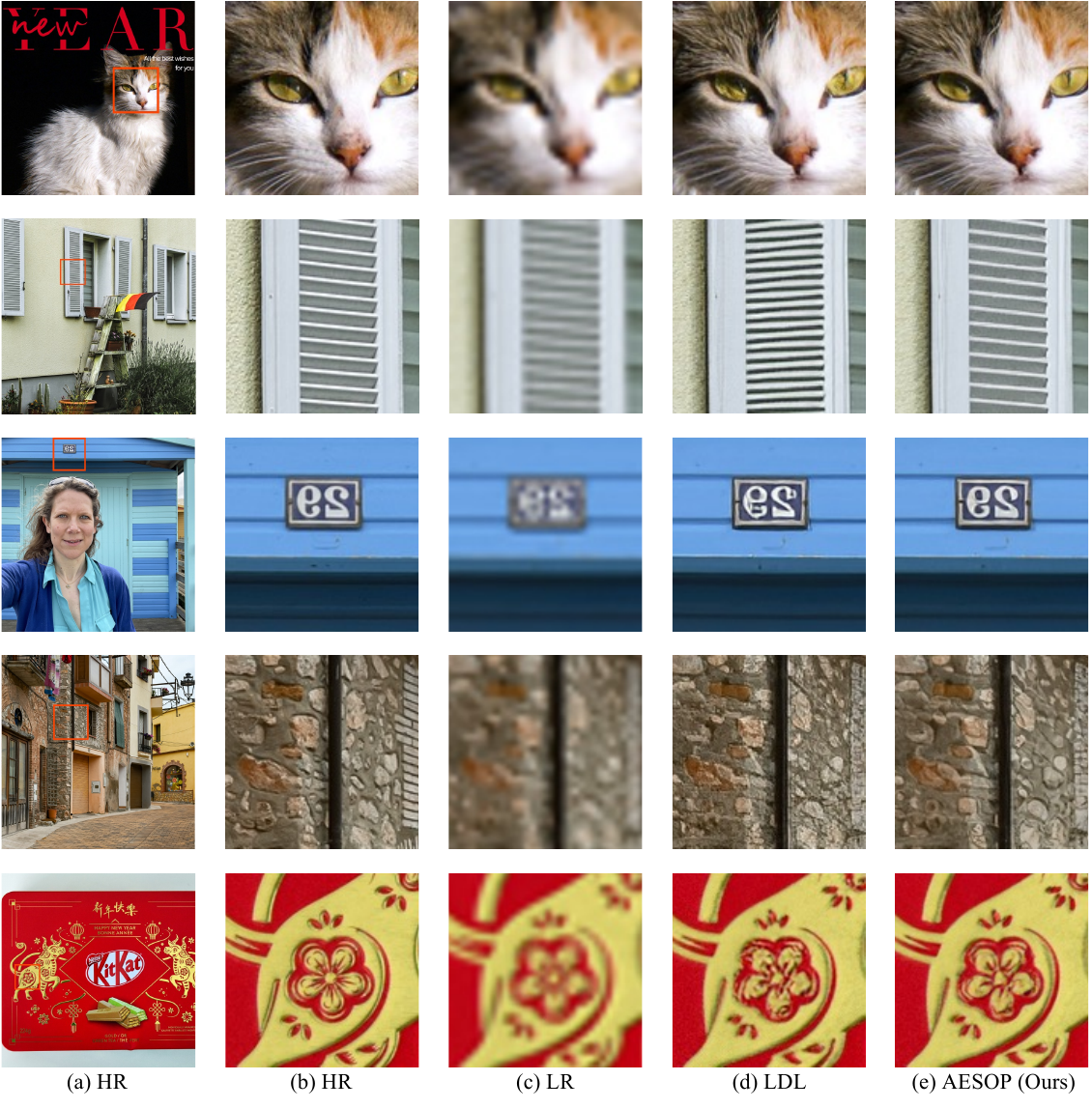}
    \caption{Visualization of AESOP (ours) and the baseline method for the bicubic ×4 SR task with RRDB backbone. AESOP can generate fine-grained textures with a lower level of visual artifacts. \textbf{Zoom in for best view}.}
    \label{fig:supp_rrdb2}
\end{figure*}

\vspace{5pt}
\noindent\textbf{Qualitative examples on benchmark datasets.}
To further illustrate the effectiveness of our method, we present an additional qualitative comparison between AESOP against the baseline method LDL \cite{details_or_artifact} on the bicubic $\times$4 SR task.
We provide results of tested methods, AESOP and LDL, on top of the SwinIR backbone (Fig.\ref{fig:supp_swinir1} and Fig.\ref{fig:supp_swinir2}) and the RRDB backbone (Fig.\ref{fig:supp_rrdb1} and Fig.\ref{fig:supp_rrdb2}). 
As can be seen, AESOP significantly improves perceptual quality while effectively suppressing visual artifacts observed in the baseline method.

\vspace{5pt}
\noindent\textbf{Additional perception-distortion trade-off curves.}
We provide extensive visualizations of the perception-distortion trade-off curves in Fig.\ref{fig:supp_pdtradeoff_PSNR_LPIPS_(RRDB, 128)}-\ref{fig:supp_pdtradeoff_PSNR_DISTS_(RRDB, 256)}. For CALGAN~\cite{calgan}, we present only a single data point rather than the full perception-distortion trade-off curve, as its official weights are not publicly available. 
Extensive results show that AESOP leads to substantial performance improvements against baselines in terms of the perception-distortion trade-off.
Aligning to Tab.\ref{tab:supp_maintable_supp}, AESOP also often outperforms CALGAN even without MoE-discriminator proposed in CALGAN. Additionally, we observe that AESOP often results in larger improvements for Swin Transformer-based methods (e.g., SwinIR, DRCT) compared to CNN-based methods (e.g., RRDB). This is likely because these models have greater capacity and benefit more from the enhanced reconstruction guidance provided by AESOP.
However, there are instances where AESOP does not always lead to improved performance. Specifically, AESOP often fails to enhance performance on the Manga109~\cite{manga109} dataset, which is consistent with the unexpected trade-off behaviors observed across most methods in this dataset. This limitation arises because Manga109 consists predominantly of comic images, which typically lack the fine-grained textures found in photorealistic datasets. The absence of such textures poses a challenge for perceptual SR methods, including AESOP, which are specifically designed to enhance and preserve realistic textures. Consequently, without the presence of these detailed textures, AESOP's advantages in minimizing fidelity bias and preserving perceptual variance are less pronounced, leading to suboptimal performance in this particular dataset.

\clearpage

%%%%%%%%%%%%%%%% PD trade-off %%%%%%%%%%%%%%%%%%%%%%%%% %%%%%%%%%%%%%%%% PD trade-off %%%%%%%%%%%%%%%%%%%%%%%%%
%%%%%%%%%%%%%%% PD trade-off %%%%%%%%%%%%%%%%%%%%%%%%% 

\input{PD/pdtrade_off_figure_list}

%%%%%%%%%%%%%%%% PD trade-off %%%%%%%%%%%%%%%%%%%%%%%%%
%%%%%%%%%%%%%%%% PD trade-off %%%%%%%%%%%%%%%%%%%%%%%%% %%%%%%%%%%%%%%%% PD trade-off %%%%%%%%%%%%%%%%%%%%%%%%%

%% file: supp/realworld_nriqa.tex
\begin{table*}[t]
\centering
\small
\renewcommand{\arraystretch}{0.7} % Adjust row spacing
\resizebox{0.7\textwidth}{!}{%
\begin{tabular}{l|l|cccc}
\toprule
\multicolumn{1}{l|}{Dataset} & \multicolumn{1}{l|}{Method} & \multicolumn{1}{c}{NIQE$\downarrow$} & \multicolumn{1}{c}{MANIQA} & \multicolumn{1}{c}{MUSIQ} & \multicolumn{1}{c}{CLIP-IQA} \\ \midrule\midrule
\multirow{5}{*}{\centering RealSRv3 \cite{RealSR_dataset1_RealSRv3}} 
                          & ESRGAN~\cite{SISR7_ESRGAN}     & 7.7326 & 0.2043 &	  29.0494 &	  0.2362 \\
                          & BSRGAN~\cite{BSRGAN}                           & 4.6519 &	0.3698 &	63.5908 &	0.5439 \\
                          & Real-ESRGAN~\cite{RealESRGAN}                  & 4.6790 &	0.3662 &	59.6855 &	0.4901 \\
                          & LDL~\cite{details_or_artifact}                 & 4.8869 &	0.3706 &	60.1015 &	0.4883 \\
                          &  AESOP~(Ours) & \textbf{4.2337} &	\textbf{0.4136} &	\textbf{63.6489} &	\textbf{0.5687} \\ \midrule
\multirow{5}{*}{\centering DRealSR \cite{RealSR_datasets3_DRealSR}}  
                          & ESRGAN~\cite{SISR7_ESRGAN}      & 8.3949 & 0.2115 &	  20.2861 &	  0.2468 \\
                          & BSRGAN~\cite{BSRGAN}                           & 4.6809 &	0.3381 &	35.4973 &	0.5614 \\
                          & Real-ESRGAN~\cite{RealESRGAN}                  & 4.7152 &	0.3404 &	35.2747 &	0.5098 \\
                          & LDL~\cite{details_or_artifact}                 & 5.0974 &	0.3393 &	35.9026 &	0.5137 \\
                          &  AESOP~(Ours) & \textbf{4.1922} &	\textbf{0.3917} &	\textbf{36.5533} &	\textbf{0.6025} \\ \midrule
                          
\end{tabular}%
}
\caption{Quantitative results of AESOP and baseline methods in real-world settings. All methods except ESRGAN are trained for the real-world SR task. The best results of each group are highlighted in \textbf{bold}. $\downarrow$ means lower is better. If not specified, higher is better. Due to memory constraints, images were cropped before evaluating CLIP-IQA scores for the DRealSR dataset.}
\label{tab:supp_full_realworldsr_nriqa}
\end{table*}

%% file: supp/fid_table.tex
\begin{table*}[!t]
\centering
\resizebox{\textwidth}{!}{%
\begin{tabular}{@{}cl|ccccc|c|cccc@{}}
\toprule
 & Backbone & \multicolumn{6}{c|}{RRDB} & \multicolumn{4}{c}{SwinIR} \\ 
\midrule\midrule
\multicolumn{1}{l}{Metrics} &
  \multicolumn{1}{l|}{Benchmark} &
  ESRGAN &
  SPSR &
  LDL* &
  LDL &
  AESOP &
  AESOP &
  +GAN &
  LDL* &
  LDL &
  AESOP \\
% & Backbone & RRDB & RRDB & RRDB & RRDB & RRDB & RRDB & SwinIR & SwinIR & SwinIR & SwinIR  \\ 
\midrule
\multicolumn{2}{l|}{\makecell{Recon. Objective}} & $\Loss{}_\text{pix}$ & $\Loss{}_\text{pix}$ & $\Loss{}_\text{pix}$ & $\Loss{}_\text{pix}$ & \multicolumn{1}{c|}{\textbf{$\Loss{}_\text{\AAEESOP{}}$}} & \multicolumn{1}{c|}{\textbf{$\Loss{}_\text{\AAEESOP{}}$}} & $\Loss{}_\text{pix}$ & $\Loss{}_\text{pix}$ & $\Loss{}_\text{pix}$ & \multicolumn{1}{c}{\textbf{$\Loss{}_\text{\AAEESOP{}}$}} \\
\midrule
% \multicolumn{1}{l}{}   & \multicolumn{1}{l}{\makecell{Patch size \\ \footnotesize{(Training)}}} & \multicolumn{4}{c}{128}               & \multicolumn{6}{c}{256}                               \\
% \multicolumn{1}{l}{}   & \multicolumn{1}{l|}{\makecell{Patch size \\ \footnotesize{(Training)}}} & 128 & 128 & 128 & 128 & 256 & 256 & 256 & 256 & 256 & 256 & 256\\
\multicolumn{2}{l|}{\makecell{Patch size (Training)}} & 128 & 128 & 128 & 128 & 128  & 256 & 256 & 256 & 256 & 256 \\
\midrule
\multirow{7}{*}{pFID~$\downarrow$}  & Set14            & 65.220         & 70.990       &   -           & 57.132               & \textbf{56.727}      & 54.792               & -          & -          & 55.367                & \textbf{53.175 }\\ 
                                    & Manga109         & 29.326         & 28.314       &   -           & 23.895               & \textbf{23.384}      & 22.833               & -          & -          & 21.766                & \textbf{21.290 }\\ 
                                    & General100       & 50.062         & 50.053       &   -           & 43.406               & \textbf{42.117}      & 41.041               & -          & -          & 42.028                & \textbf{40.199 }\\ 
                                    & Urban100         & 32.094         & 31.105       &   -           & 28.380               & \textbf{27.875}      & 27.017               & -          & -          & 26.972                & \textbf{25.613 }\\ 
                                    % & DIV2K100         & 14.624         & 13.719       &   -           & \textbf{12.822}      & 13.019               & 12.175               & -          & -          & 12.119                & \textbf{11.814 }\\ 
                                    & BSD100           & 69.943         & 68.370       &   -           & 64.058               & \textbf{57.864}      & 56.844               & -          & -          & 59.653                & \textbf{57.118 }\\ 
                                    & LSDIR            & 14.579         & 14.110       &   -           & 12.537               & \textbf{12.220}      & 11.718               & -          & -          & 12.056                & \textbf{11.387 }\\ \midrule
\multirow{7}{*}{FID~$\downarrow$}   & Set14            & 54.939         & 53.919       & 43.454        & \textbf{43.479}      & 46.828               & 38.907               & 48.910     & 46.057     & 46.110                & \textbf{45.411 }\\ 
                                    & Manga109         & 11.559         & 10.663       & 10.161        & 10.162               & \textbf{9.230 }      & 9.446                & 9.703      & 8.680      & \textbf{8.677}        & 9.256  \\ 
                                    & General100       & 29.850         & 30.172       & 27.211        & \textbf{27.220}      & 27.425               & 25.201               & 27.557     & 25.304     & 25.301                & \textbf{24.592 }\\ 
                                    & Urban100         & 20.354         & 18.676       & 16.351        & \textbf{16.355}      & 16.865               & 16.723               & 17.555     & 16.282     & 16.287                & \textbf{15.547 }\\ 
                                    % & DIV2K100         & 13.558         & 13.779       & 12.121        & 12.128               & \textbf{11.673}      & 12.160               & 12.736     & 12.075     & \textbf{12.075}       & 12.506 \\ 
                                    & BSD100           & 50.752         & 48.349       &   -           & 44.053               & \textbf{41.130}      & 40.751               & -          & -          & 41.954                & \textbf{41.721 }\\ 
                                    & LSDIR            & 17.552         & 16.056       &   -           & 15.229               & \textbf{14.748}      & 14.802               & -          & -          & 14.510                & \textbf{14.397 }\\ \hline\hline
\end{tabular}%
}
\caption{
Quantitative results of the proposed method~(\AAEESOP{}) against baseline methods. We report Frechet Inception Distance~(FID) and patch-FID~(pFID) scores. LDL* indicates that scores are from the official paper. All other scores are evaluated in our settings, with officially provided pretrained weights. The best results of each group are highlighted in \textbf{bold}, based on scores evaluated in our settings. 
}
\label{tab:fid_table}
\end{table*}

%% file: supp/lraepsnr_rrdb_table.tex
\begin{table*}[h]
\centering
\setlength{\tabcolsep}{3pt}
\footnotesize
\renewcommand{\arraystretch}{0.9} % Adjust row spacing
\resizebox{0.8\textwidth}{!}{%
\begin{tabular}{@{}c|l|ccccccc@{}}
\midrule
\multirow{2}{*}{\rotatebox{0}{}}{Metric~} &  \multicolumn{1}{c|}{Method} & Set14 & Manga109 & General100 & Urban100 & DIV2K-val & BSD100 & LSDIR \\
\midrule
\multirow{4}{*}{\rotatebox{0}{AE-PSNR~}} 
& ESRGAN~\cite{SISR7_ESRGAN}        & 30.280	& 31.165 &	32.663 &	27.198 &	31.668 &	28.991 &	27.636 \\
& SPSR~\cite{SPSR}                  & 30.602	& 31.351 &	32.670 &	27.508 &	31.737 &	29.029 &	27.881 \\
& LDL~\cite{details_or_artifact}    & 31.180	& 32.608 &	33.823 &	28.488 &	32.597 &	29.595 &	28.625 \\
&  AESOP~(Ours)  & \textbf{31.341}	& \textbf{32.843} &	\textbf{33.956} &	\textbf{28.529} &	\textbf{32.740} &	\textbf{29.737} &	\textbf{28.812} \\
\midrule
\multirow{4}{*}{\rotatebox{0}{LR-PSNR~}} 
& ESRGAN~\cite{SISR7_ESRGAN}        & 43.892 & 43.908 & 45.259 & 42.879 & 45.689 & 43.823 & 42.718 \\
& SPSR~\cite{SPSR}                  & 43.835 & 44.359 & 44.656 & 42.666 & 44.717 & 42.719 & 42.364 \\
& LDL~\cite{details_or_artifact}    & 46.497 & 47.603 & 48.184 & 45.975 & 47.793 & 45.307 & 45.295 \\
&  AESOP~(Ours)  & \textbf{46.625} & \textbf{48.188} & \textbf{48.653} & \textbf{46.280} & \textbf{48.272} & \textbf{45.837} & \textbf{45.571} \\
\midrule
\end{tabular}
}
\caption{
Quantitative comparison between the proposed method (AESOP) and baseline methods. We report AE-PSNR and LR-PSNR scores using the RRDB backbone. AE-PSNR measures how accurately the method estimates fidelity bias factors, while LR-PSNR evaluates how well the generated images align with the input LR image. The best result in each group is highlighted in bold.
}
\label{tab:lrpsnr_aepsnr_table_rrdb}
\end{table*}

%% file: supp/supp_maintable_supp.tex
% \usepackage{booktabs}
% \usepackage{multirow}
% \usepackage{graphicx}
\begin{table*}[]
\centering
\footnotesize
\renewcommand{\arraystretch}{0.7} % Adjust row spacing
\resizebox{0.7\textwidth}{!}{%
\begin{tabular}{@{}cl|cc|cc|cc@{}}
\toprule
                          & Backbone      & \multicolumn{2}{|c|}{RRDB}                                                         & \multicolumn{2}{c|}{SwinIR}                                                       & \multicolumn{2}{c}{DRCT}                            \\  \midrule\midrule
Metrics                   & Benchmark     & \multicolumn{1}{c}{CALGAN} & \multicolumn{1}{c|}{AESOP} & \multicolumn{1}{c}{CALGAN} & \multicolumn{1}{c|}{AESOP} & \multicolumn{1}{c}{LDL} & \multicolumn{1}{c}{AESOP} \\ \midrule
\multicolumn{2}{c|}{Recon. Objective}      & $\Loss{}_\text{pix}$                        & \multicolumn{1}{c|}{$\Loss{}_\text{AESOP}$}                     & $\Loss{}_\text{pix}$                        & \multicolumn{1}{c|}{$\Loss{}_\text{AESOP}$}                     & $\Loss{}_\text{pix}$                     & \multicolumn{1}{c}{$\Loss{}_\text{AESOP}$}                     \\ \midrule
\multirow{7}{*}{LPIPS $\downarrow$}    
                          & Set14         & -                          & 0.1053                      & -                          & 0.1027                      & 0.1086                  & \textbf{0.1022}           \\
                          & Manga109      & -                          & 0.0494                      & -                          & 0.0461                      & 0.0459                  & \textbf{0.0447}           \\
                          & General100    & 0.077                      & \textbf{0.0734}             & 0.074                      & \textbf{0.0710}             & 0.0727                  & \textbf{0.0722}           \\
                          & Urban100      & 0.108                      & \textbf{0.1033}             & 0.098                      & \textbf{0.0945}             & 0.1006                  & \textbf{0.0972}           \\
                          & DIV2K-val     & \textbf{0.091}             & 0.0936                      & \textbf{0.087}             & 0.0893                      & \textbf{0.0934}         & 0.0949                    \\
                          & BSD100        & 0.151                      & \textbf{0.1443}             & 0.147                      & \textbf{0.1385}             & 0.1462                  & \textbf{0.1451}           \\
                          & LSDIR         & -                          & 0.1123                      & -                          & 0.1071                      & 0.1131                  & \textbf{0.1129}           \\ \midrule
\multirow{7}{*}{DISTS $\downarrow$}    
                          & Set14         & -                          & 0.0825                      & -                          & 0.0819                      & 0.0889                  & \textbf{0.0830}           \\
                          & Manga109      & -                          & 0.0356                      & -                          & 0.0328                      & \textbf{0.0316}         & 0.0338                    \\
                          & General100    & 0.083                      & \textbf{0.0773}             & 0.081                      & \textbf{0.0762}             & 0.0782                  & \textbf{0.0775}           \\
                          & Urban100      & 0.082                      & \textbf{0.0768}             & 0.083                      & \textbf{0.0742}             & 0.0803                  & \textbf{0.0771}           \\
                          & DIV2K-val     & 0.049                      & \textbf{0.0484}             & 0.048                      & \textbf{0.0459}             & 0.0487                  & \textbf{0.0485}           \\
                          & BSD100        & 0.118                      & \textbf{0.1089}             & 0.128                      & \textbf{0.1072}             & 0.1136                  & \textbf{0.1072}           \\
                          & LSDIR         & -                          & 0.0612                      & -                          & 0.0591                      & 0.0635                  & \textbf{0.0621}           \\  \midrule
\multirow{7}{*}{PSNR $\uparrow$}     
                          & Set14         & -                          & 27.246                      & -                          & 27.421                      & 27.314                  & \textbf{27.796}           \\
                          & Manga109      & -                          & 29.747                      & -                          & 30.061                      & 29.979                  & \textbf{30.398}           \\
                          & General100    & 30.182                     & \textbf{30.251}             & -                          & 30.401                      & 30.143                  & \textbf{30.646}           \\
                          & Urban100      & 25.290                     & \textbf{25.541}             & -                          & 26.148                      & 26.038                  & \textbf{26.360}           \\
                          & DIV2K-val     & 28.863                     & \textbf{28.910}             & -                          & 29.137                      & 29.030                  & \textbf{29.456}           \\
                          & BSD100        & \textbf{25.925}            & 25.904                      & -                          & 25.930                      & 25.942                  & \textbf{26.324}           \\
                          & LSDIR         & -                          & 24.845                      & -                          & 25.038                      & 24.943                  & \textbf{25.354}           \\ \midrule
\multirow{7}{*}{SSIM $\uparrow$}      
                          & Set14         & -                          & 0.7371                      & -                          & 0.7438                      & 0.7403                  & \textbf{0.7546}           \\
                          & Manga109      & -                          & 0.8802                      & -                          & 0.8880                      & 0.8888                  & \textbf{0.8936}           \\
                          & General100    & 0.825                      & \textbf{0.8269}             & -                          & 0.8327                      & 0.8288                  & \textbf{0.8382}           \\
                          & Urban100      & 0.763                      & \textbf{0.7697}             & -                          & 0.7884                      & 0.7855                  & \textbf{0.7926}           \\
                          & DIV2K-val     & 0.790                      & \textbf{0.7951}             & -                          & 0.8023                      & 0.7994                  & \textbf{0.8085}           \\
                          & BSD100        & 0.676                      & \textbf{0.6783}             & -                          & 0.6813                      & 0.6812                  & \textbf{0.6921}           \\
                          & LSDIR         & -                          & 0.7202                      & -                          & 0.7289                      & 0.7253                  & \textbf{0.7353}           \\ \midrule\midrule
\end{tabular}%
}
\vspace{-5pt}
\caption{Additional quantitative evaluation on benchmark datasets. We also provide quantitative results of CALGAN~\cite{calgan} and DRCT~\cite{drct}. CALGAN is a recent work improving perceptual-oriented losses, while DRCT made improvements in the SR network architecture. AESOP mostly outperforms CALGAN~\cite{calgan} even without the MoE-discriminator. However, note that enhancements to network architectures and perceptual-oriented losses are beyond the scope of this work. The focus of this work is on the fidelity loss term $\Loss{}_\text{pix}$ under the perceptual SR framework. The best results of each group are highlighted in \textbf{bold}. Additionally, refer to the PD trade-off curve in Fig.\ref{fig:supp_pdtradeoff_PSNR_LPIPS_(RRDB, 128)}-\ref{fig:supp_pdtradeoff_PSNR_DISTS_(RRDB, 256)}.}
\label{tab:supp_maintable_supp}
\end{table*}

%% file: PD/pdtrade_off_figure_list.tex
%%%%%%%%%%%%% ============================ (RRDB, 128)

\clearpage
\begin{figure*}[htp]
    \centering
    % First Row
    \begin{tabular}{cc}
        \includegraphics[width=0.4\textwidth]{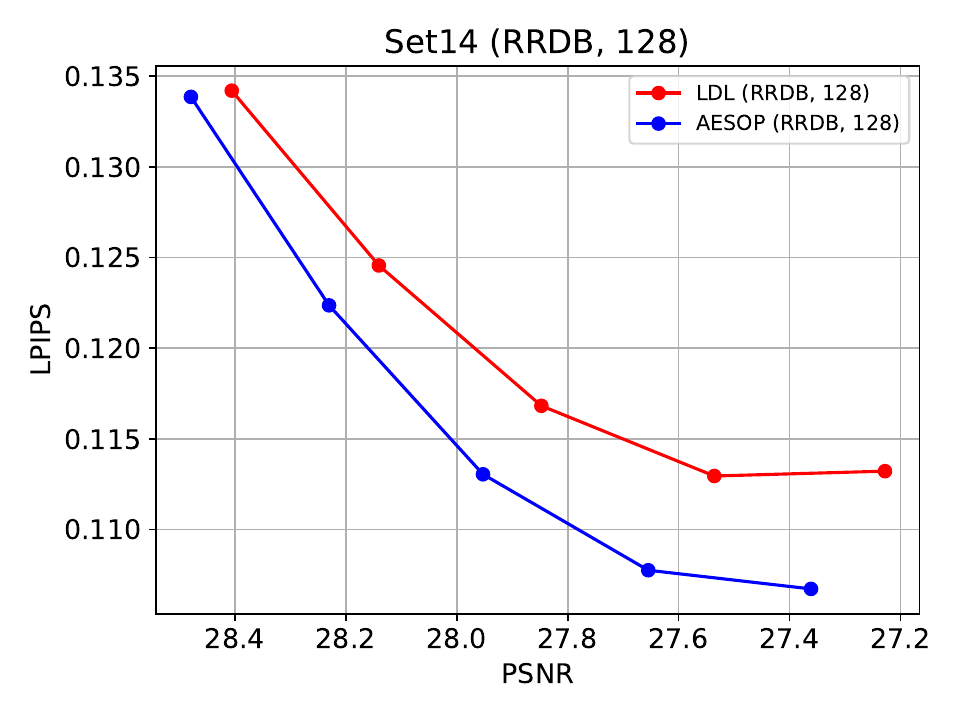}         & \includegraphics[width=0.4\textwidth]{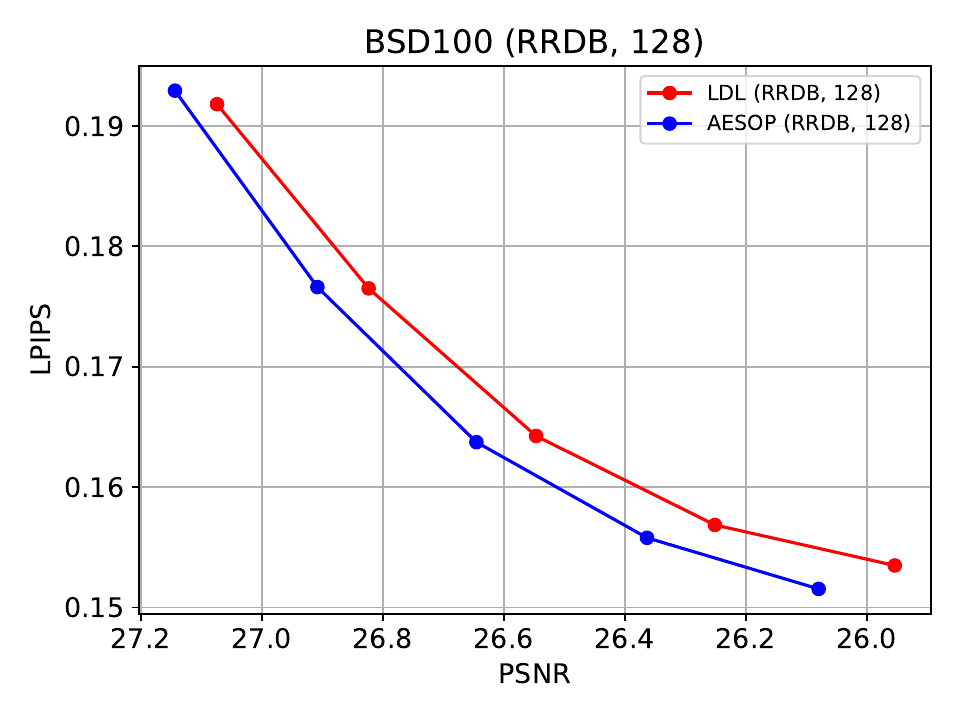} \\
        \includegraphics[width=0.4\textwidth]{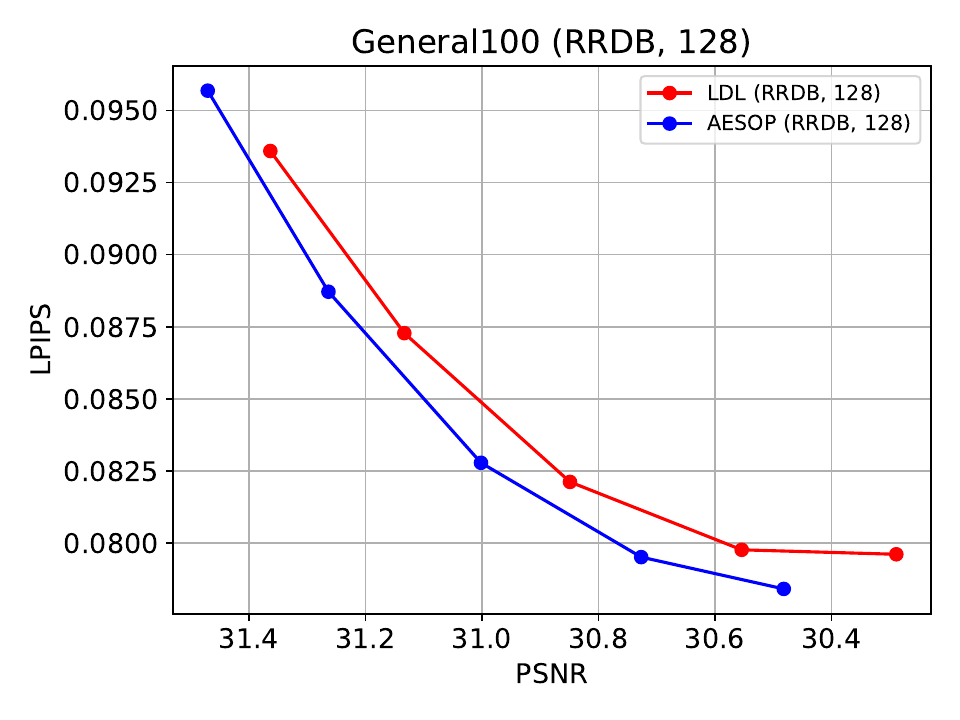}    & \includegraphics[width=0.4\textwidth]{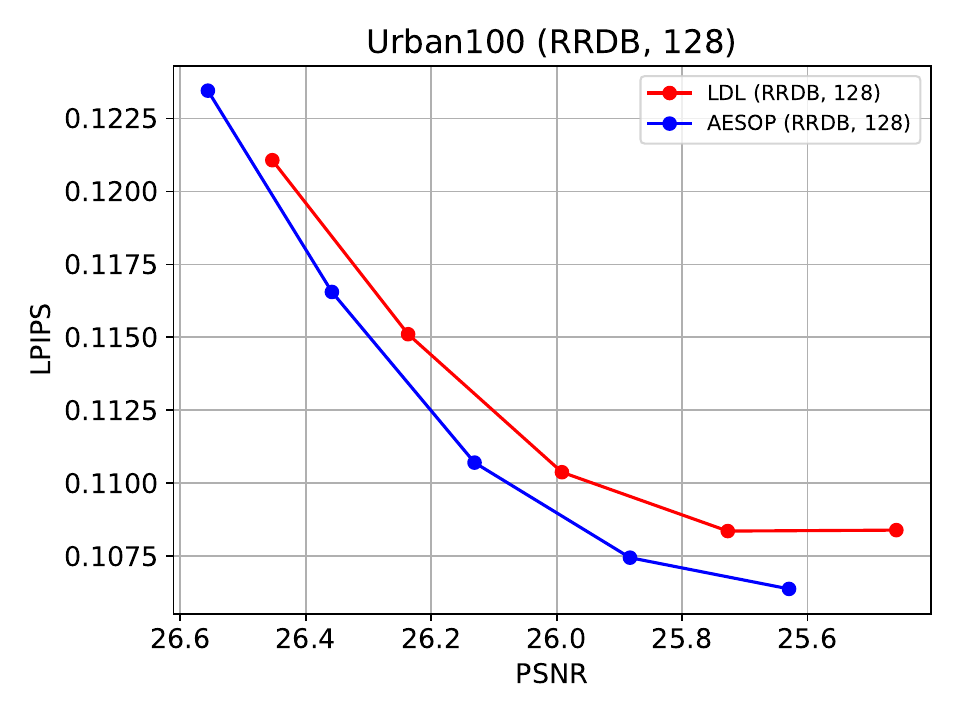} \\
        \includegraphics[width=0.4\textwidth]{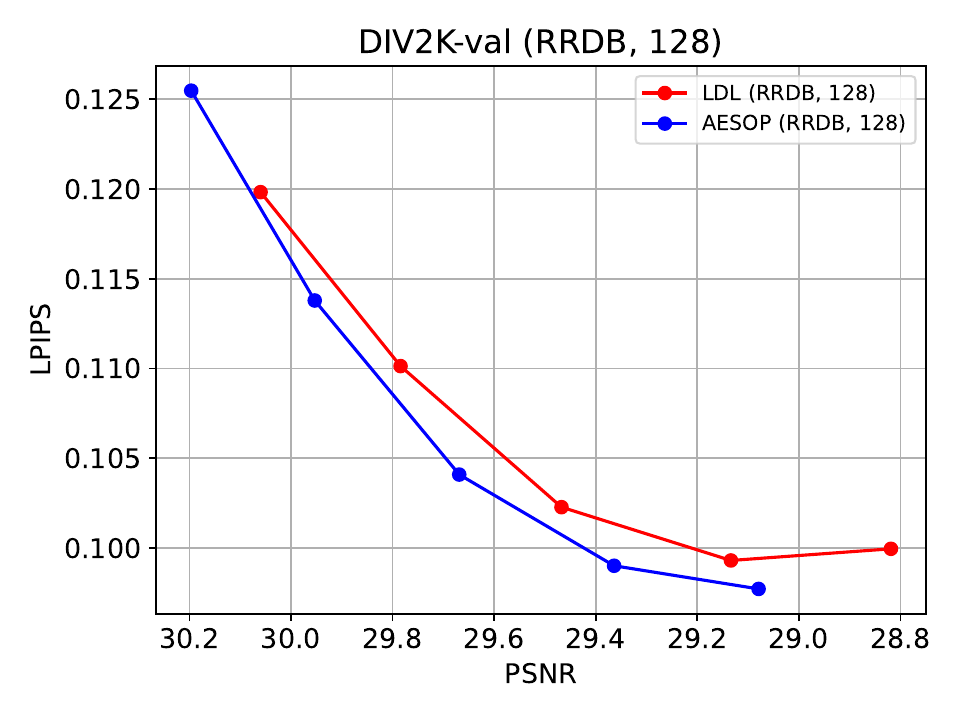}      & \includegraphics[width=0.4\textwidth]{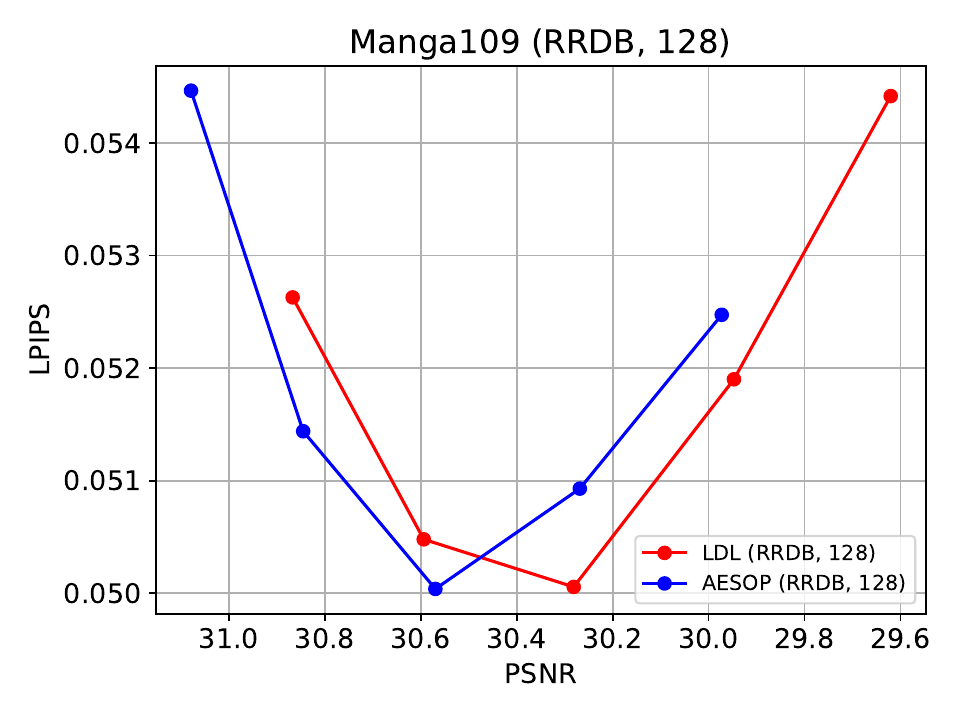} \\
        \includegraphics[width=0.4\textwidth]{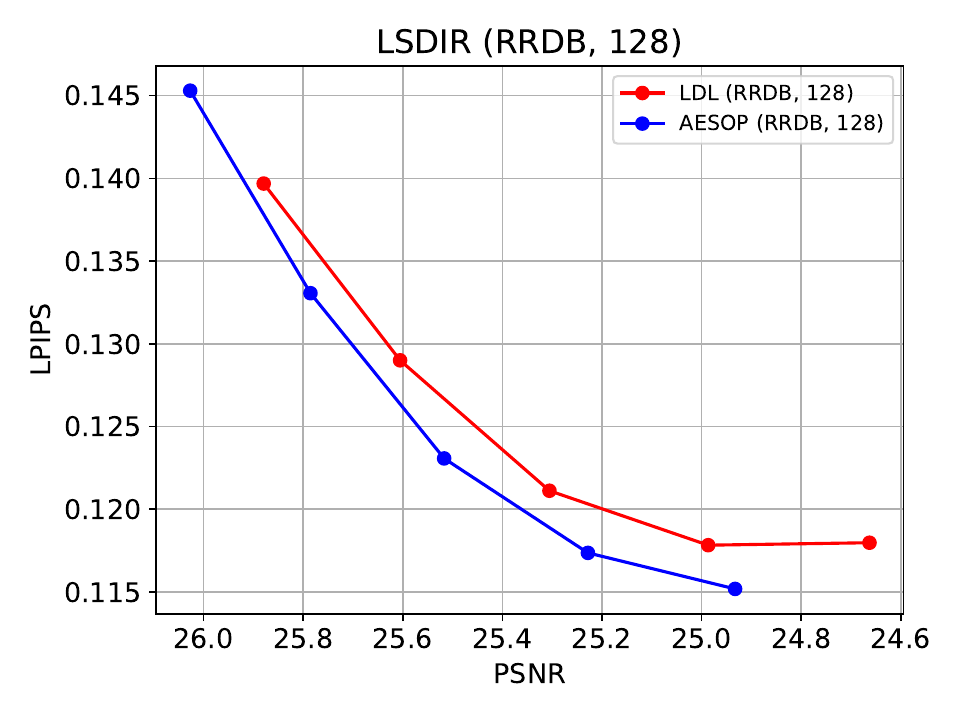}\\
    \end{tabular}
    \caption{The perception-distortion trade-off curve between AESOP and baseline methods on top of the RRDB~\cite{SISR7_ESRGAN} backbone. The training HR patch size is 128. AESOP often fails to improve the performance on the Manga109 dataset.}
    \label{fig:supp_pdtradeoff_PSNR_LPIPS_(RRDB, 128)}
\end{figure*}

\clearpage
\begin{figure*}[htp]
    \centering
    % First Row
    \begin{tabular}{cc}
        \includegraphics[width=0.4\textwidth]{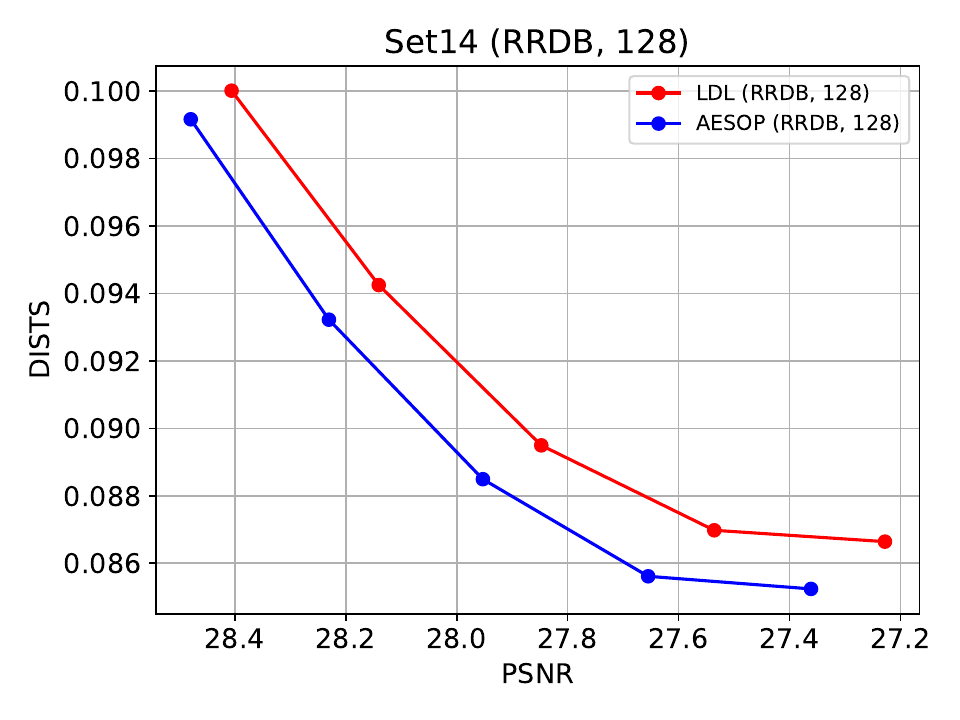}         & \includegraphics[width=0.4\textwidth]{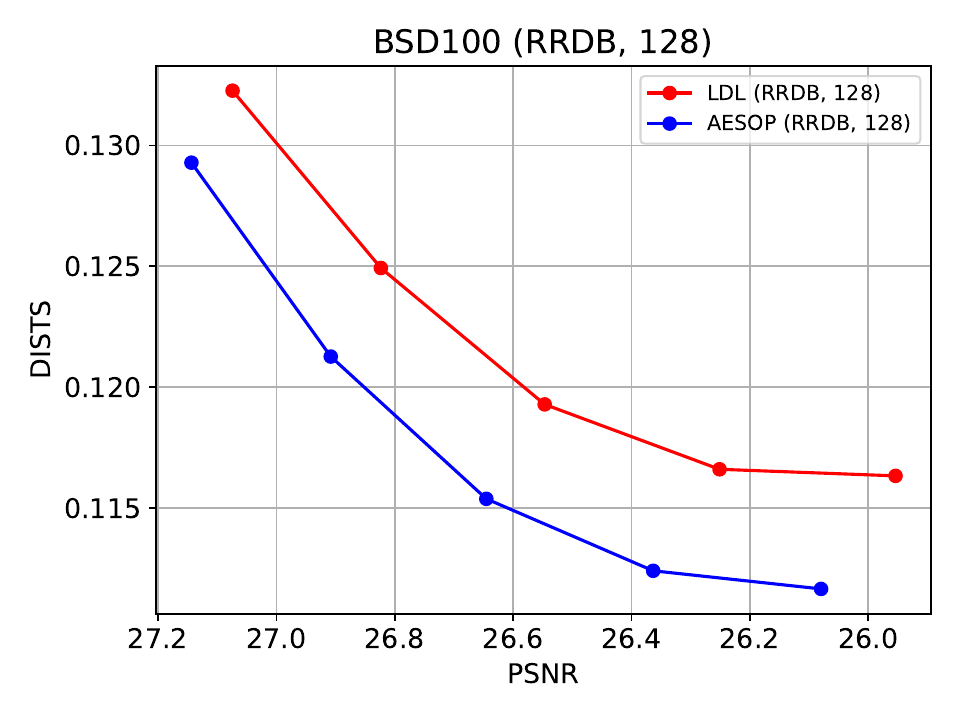} \\
        \includegraphics[width=0.4\textwidth]{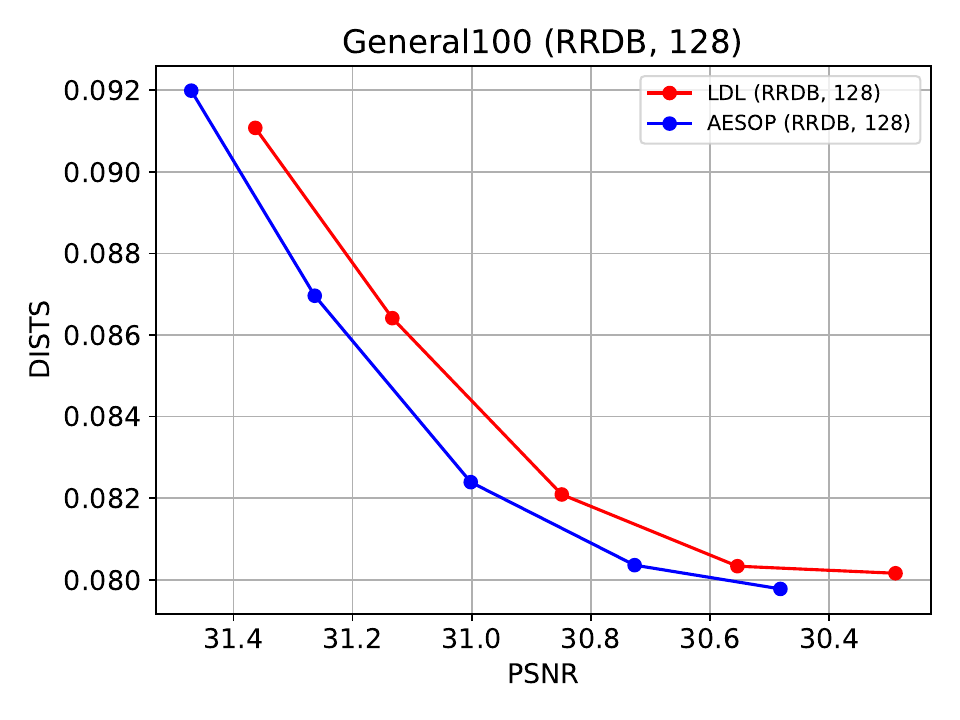}    & \includegraphics[width=0.4\textwidth]{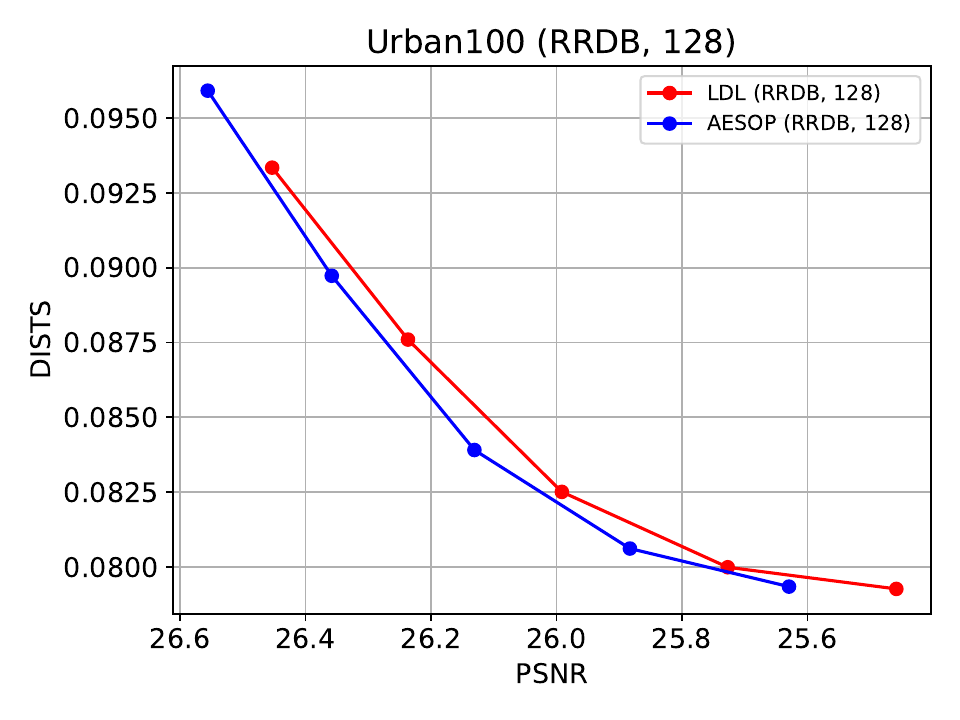} \\
        \includegraphics[width=0.4\textwidth]{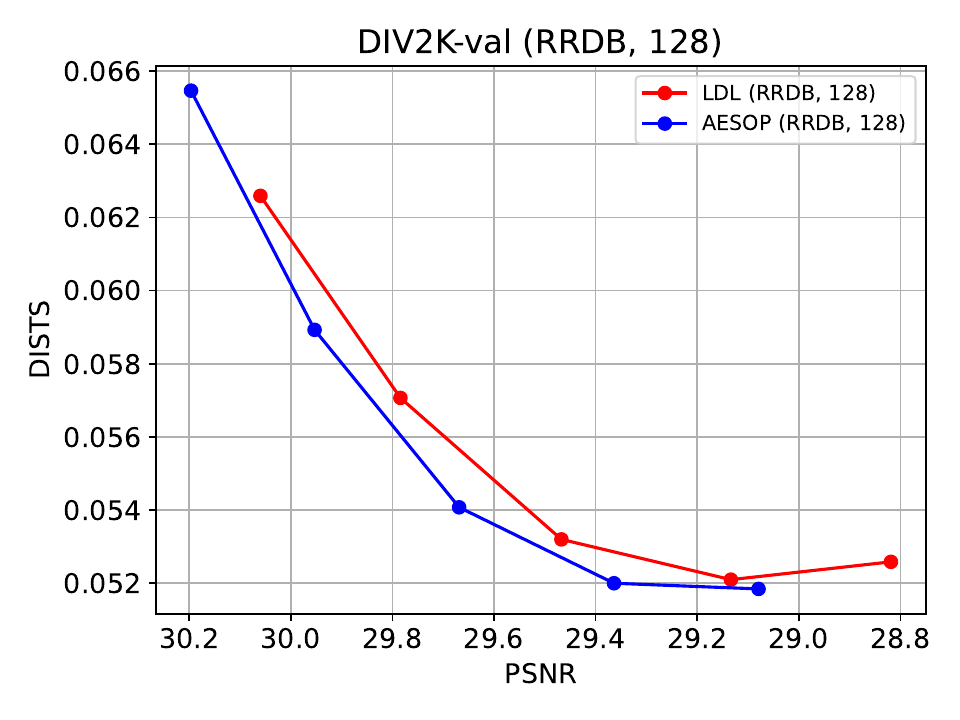}      & \includegraphics[width=0.4\textwidth]{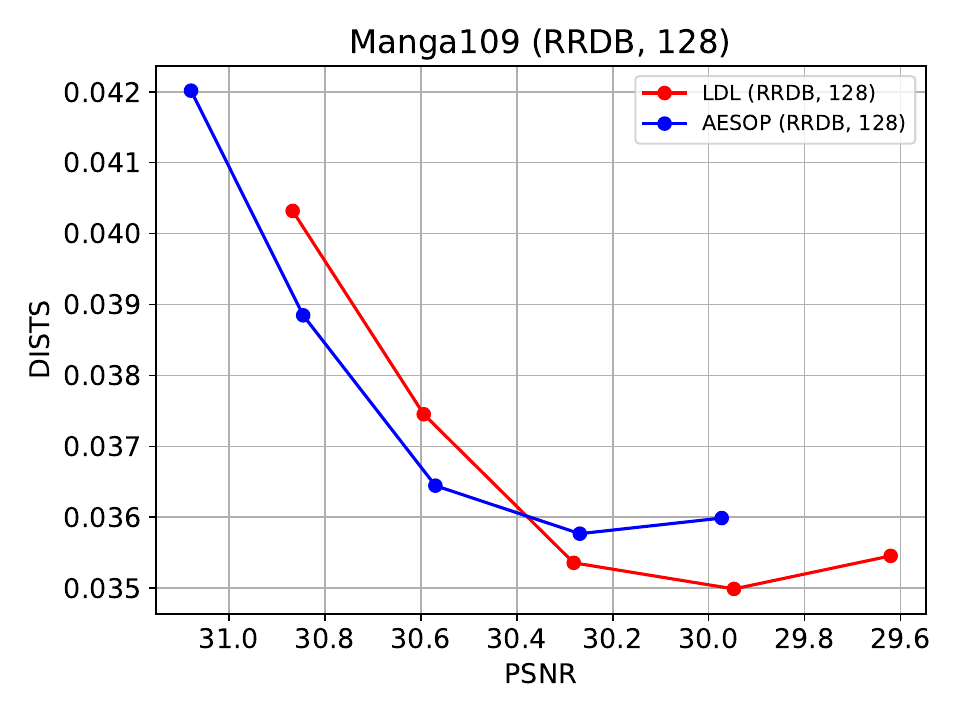} \\
        \includegraphics[width=0.4\textwidth]{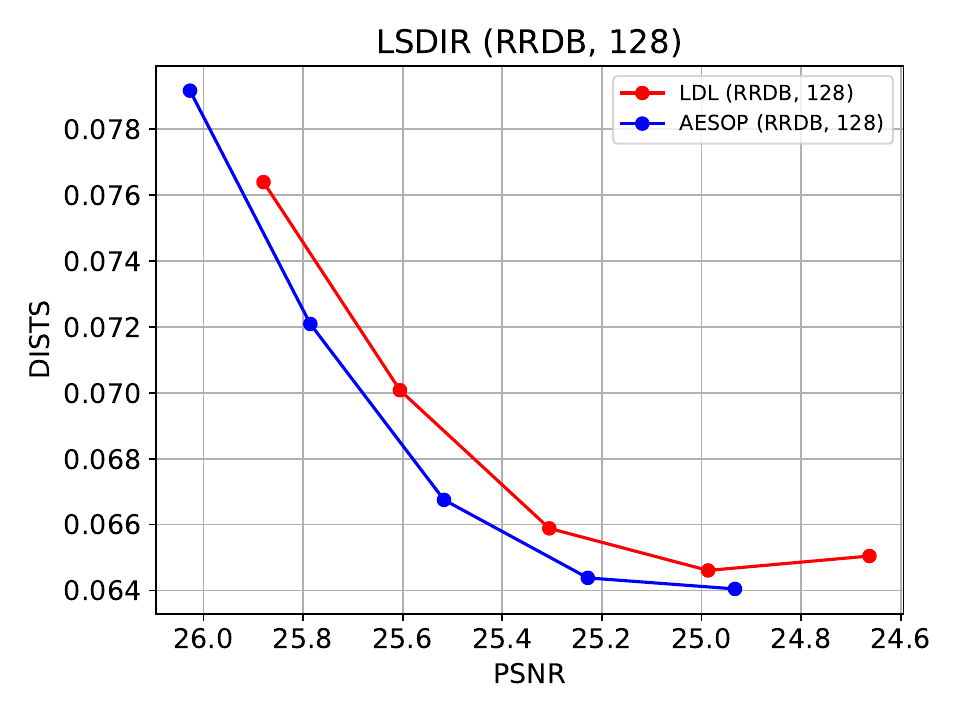}\\
    \end{tabular}
    \caption{The perception-distortion trade-off curve between AESOP and baseline methods on top of the RRDB~\cite{SISR7_ESRGAN} backbone. The training HR patch size is 128. AESOP often fails to improve the performance on the Manga109 dataset.}
    \label{fig:supp_pdtradeoff_PSNR_DISTS_(RRDB, 128)}
\end{figure*}

%%%%%%%%%%%%% ============================ (DRCT, 256)

\clearpage
\begin{figure*}[htp]
    \centering
    % First Row
    \begin{tabular}{cc}
        \includegraphics[width=0.4\textwidth]{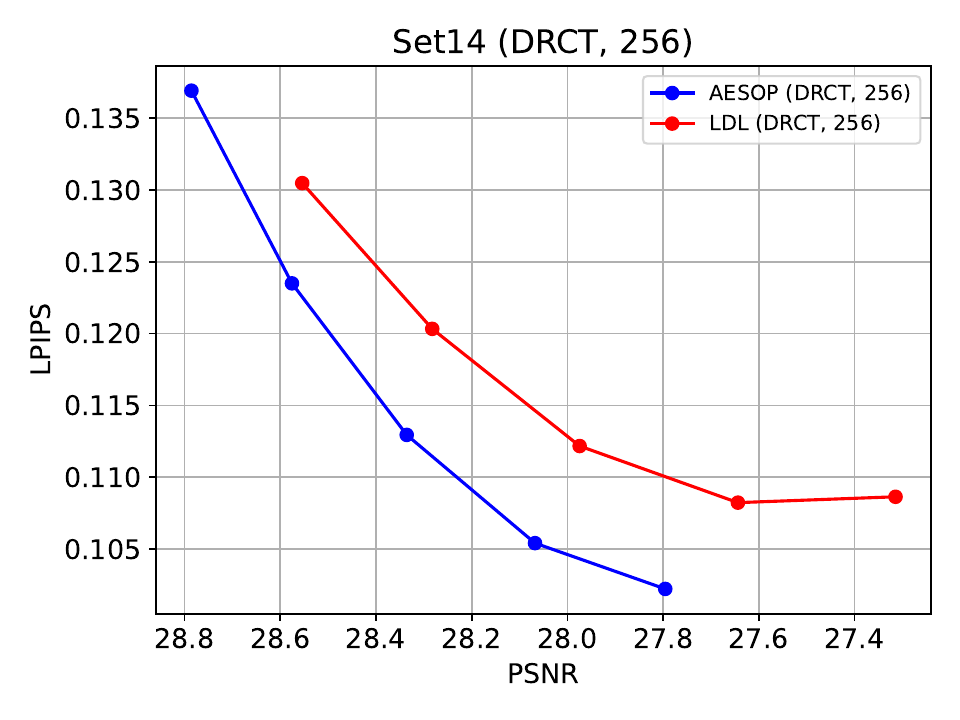}         & \includegraphics[width=0.4\textwidth]{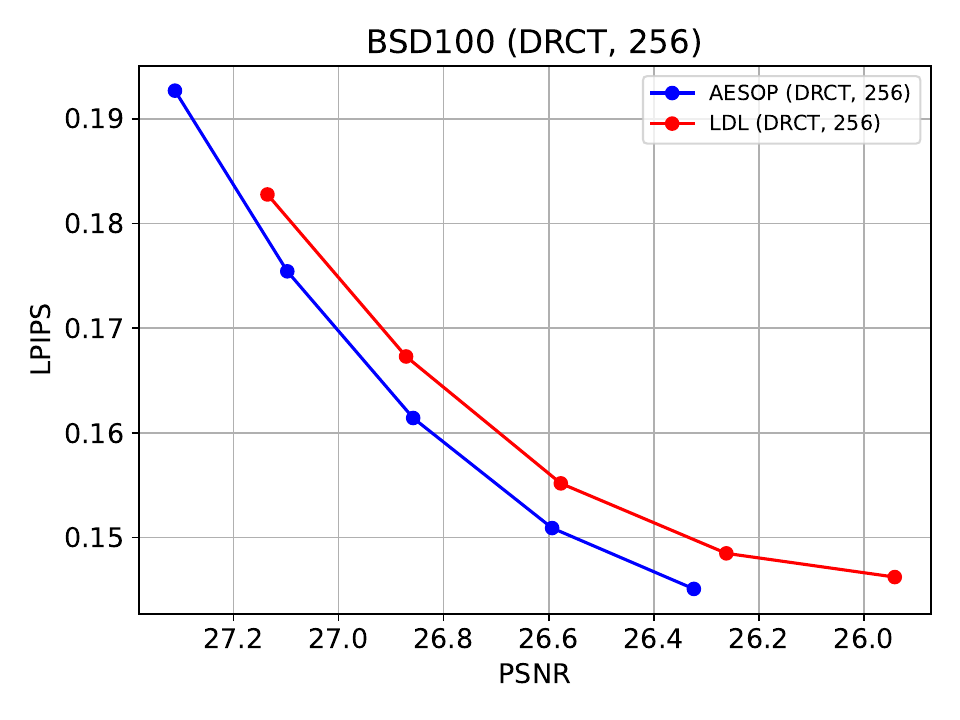} \\
        \includegraphics[width=0.4\textwidth]{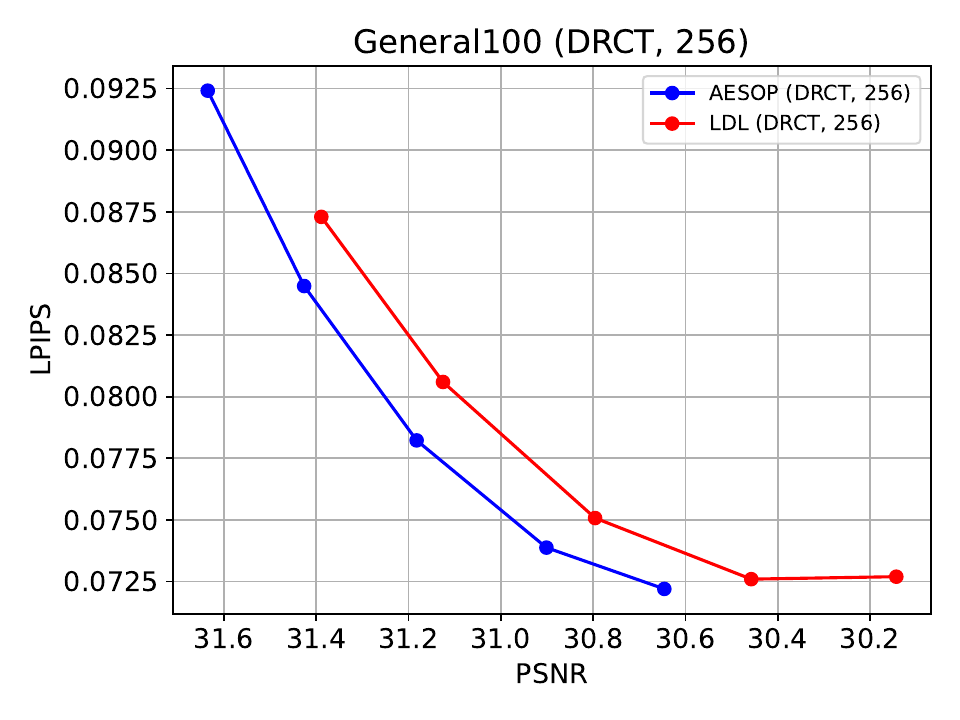}    & \includegraphics[width=0.4\textwidth]{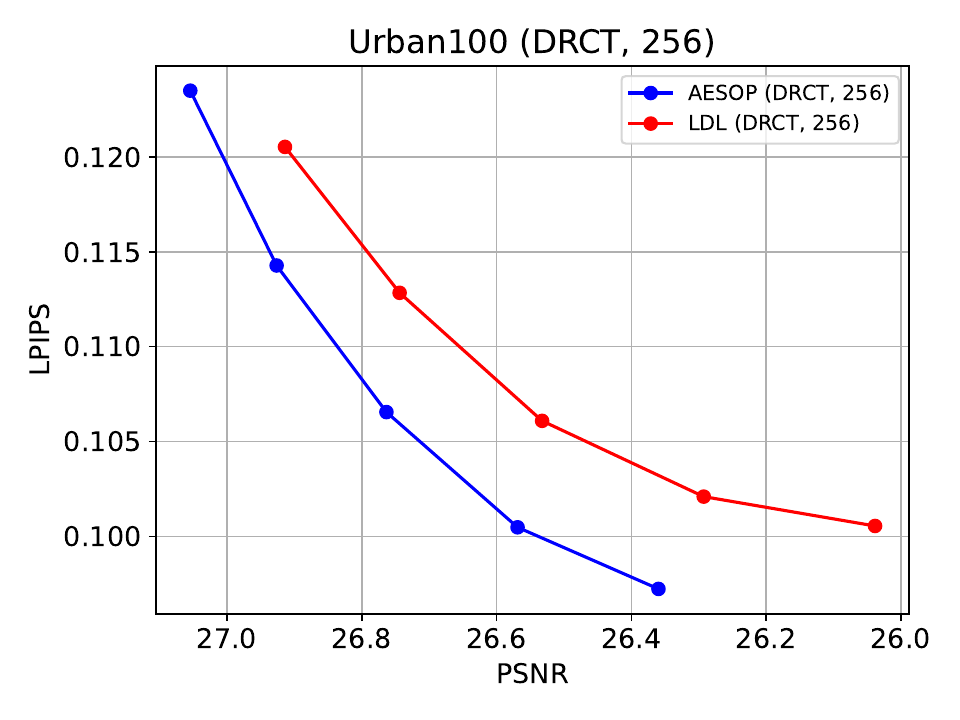} \\
        \includegraphics[width=0.4\textwidth]{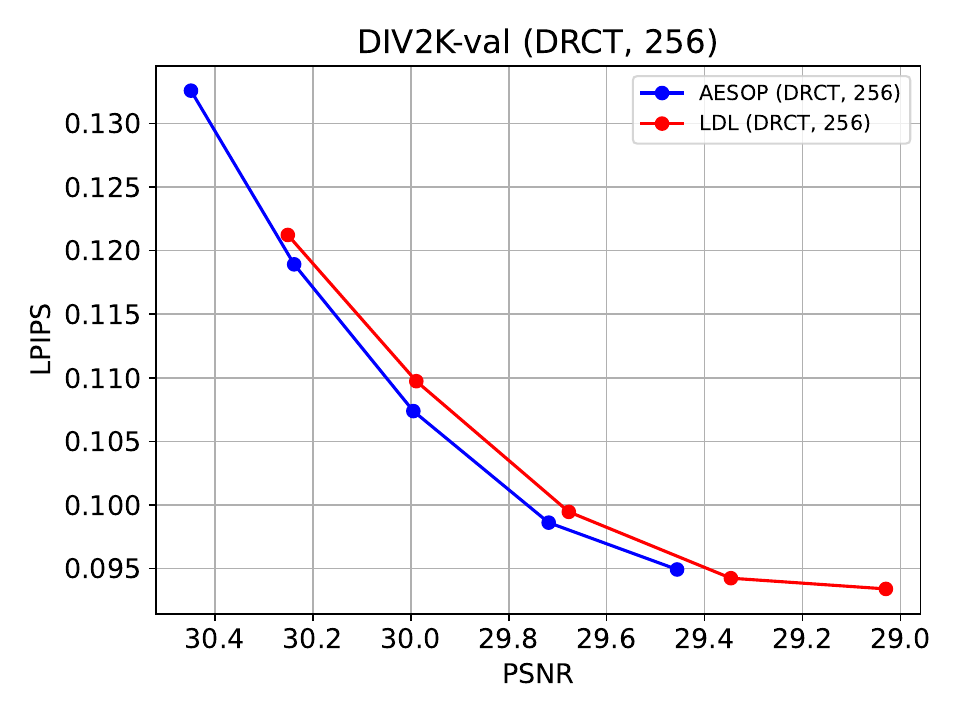}      & \includegraphics[width=0.4\textwidth]{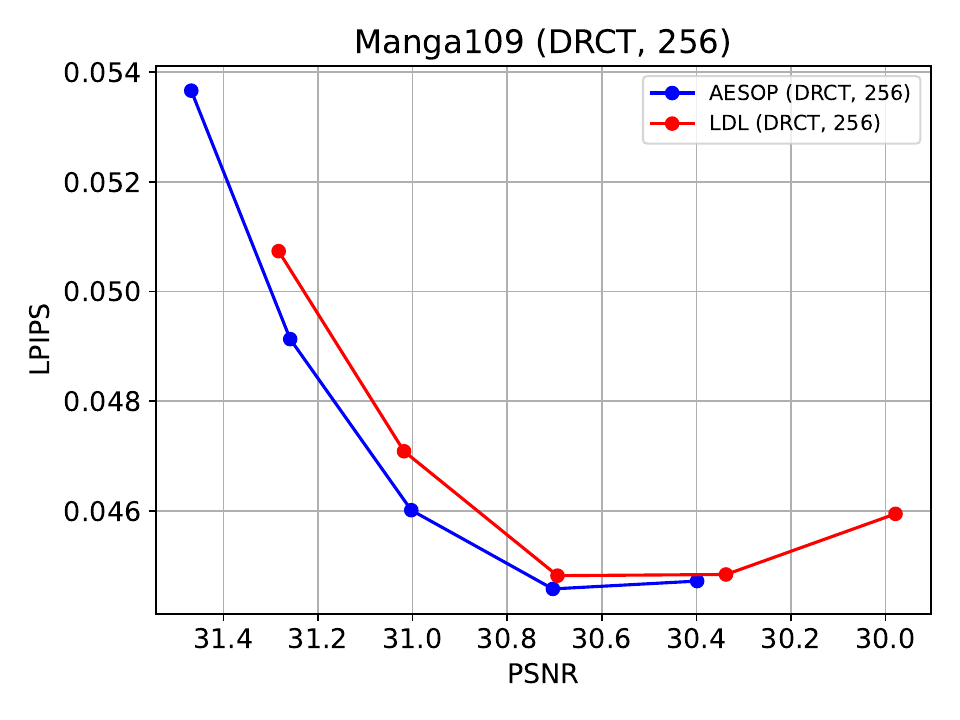} \\
        \includegraphics[width=0.4\textwidth]{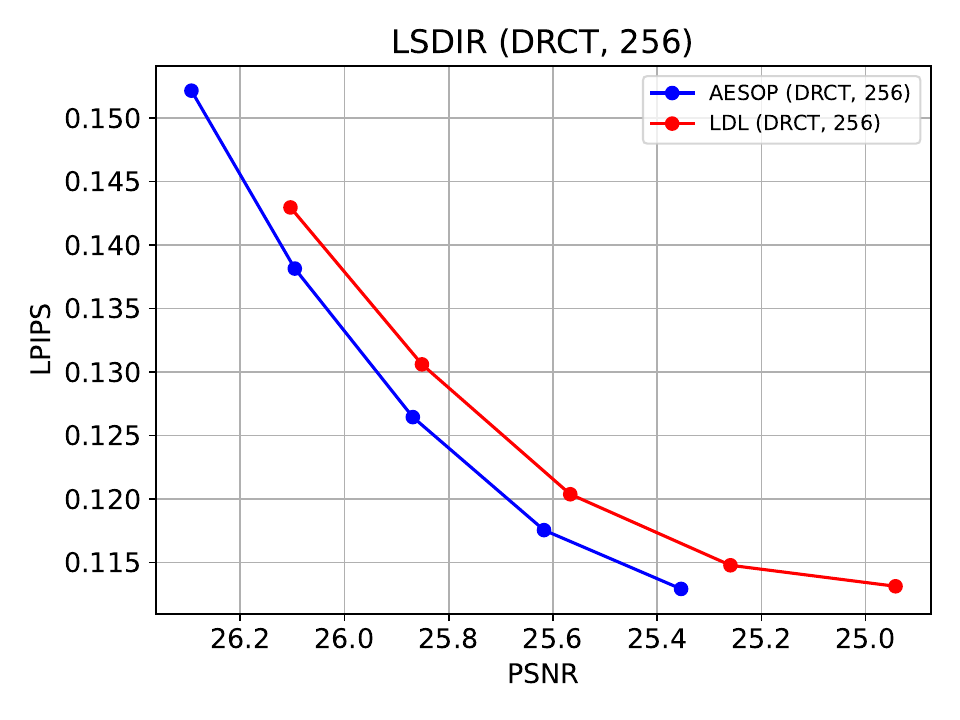}\\
    \end{tabular}
    \caption{The perception-distortion trade-off curve between AESOP and baseline methods on top of the DRCT~\cite{drct} backbone. The training HR patch size is 256. AESOP often fails to improve the performance on the Manga109 dataset.}
    \label{fig:supp_pdtradeoff_PSNR_LPIPS_(DRCT, 256)}
\end{figure*}

\clearpage
\begin{figure*}[htp]
    \centering
    % First Row
    \begin{tabular}{cc}
        \includegraphics[width=0.4\textwidth]{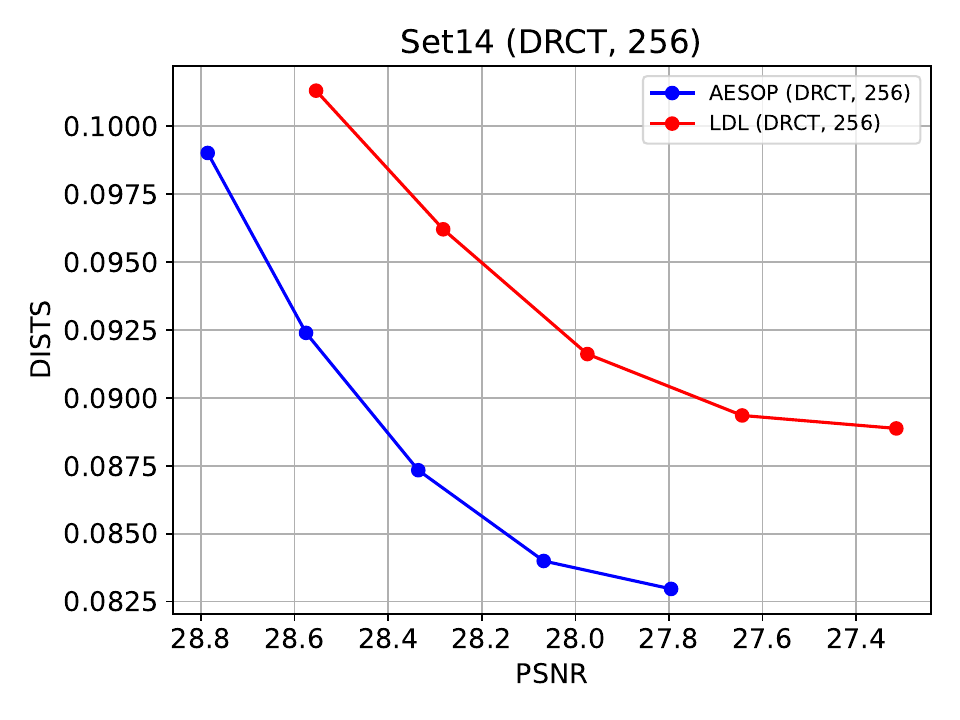}         & \includegraphics[width=0.4\textwidth]{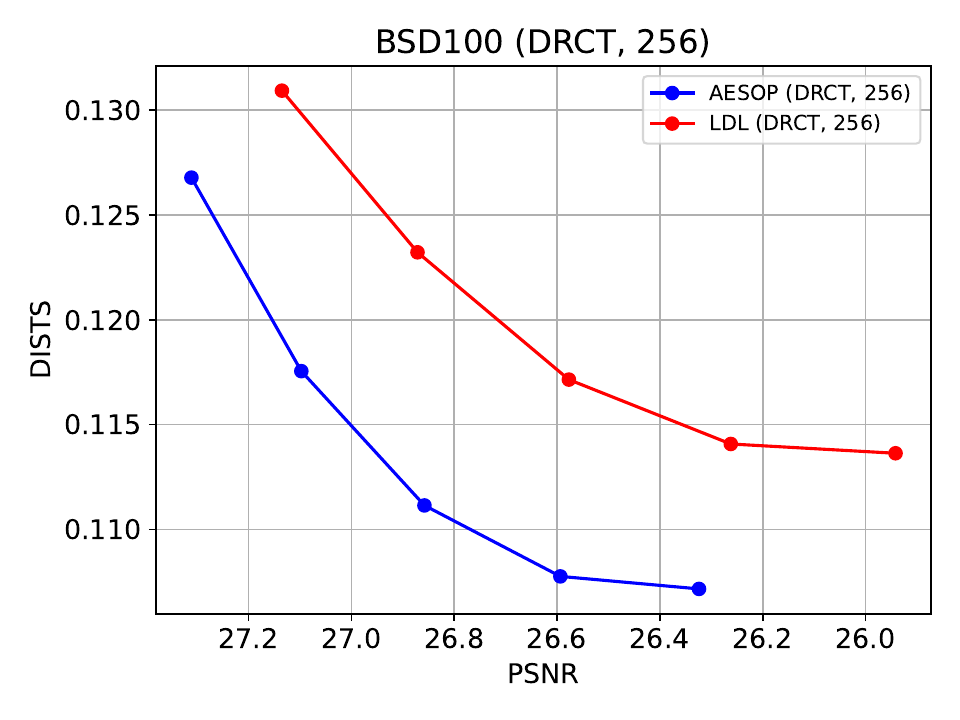} \\
        \includegraphics[width=0.4\textwidth]{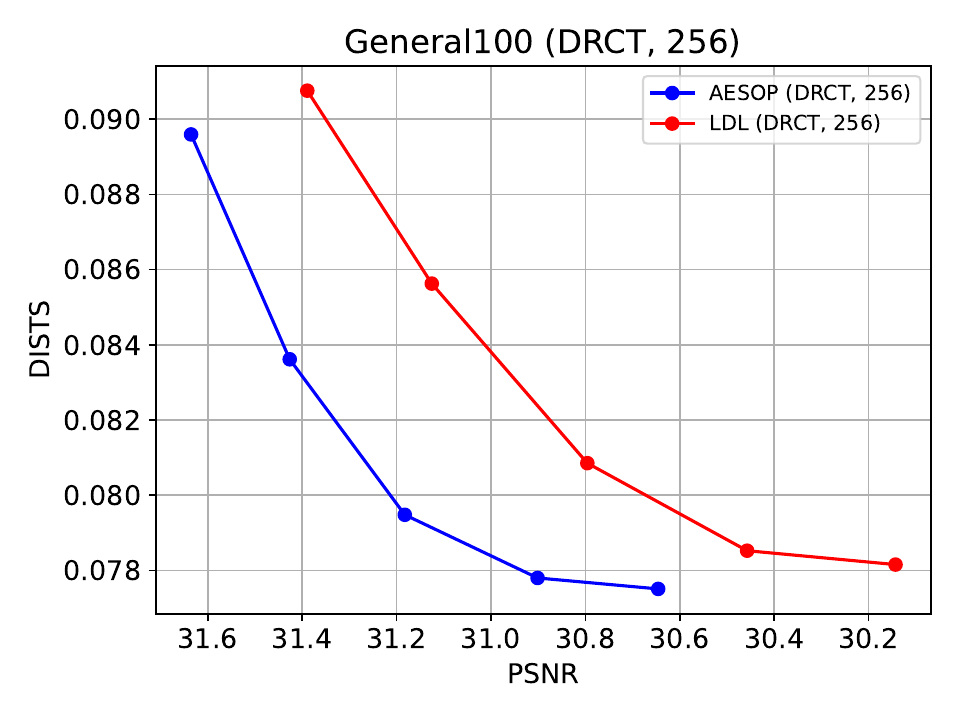}    & \includegraphics[width=0.4\textwidth]{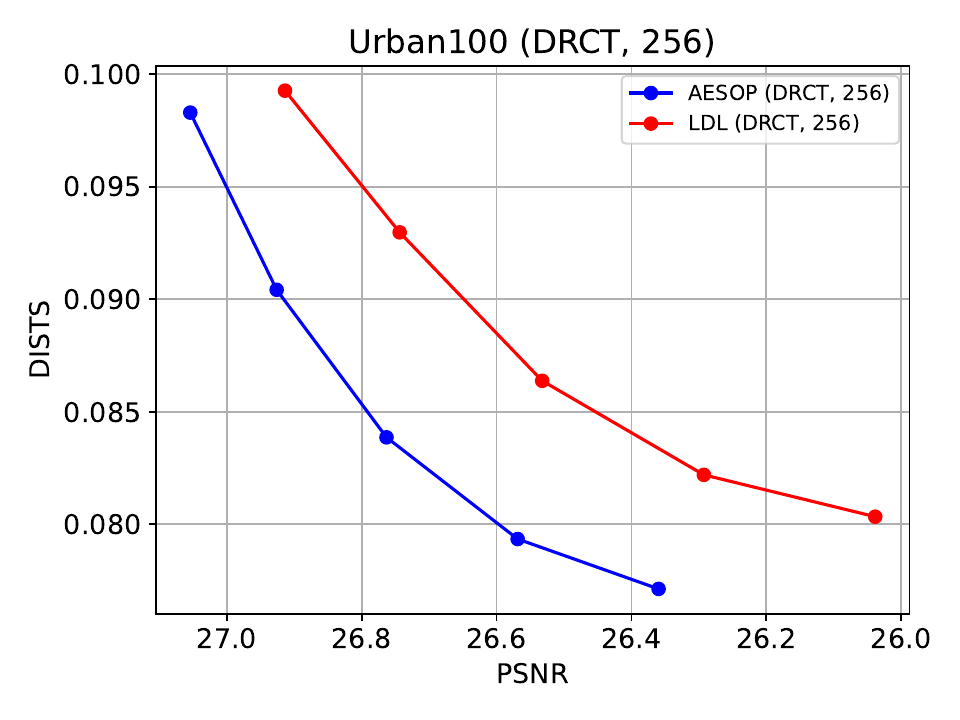} \\
        \includegraphics[width=0.4\textwidth]{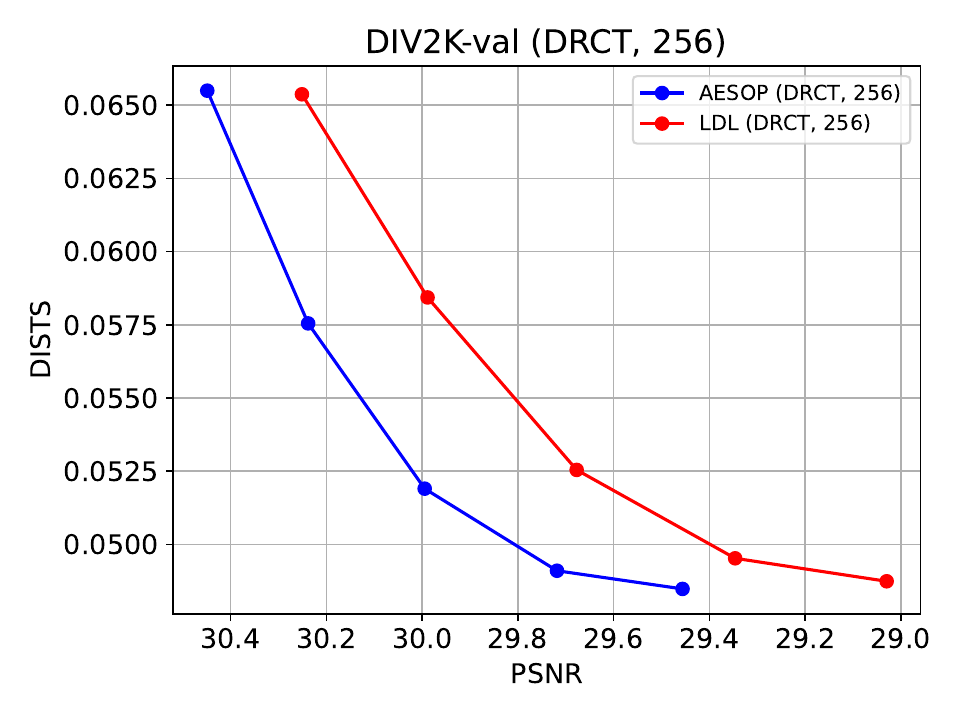}      & \includegraphics[width=0.4\textwidth]{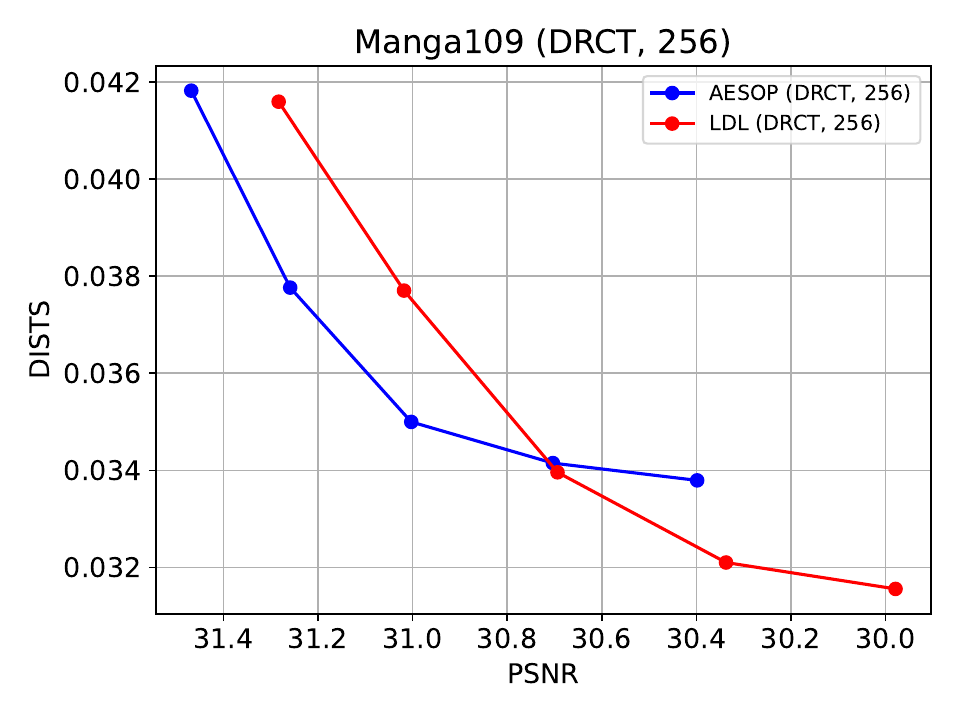} \\
        \includegraphics[width=0.4\textwidth]{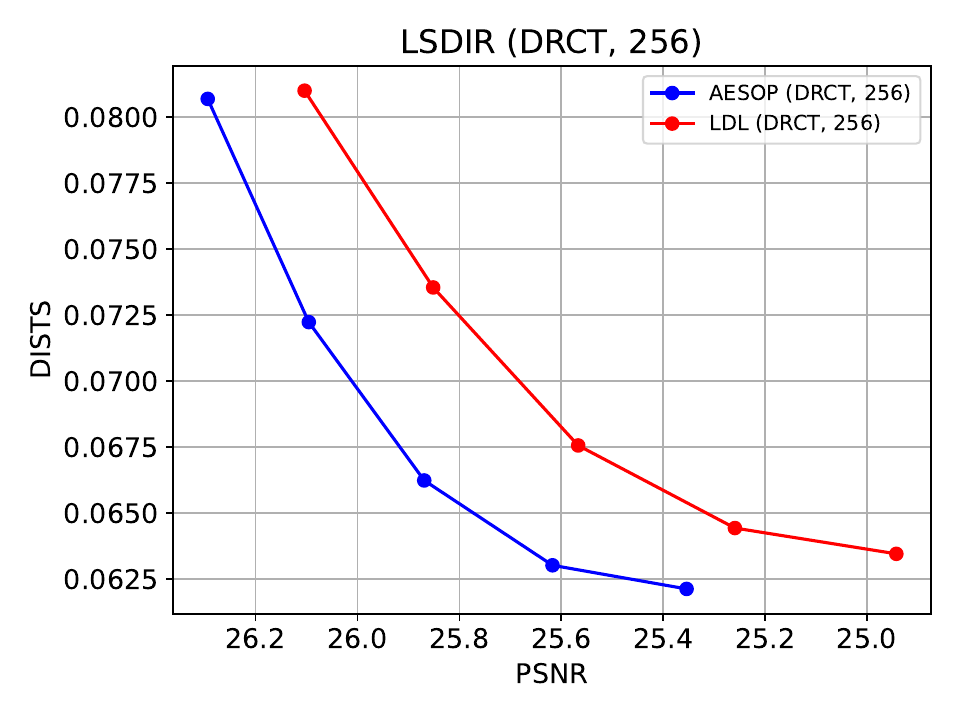}\\
    \end{tabular}
    \caption{The perception-distortion trade-off curve between AESOP and baseline methods on top of the DRCT~\cite{drct} backbone. The training HR patch size is 256. AESOP often fails to improve the performance on the Manga109 dataset.}
    \label{fig:supp_pdtradeoff_PSNR_DISTS_(DRCT, 256)}
\end{figure*}

%%%%%%%%%%%%% ============================ (SwinIR, 256)

\clearpage
\begin{figure*}[htp]
    \centering
    % First Row
    \begin{tabular}{cc}
        \includegraphics[width=0.4\textwidth]{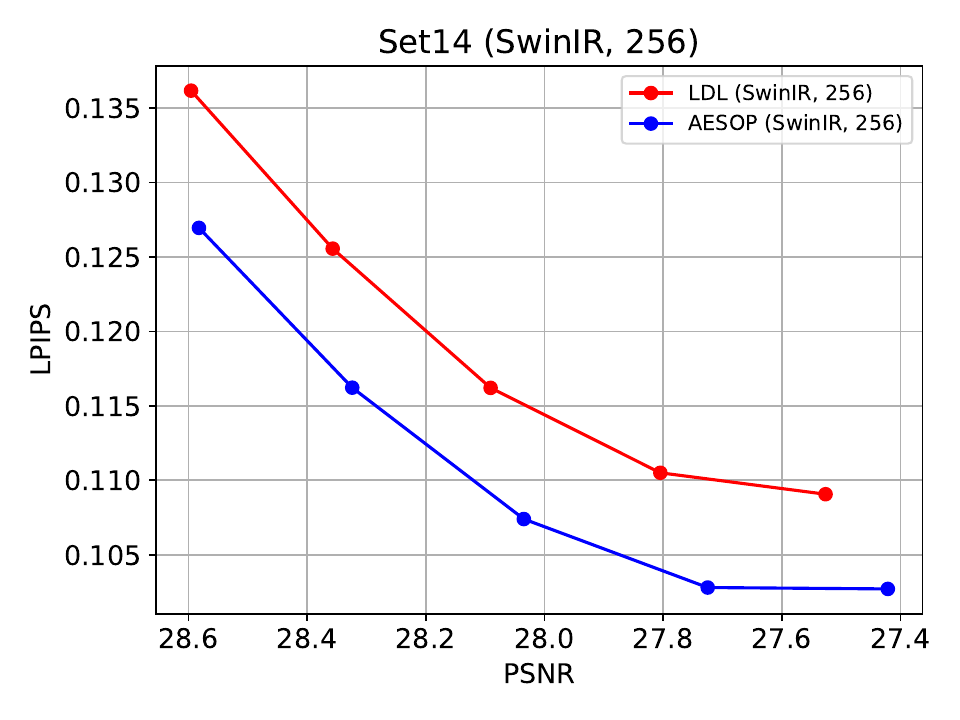}         & \includegraphics[width=0.4\textwidth]{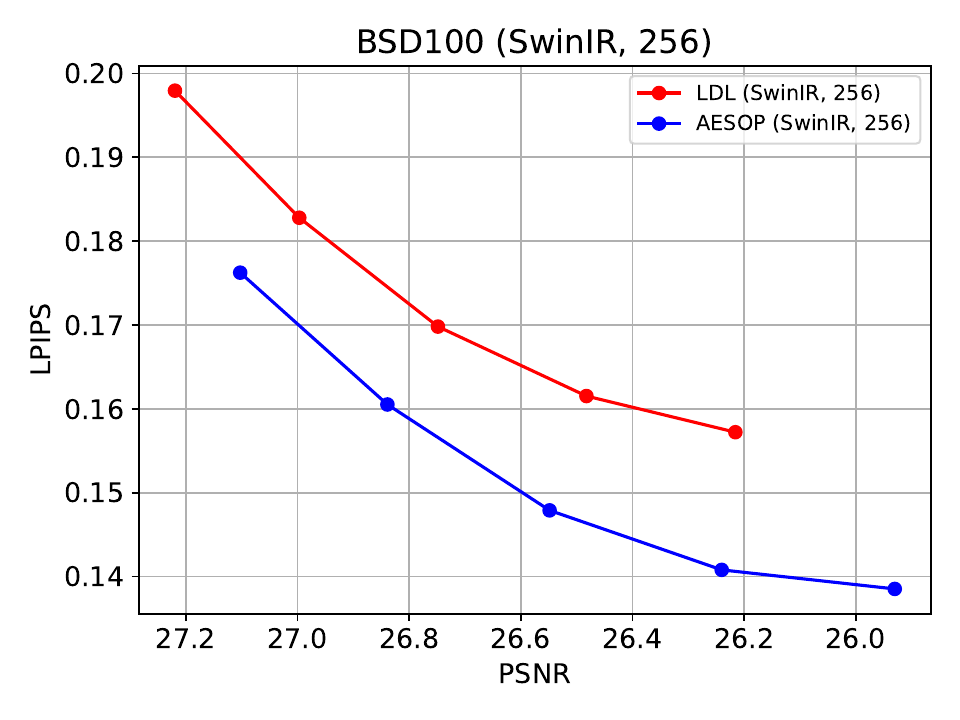} \\
        \includegraphics[width=0.4\textwidth]{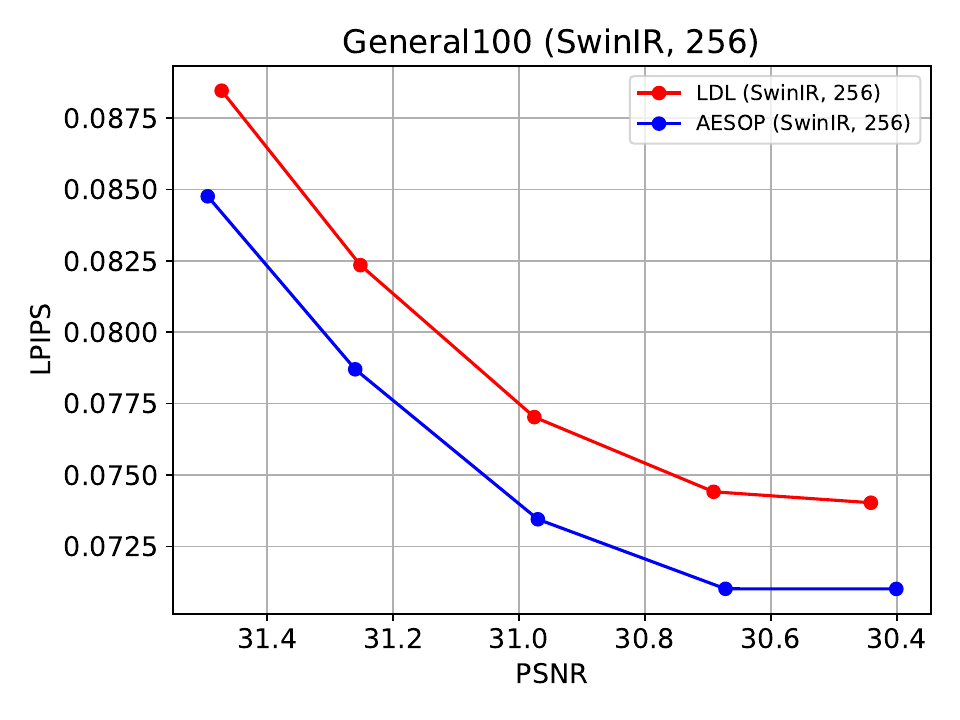}    & \includegraphics[width=0.4\textwidth]{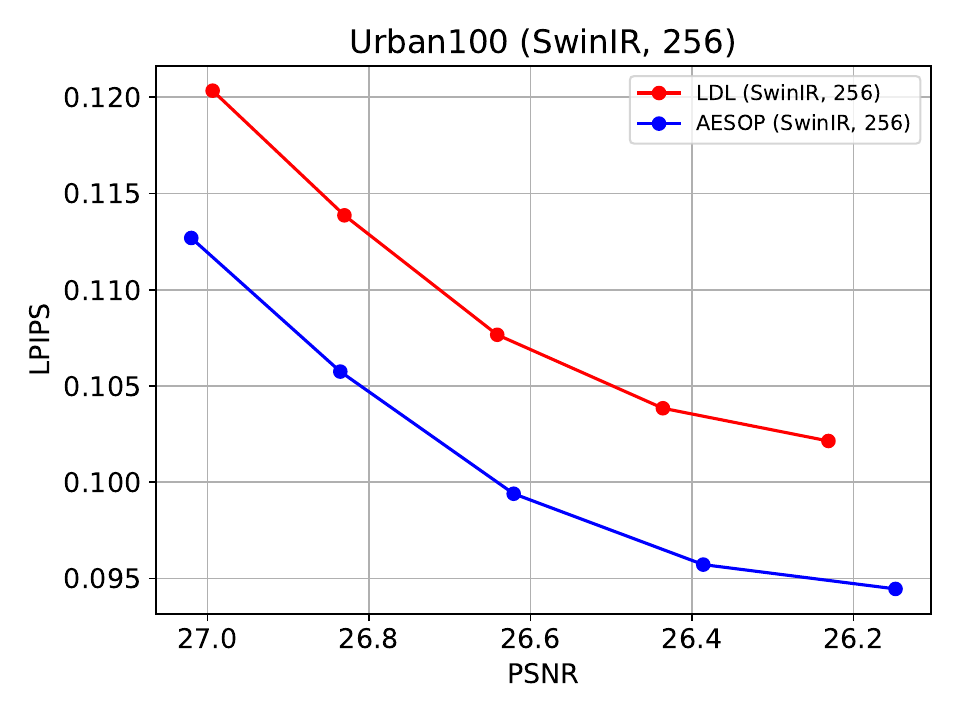} \\
        \includegraphics[width=0.4\textwidth]{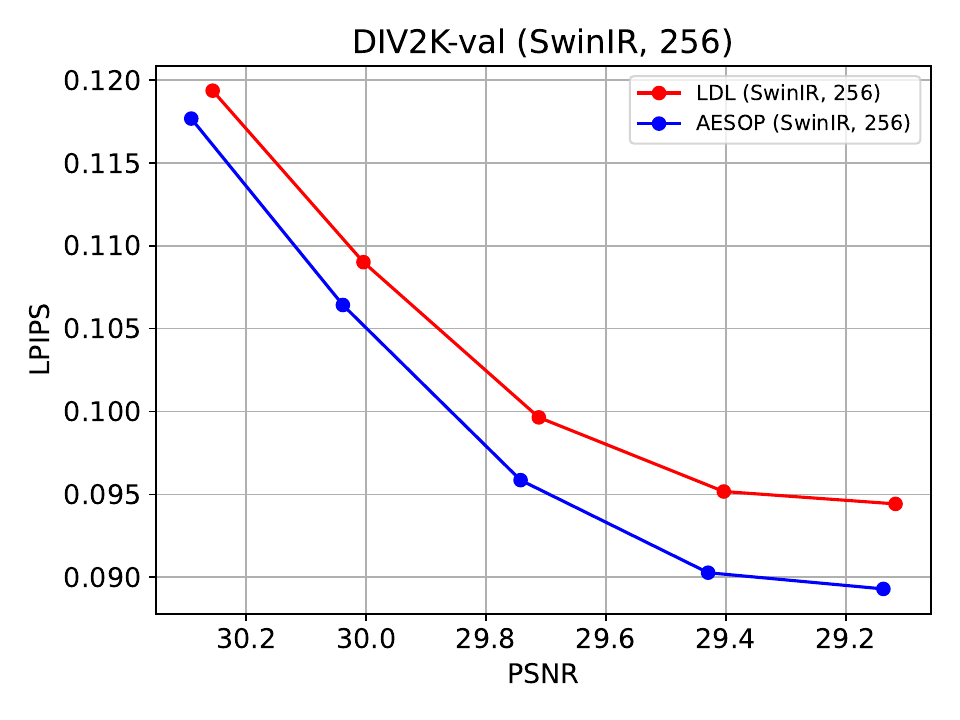}      & \includegraphics[width=0.4\textwidth]{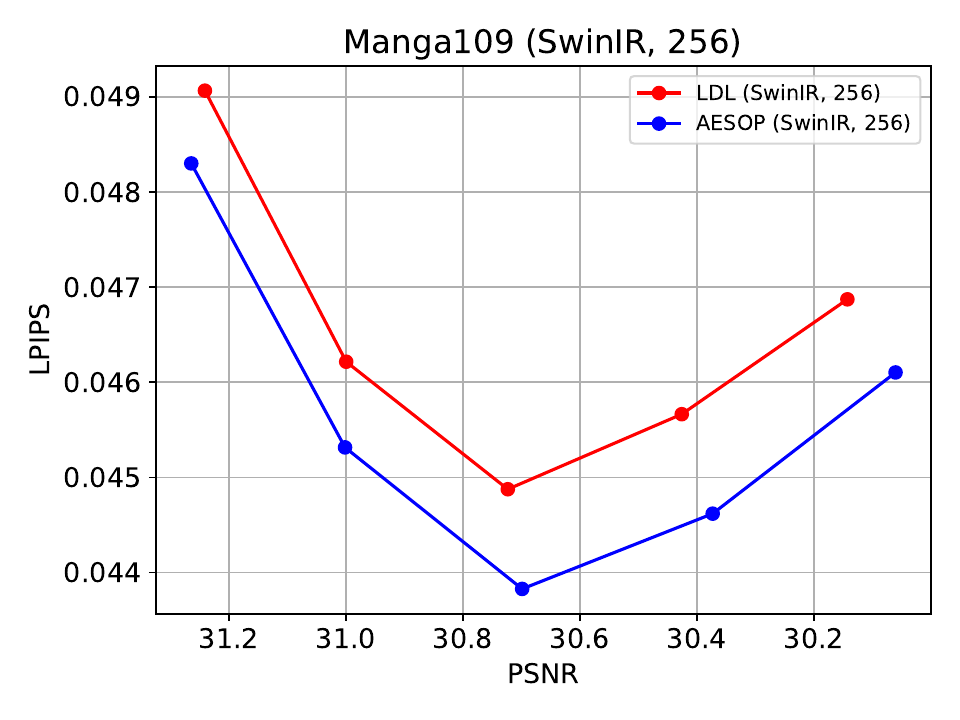} \\
        \includegraphics[width=0.4\textwidth]{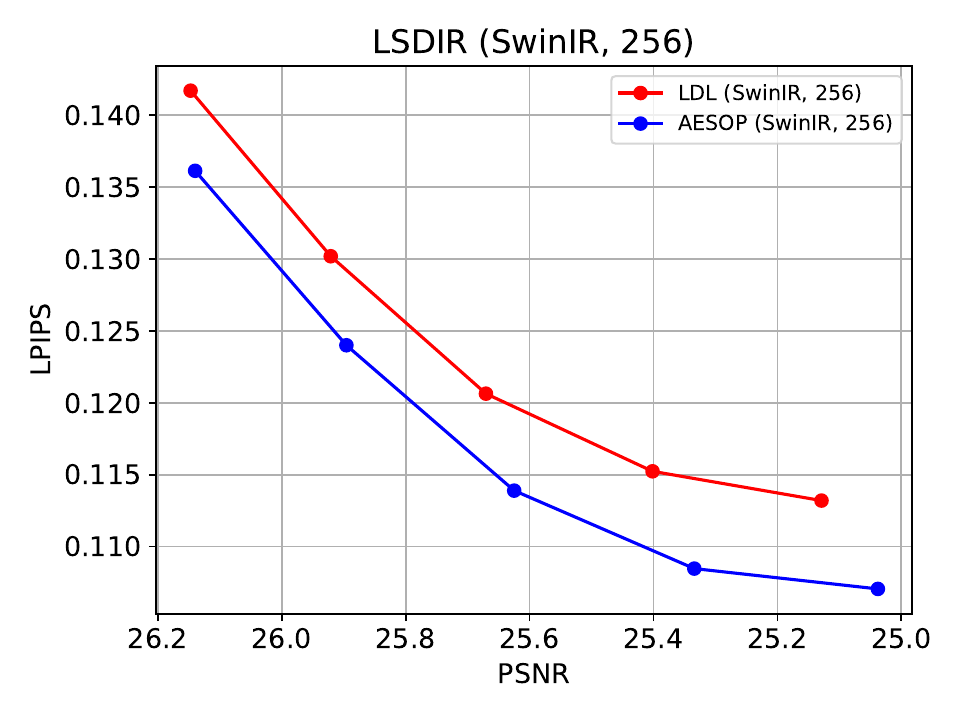}\\
    \end{tabular}
    \caption{The perception-distortion trade-off curve between AESOP and baseline methods on top of the SwinIR~\cite{SwinIR} backbone. The training HR patch size is 256. AESOP often fails to improve the performance on the Manga109 dataset.}
    \label{fig:supp_pdtradeoff_PSNR_LPIPS_(SwinIR, 256)}
\end{figure*}

\clearpage
\begin{figure*}[htp]
    \centering
    % First Row
    \begin{tabular}{cc}
        \includegraphics[width=0.4\textwidth]{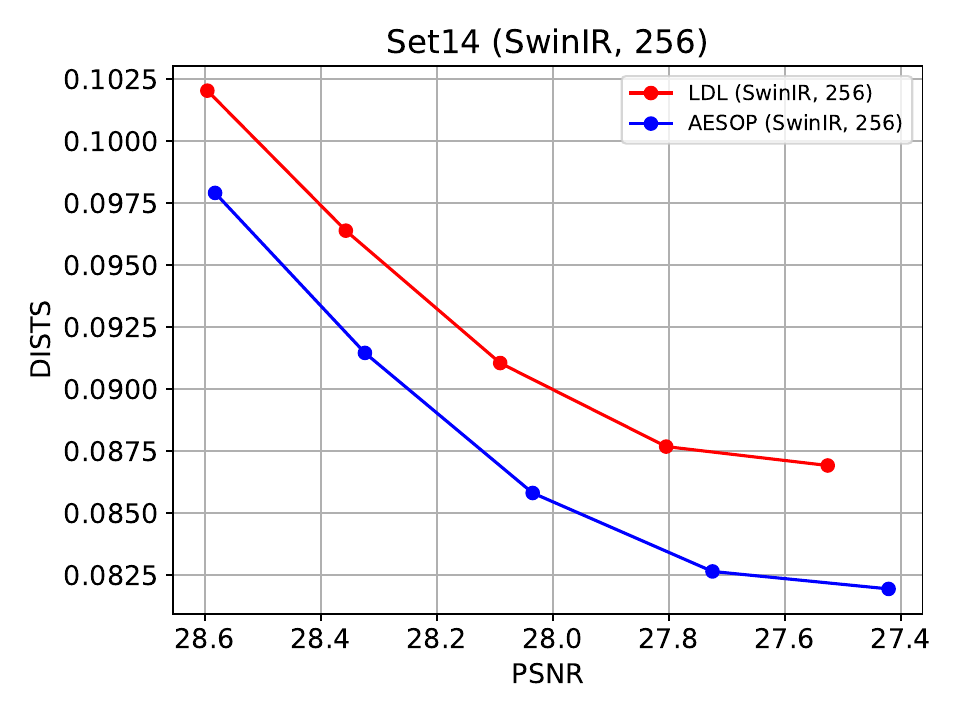}         & \includegraphics[width=0.4\textwidth]{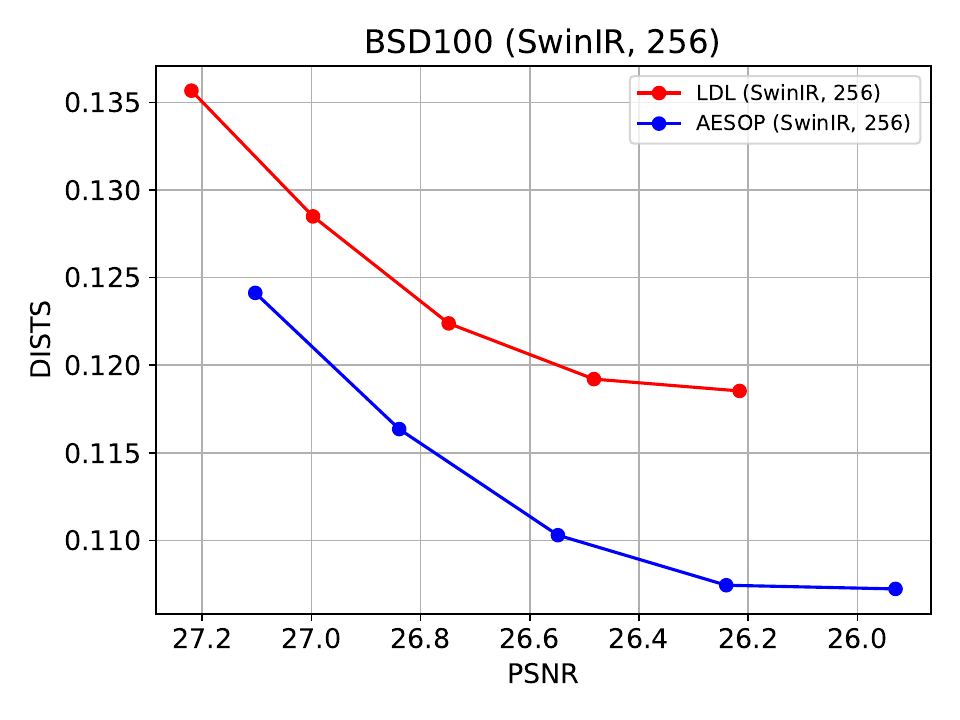} \\
        \includegraphics[width=0.4\textwidth]{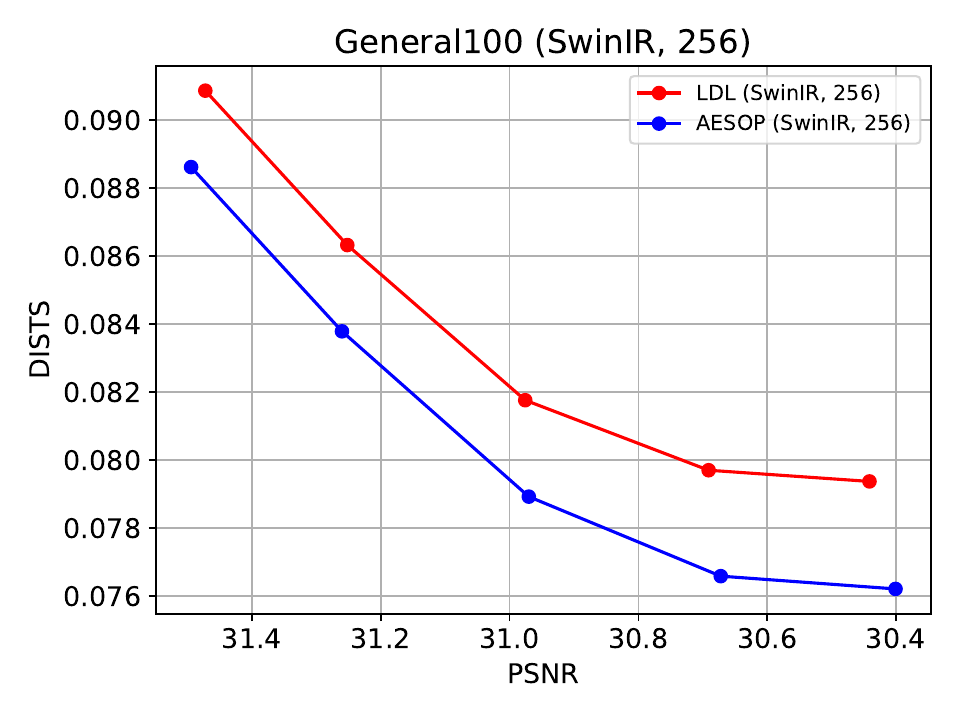}    & \includegraphics[width=0.4\textwidth]{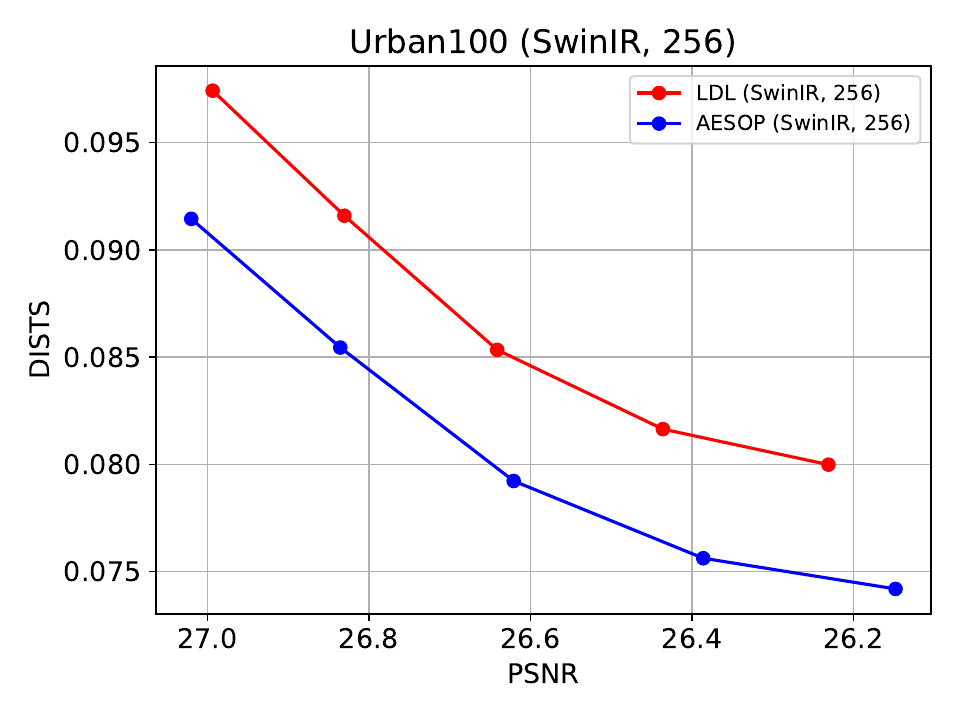} \\
        \includegraphics[width=0.4\textwidth]{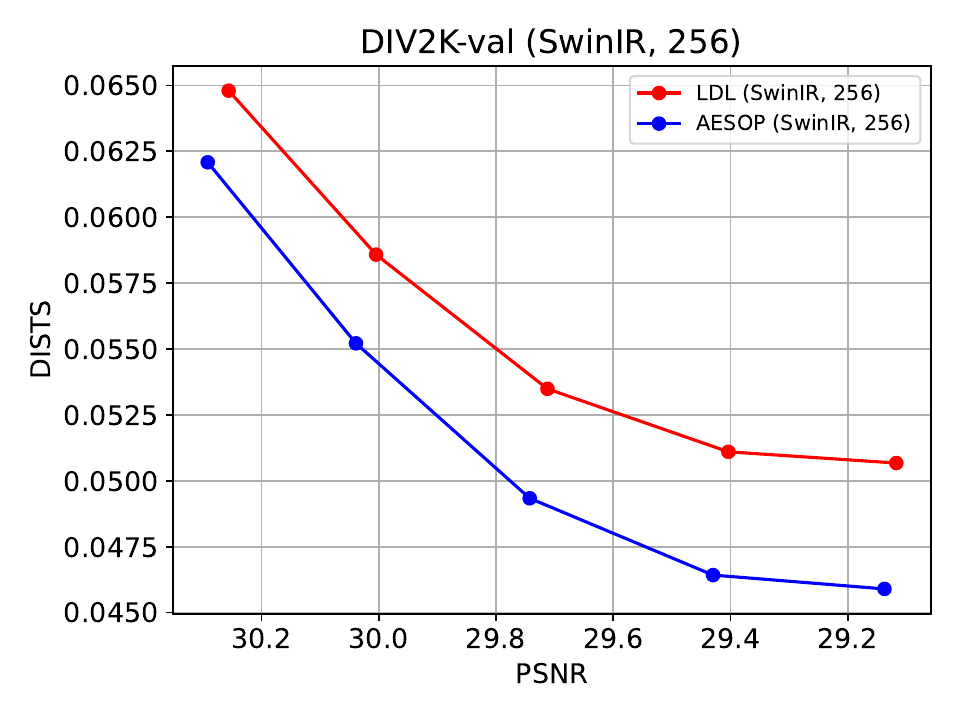}      & \includegraphics[width=0.4\textwidth]{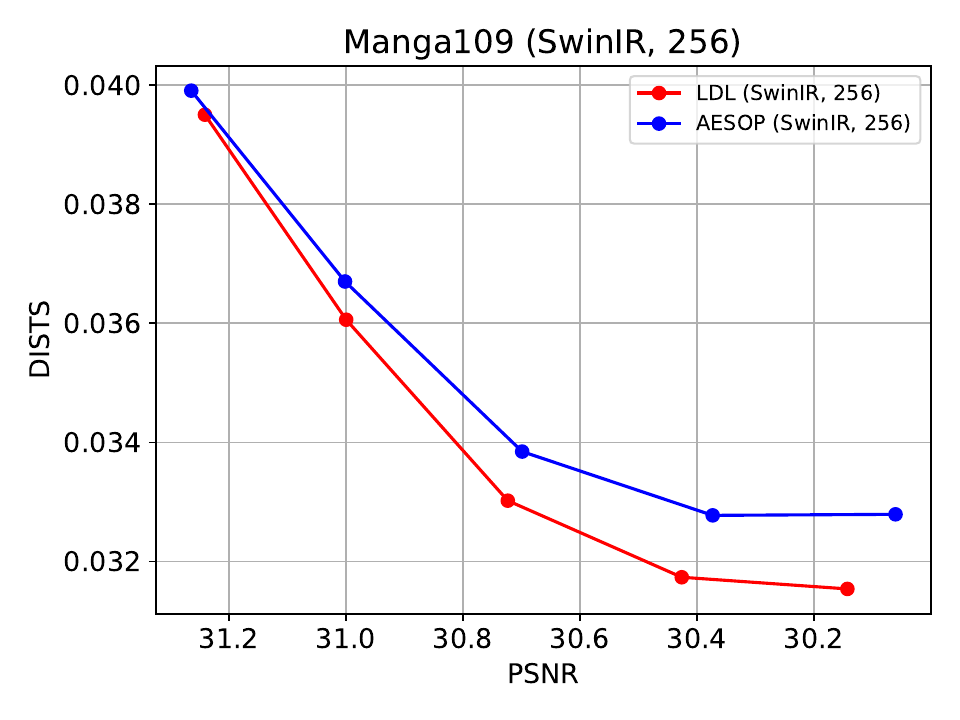} \\
        \includegraphics[width=0.4\textwidth]{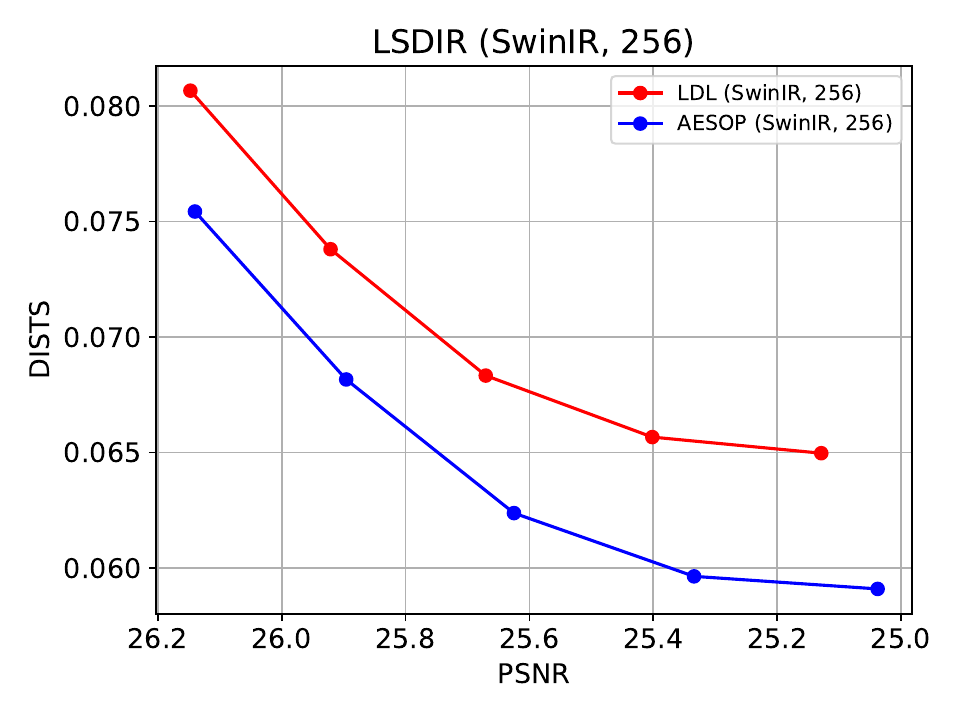}\\
    \end{tabular}
    \caption{The perception-distortion trade-off curve between AESOP and baseline methods on top of the SwinIR~\cite{SwinIR} backbone. The training HR patch size is 256. AESOP often fails to improve the performance on the Manga109 dataset.}
    \label{fig:supp_pdtradeoff_PSNR_DISTS_(SwinIR, 256)}
\end{figure*}

%%%%%%%%%%%%% ============================ (RRDB, 256)

\clearpage
\begin{figure*}[htp]
    \centering
    % First Row
    \begin{tabular}{cc}
        \includegraphics[width=0.4\textwidth]{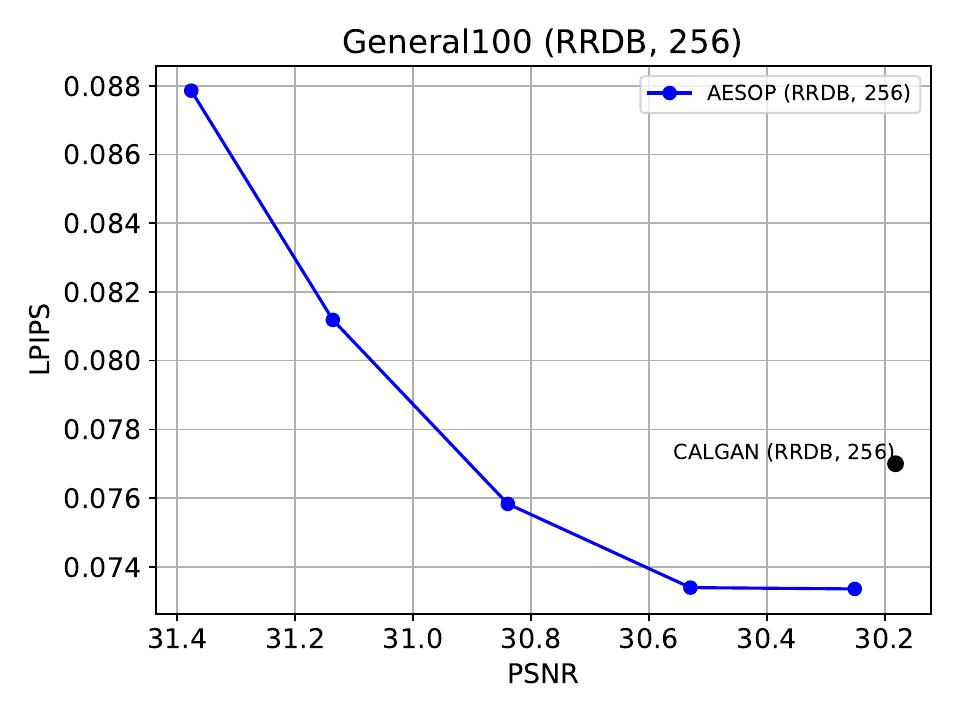}         & \includegraphics[width=0.4\textwidth]{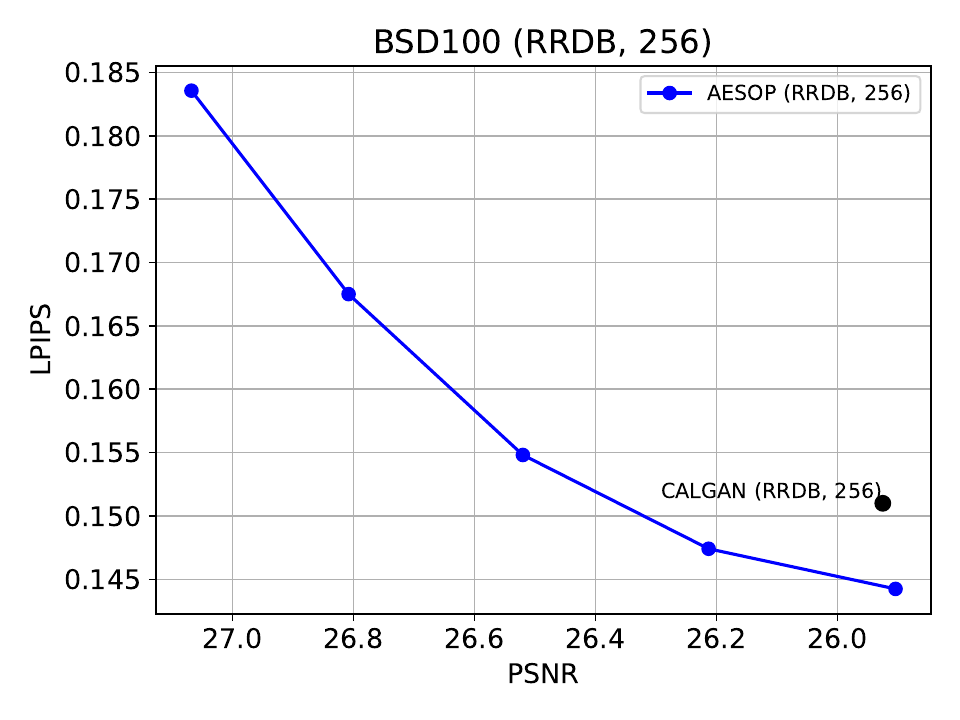} \\
        \includegraphics[width=0.4\textwidth]{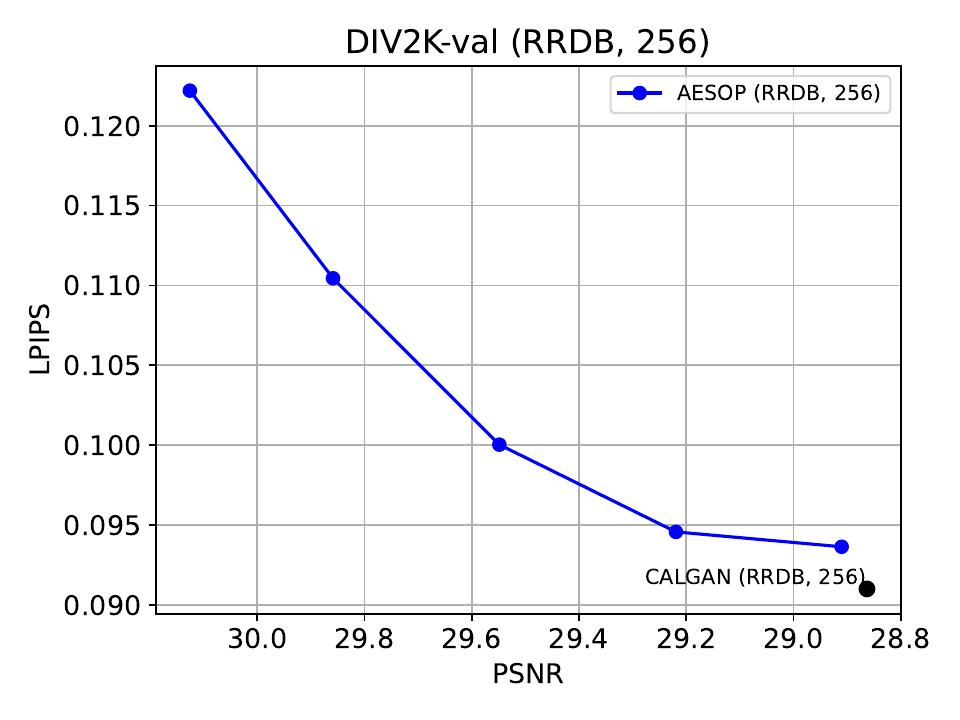}    & \includegraphics[width=0.4\textwidth]{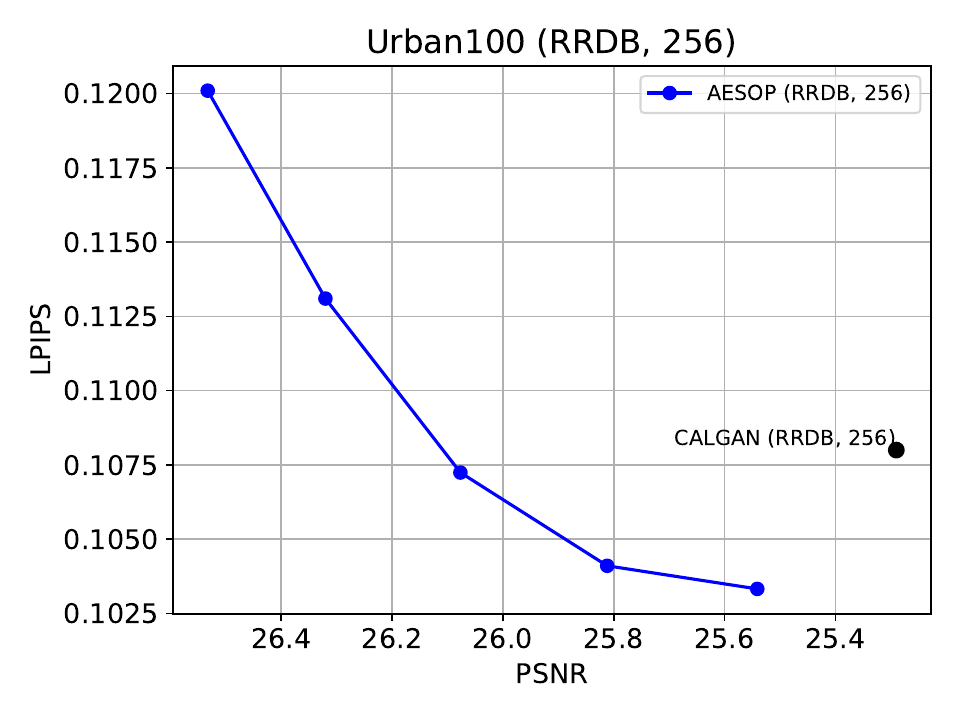} \\
%     \end{tabular}
%     \caption{The perception-distortion trade-off curve between AESOP and baseline methods on top of the RRDB~\cite{SISR7_ESRGAN} backbone. The training HR patch size is 256. AESOP mostly outperforms CALGAN~\cite{calgan} even without the MoE-discriminator.}
%     \label{fig:supp_pdtradeoff_PSNR_LPIPS_(RRDB, 256)}
% \end{figure}

% \clearpage
% \begin{figure}[htp]
%     \centering
%     % First Row
%     \begin{tabular}{cc}
        \includegraphics[width=0.4\textwidth]{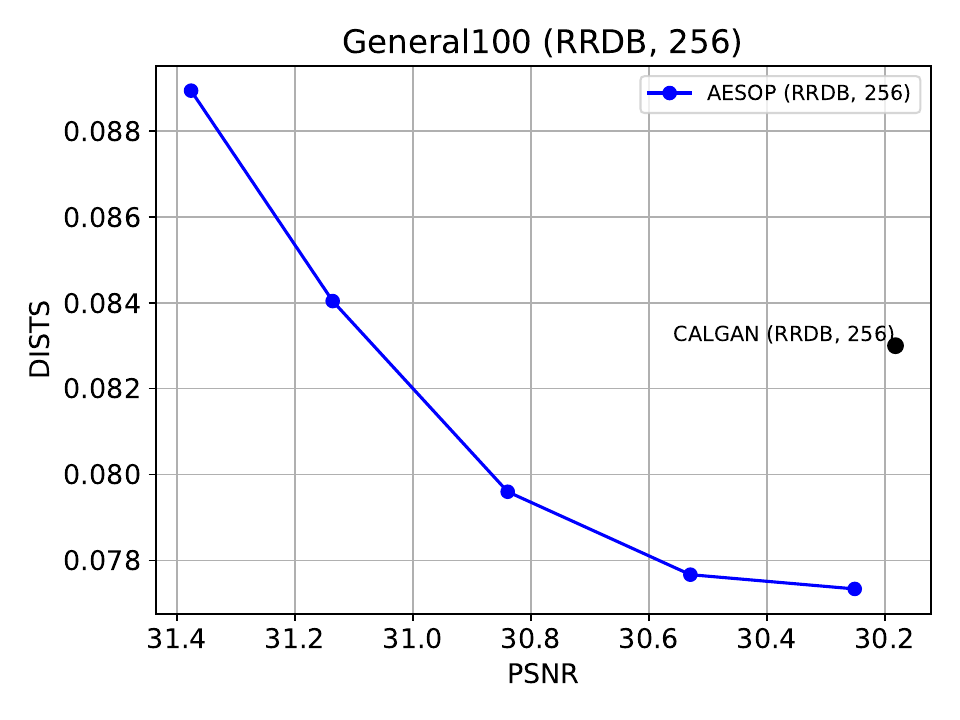}         & \includegraphics[width=0.4\textwidth]{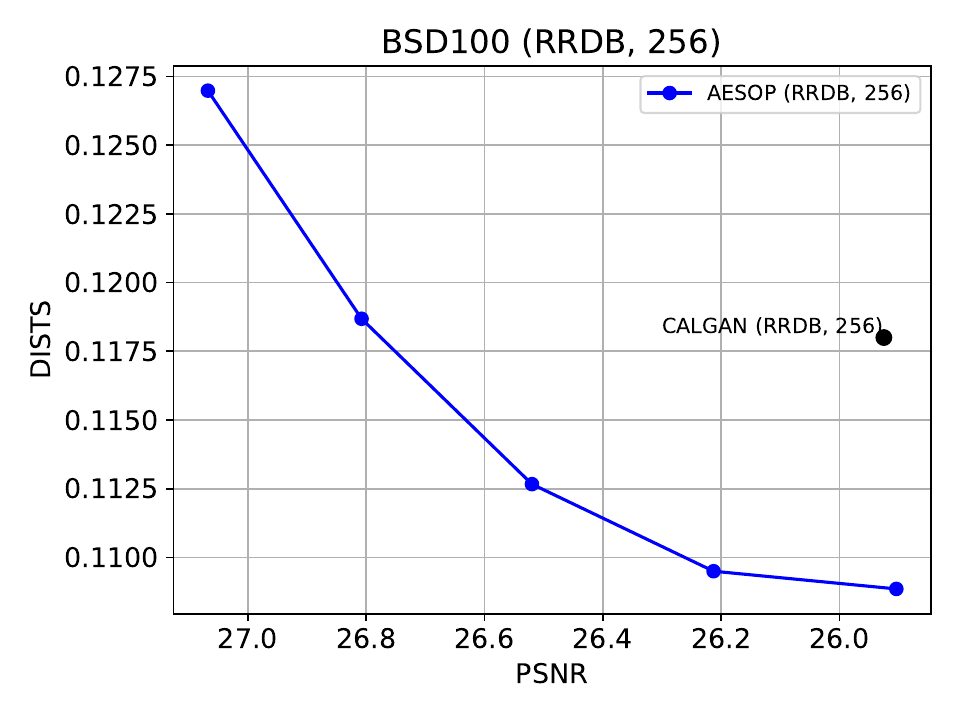} \\
        \includegraphics[width=0.4\textwidth]{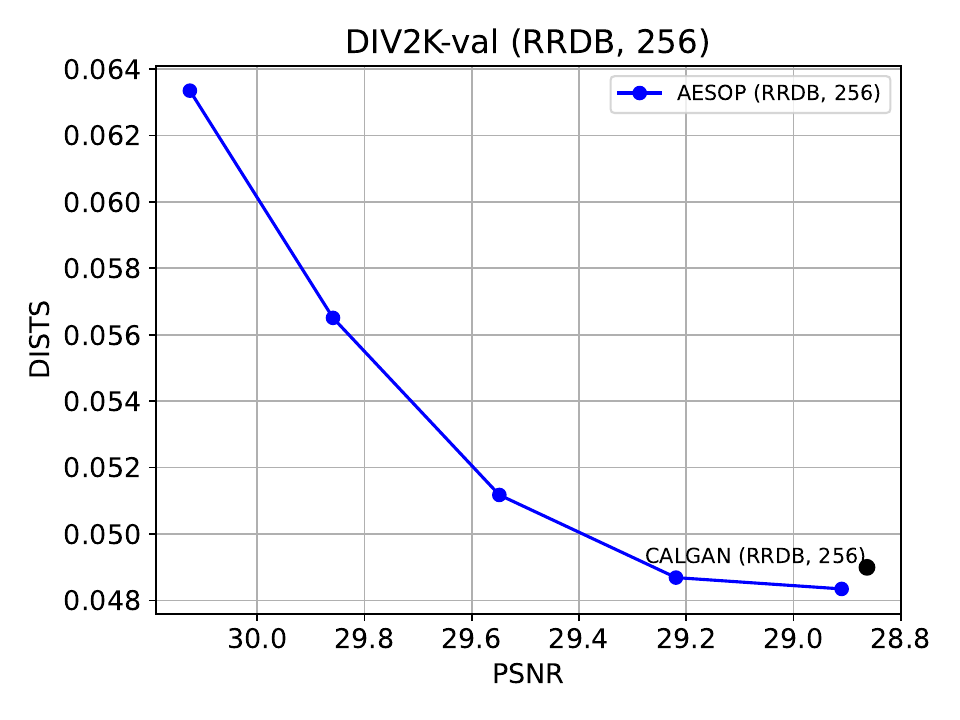}    & \includegraphics[width=0.4\textwidth]{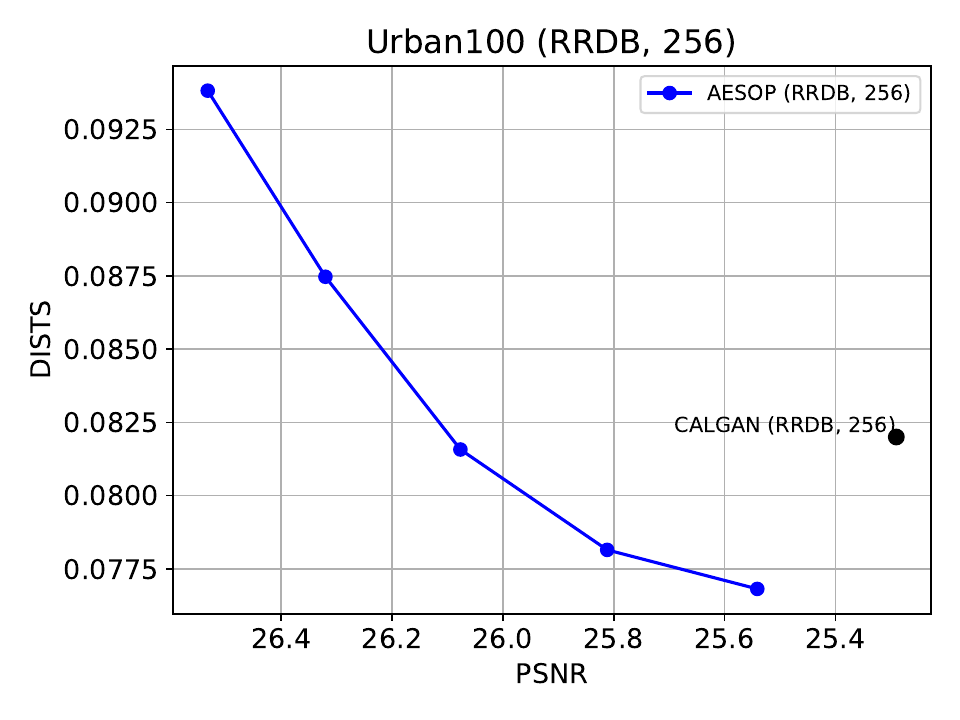} \\
    \end{tabular}
    \caption{The perception-distortion trade-off curve between AESOP and baseline methods on top of the RRDB~\cite{SISR7_ESRGAN} backbone. The training HR patch size is 256. AESOP mostly outperforms CALGAN~\cite{calgan} even without the MoE-discriminator.}
    \label{fig:supp_pdtradeoff_PSNR_DISTS_(RRDB, 256)}
\end{figure*}